\documentclass[preprint,review,12pt]{elsarticle}

\usepackage{amssymb}

\usepackage{subfig}
\usepackage{graphicx}
\usepackage{epstopdf}
\usepackage{amsmath,amssymb} 
\usepackage{upgreek}
\usepackage{color}
\usepackage[usenames,dvipsnames,table]{xcolor}
\usepackage[pagebackref=true,breaklinks=true,letterpaper=true,colorlinks,bookmarks=false]{hyperref}
\usepackage{enumitem}
\usepackage{tikz}
\usepackage{tikz-3dplot}
\usetikzlibrary{3d}
\usetikzlibrary{arrows,positioning}
\usepackage{tabularx,colortbl}
\usepackage{booktabs}
\usepackage[ruled]{algorithm2e}

\setlist{noitemsep}
\newcommand{\etal}{\textit{et al.}}

\journal{Pattern Recognition}

\begin{document}

\begin{frontmatter}



\title{TextProposals: a Text-specific Selective Search Algorithm for Word Spotting in the Wild}


\author{Llu\'is~G\'omez}
\author{Dimosthenis~Karatzas}
\address{Computer Vision Center, Universitat Autonoma de Barcelona. Edifici O, Campus UAB, 08193 Bellaterra (Cerdanyola) Barcelona, Spain. E-mail: {lgomez,dimos}@cvc.uab.cat}

\begin{abstract}
Motivated by the success of powerful while expensive techniques to recognize words in a holistic way~\cite{goel2013,Almazan2014,jaderberg2014reading}, object proposals techniques emerge as an alternative to the traditional text detectors. 
In this paper we introduce a novel object proposals method that is specifically designed for text. We rely on a similarity based region grouping algorithm that generates a hierarchy of word hypotheses. Over the nodes of this hierarchy it is possible to apply a holistic word recognition method in an efficient way. 

Our experiments demonstrate that the presented method is superior in its ability of producing good quality word proposals when compared with class-independent algorithms. We show impressive recall rates with a few thousand proposals in different standard benchmarks, including focused or incidental text datasets, and multi-language scenarios. Moreover, the combination of our object proposals with existing whole-word recognizers~\cite{Almazan2014,jaderberg2014reading} shows competitive performance in end-to-end word spotting, and, in some benchmarks, outperforms previously published results. Concretely, in the challenging ICDAR2015 Incidental Text dataset, we overcome in more than 10 percent f-score the best-performing method in the last ICDAR Robust Reading Competition~\cite{karatzas2015}. Source code of the complete end-to-end system is available at~\url{https://github.com/lluisgomez/TextProposals}.

\end{abstract}


\end{frontmatter}



\section{Introduction}
\label{sec:intro}  
Textual content in images can provide relevant information in the process of image understanding and retrieval. It can be used for image search in large collections, and in many other applications such as automatic translation, aid tools for visually impaired people, robot navigation, etc. However, robust reading of text in uncontrolled environments is a challenging task due to a multitude of factors such as the diversity of acquisition conditions, low resolution, font variability, complex backgrounds, different lighting conditions, blur, etc. 

\begin{figure}
\centering
\includegraphics[width=0.32\linewidth]{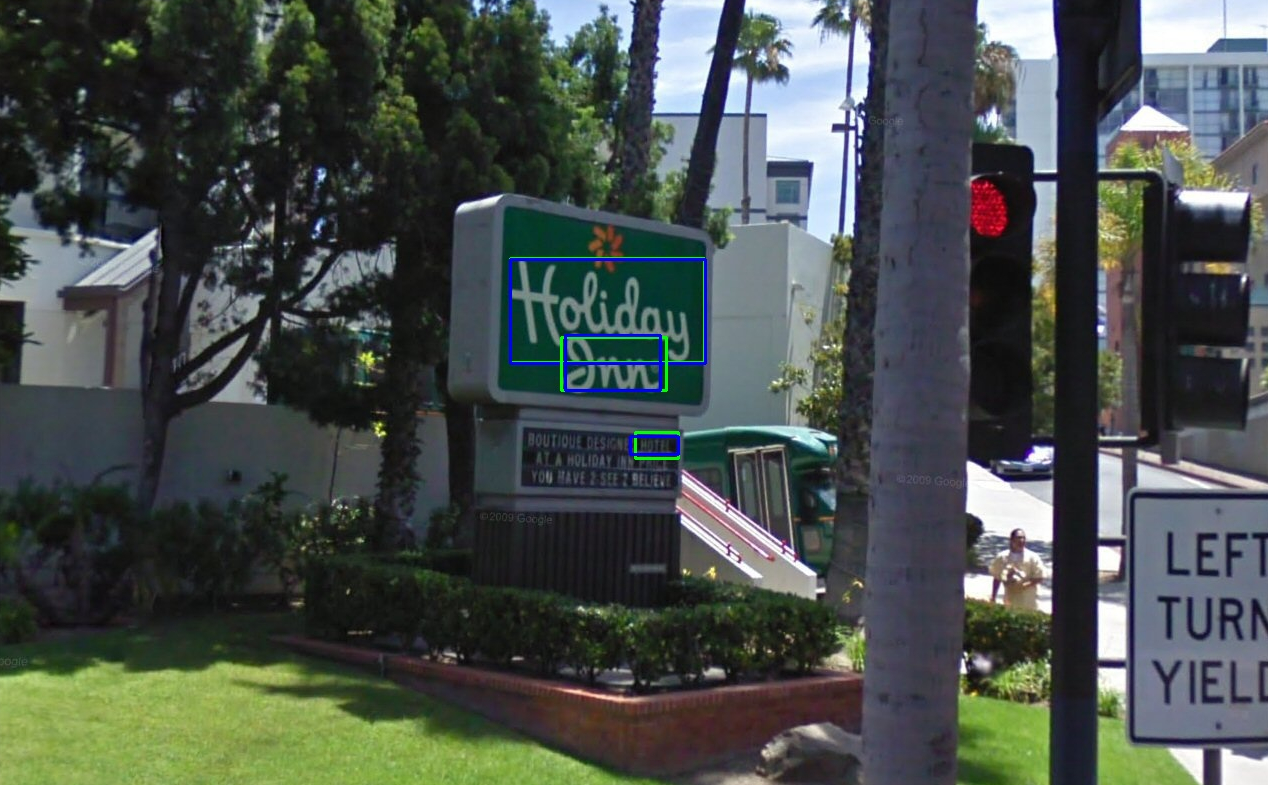}
\includegraphics[width=0.32\linewidth]{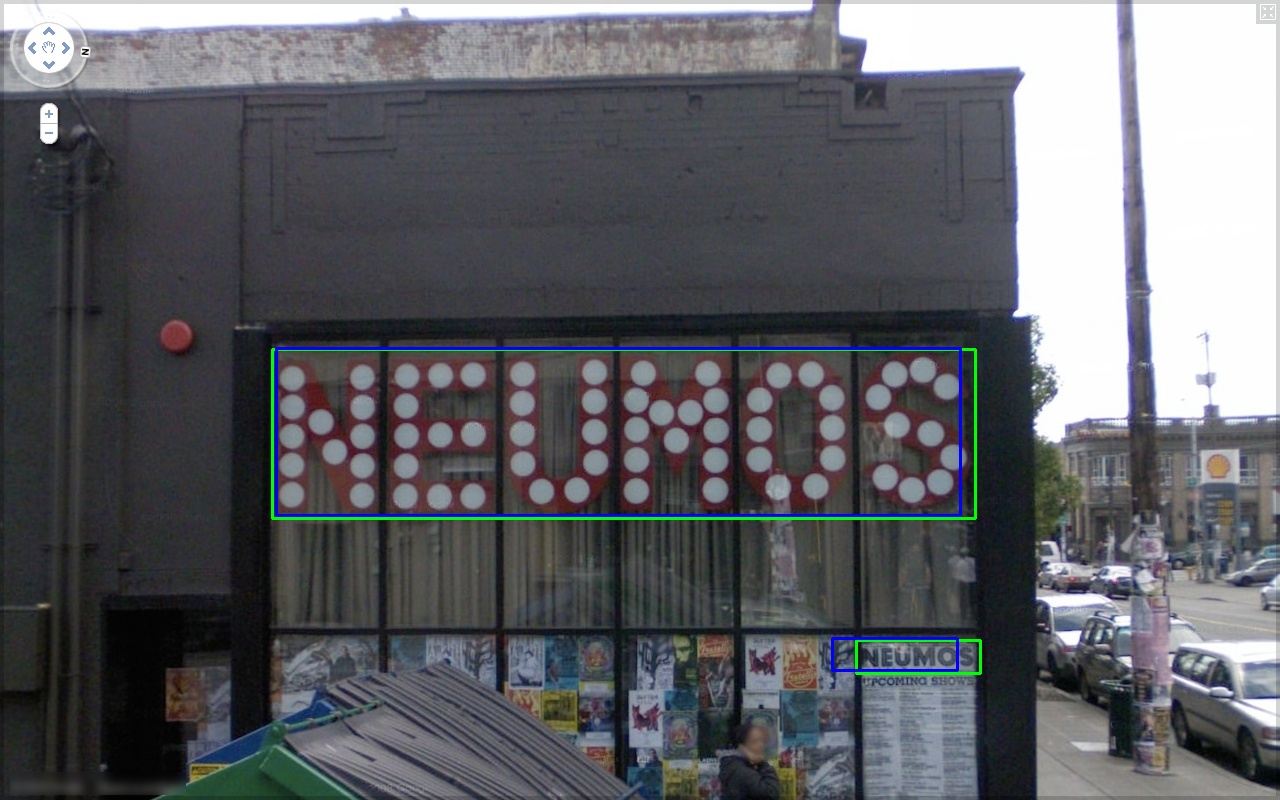}
\includegraphics[width=0.32\linewidth]{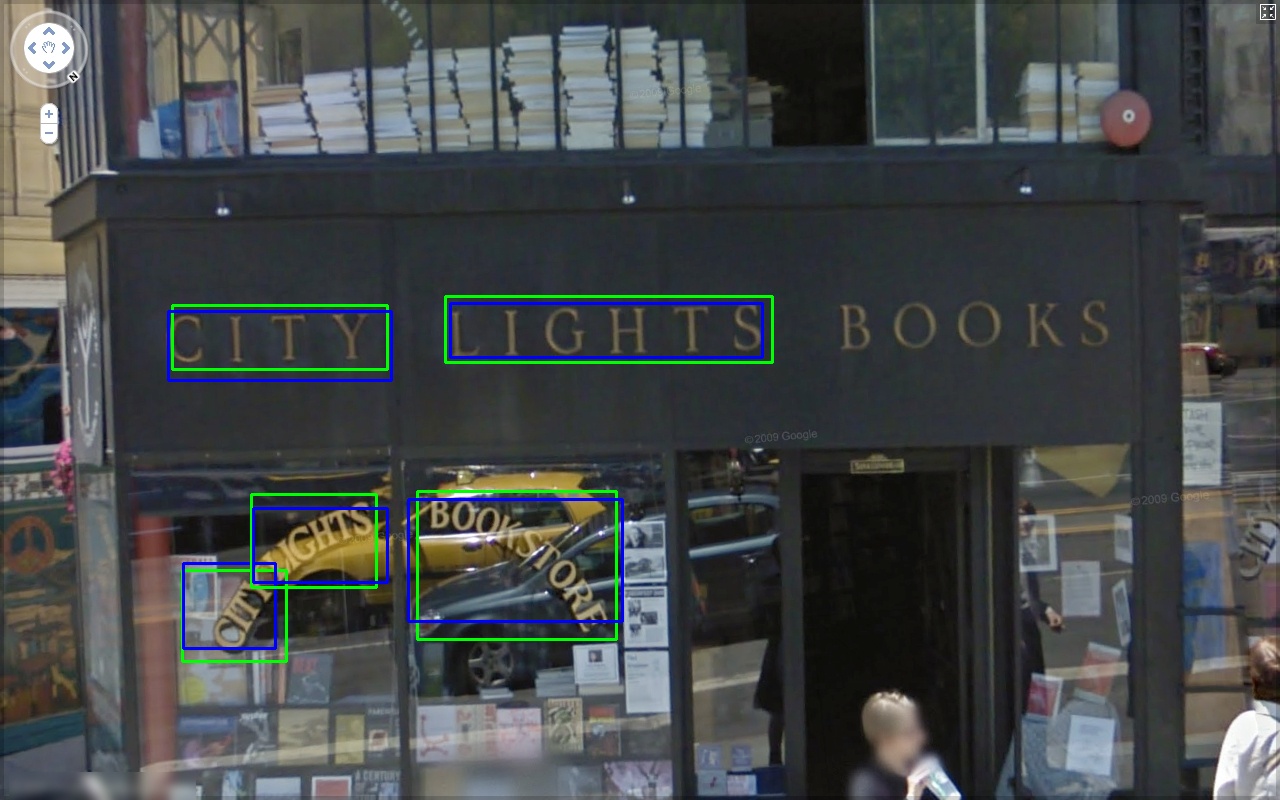}\\
\vspace{3pt}
\includegraphics[width=0.32\linewidth]{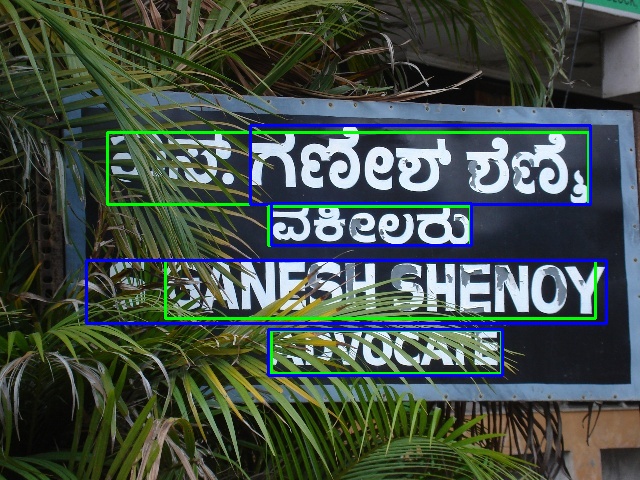}
\includegraphics[width=0.32\linewidth]{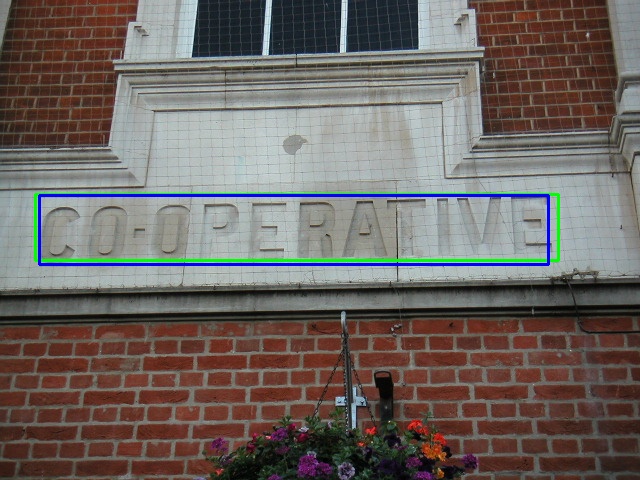}
\includegraphics[width=0.32\linewidth]{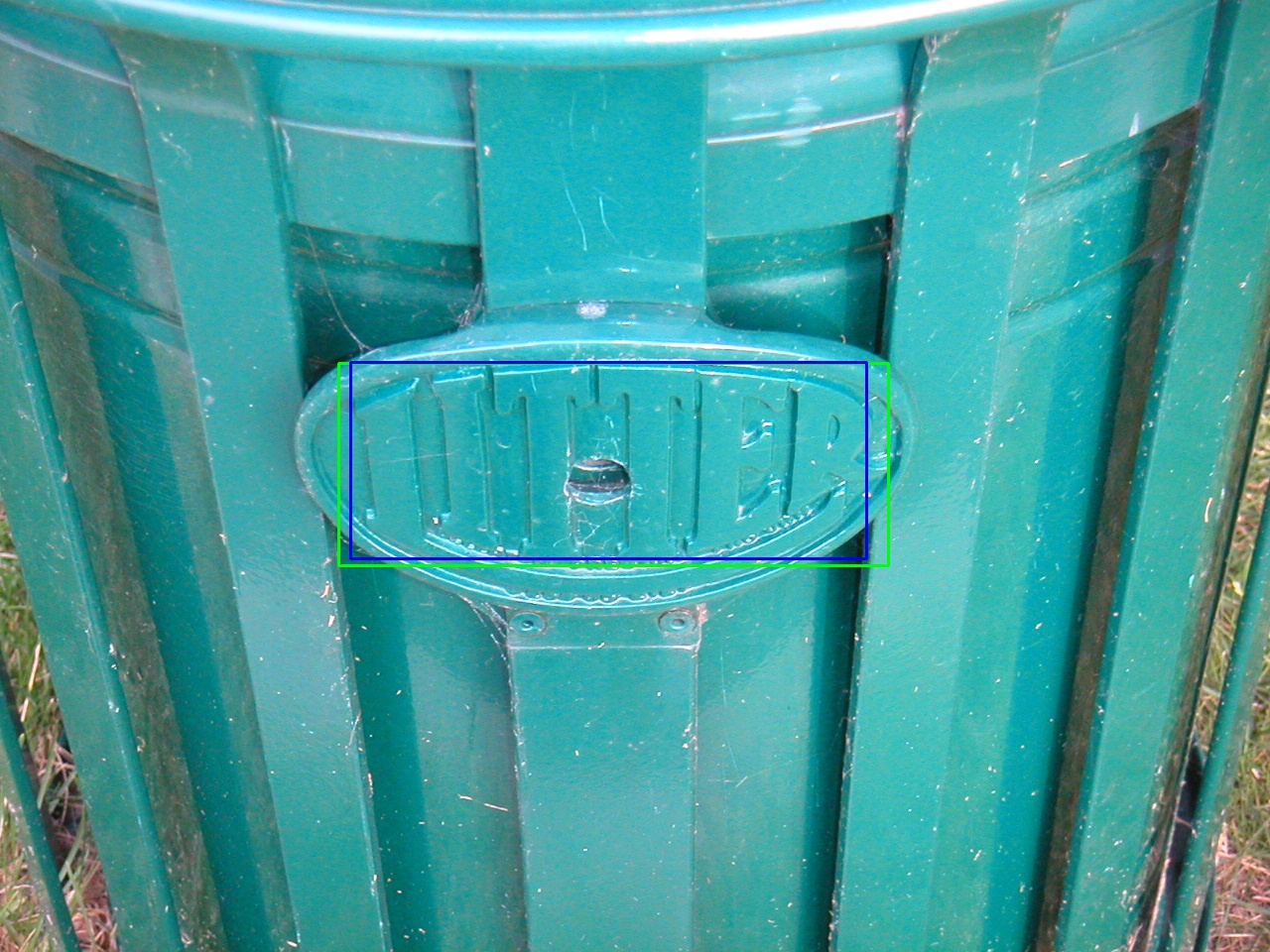}
\caption{Examples of scene text instances where region based methods performing individual character classification are prone to fail. We show the word proposals generated by our method (blue) with better Intersection over Union (IoU) over their corresponding ground truth bounding boxes (green).}
\label{fig:ill-posed-segmentation}
\end{figure}

Until recently, existing methods have approached this difficult task by relying on the detection and recognition of individual characters. So the initial set of detected character candidates are then grouped into words based on spatial and/or lexicon constraints.
Individual character segmentation is central on such methods, and has attracted vast interest from researchers in the scene text understanding field~\cite{Coates2011,Wang2012,Jaderberg2014,Neumann2012,Neumann2013b}. However, a proper character segmentation is not always feasible with existing techniques. Figure~\ref{fig:ill-posed-segmentation} shows some examples of scene text where individual character segmentation is particularly difficult: e.g. cursive text, dot-matrix text, text with low contrast, degraded characters, heavily-cluttered backgrounds, and characters affected by highlights, shadows, blur, partial occlusions, or with very low quality.

In this paper we take a different approach that generates word proposals without an explicit character segmentation. Similarly to state-of-the-art object detection systems~\cite{girshick2014,Uijlings2013}, the main idea is to produce a set of good quality word proposals and then run a classifier on the proposed bounding boxes. Jaderberg \etal~\cite{jaderberg2014reading} have recently shown that such a holistic approach can lead to impressive performance gains in end-to-end word spotting benchmarks. Their system is based on a deep Convolutional Neural Network for holistic word recognition that is applied to a set of region proposals provided by a fast detector~\cite{dollar2014fast} and a class-independent object proposals algorithm~\cite{zitnick2014}. In this paper we show how this performance gain can be even broader by using class-specific proposals.

Our main contribution is the design of a text-specific object proposals algorithm by taking into account the particular specificities of text. Our method is grounded in the same intuitions developed during the last decade for traditional text detection methods based on connected components grouping~\cite{kim2004scene,lee2009scene,yao2012detecting,koo2013scene,Yin2013,Gomez2014}: we rely on a hierarchical clustering analysis that iteratively agglomerates a set of connected components by their proximity and similarity. But there are three fundamental differences in the way we design this agglomerative process: (1) we do not assume that the initial connected components correspond to individual characters; (2) we do not aim to model the exact formation of a well-defined hierarchy of character sequences (i.e. with characters at the first level, bi-grams at the second level, etc.); and (3) we do not presume that there is a single best similarity measure that is going to generate the correct text groupings in all possible cases. Instead, we consider here of our interest any connected component extracted from the input image by casting them as potential text-parts candidates -- that may potentially be just small strokes, disjoint character-parts, or merged groups of characters such as in cursive text. Then, we build several similarity hierarchies, using complementary similarity cues, with the hope that every text instance (e.g. words) in the input image will correspond to some connected component grouping (a node) in, at least, one of such hierarchies. Figure~\ref{fig:ill-posed-segmentation} shows how our method is able to produce good quality word proposals in different real cases for which existing individual character segmentation techniques are not well-suited. 

The complete list of contributions made in this paper is as follows:

\begin{itemize}
\item We present a text-specific object proposals algorithm. To the best of our knowledge this is the first object proposals method specifically designed for text detection. As mentioned before, this approach supposes a methodological shift in the way text detectors have been traditionally designed and integrated in end-to-end pipelines. 
\item We design a novel text proposals ranking strategy, and a non-maximal suppression procedure, that are made efficient by exploiting the inclusion relation of the nodes in the hierarchies provided by our method.  
\item We provide exhaustive experimentation to compare our algorithm with well-known generic object proposals methods on the task of text detection in the wild. These experiments demonstrate that our method is superior in its ability of producing good quality word proposals in an efficient way. We show impressive recall rates with a few thousand proposals in different standard benchmarks, including focused or incidental text datasets, and multi-language scenarios.
\item We combine our text proposals algorithm with existing whole-word recognizers~\cite{Almazan2014,jaderberg2014reading}. This combination shows state-of-the-art end-to-end word spotting performance in several standard datasets, and, in some benchmarks, outperforms previously published results with a noticeable gain.
\item The source code of the complete end-to-end system is made publicly available.
\end{itemize}


\section{Related Work}
\label{sec:background}
An exhaustive survey of recent developments in scene text detection and recognition can be found in~\cite{ye2015text} and~\cite{zhu2016scene}, while corresponding surveys of earlier works on camera-based document and scene text analysis are also available in~\cite{Jung2004,Liang2005}. 

Scene text detection methods can be categorized into sliding window search methods and connected component based approaches. In the first category, Coates \etal~\cite{Coates2011} propose the use of unsupervised feature learning to generate the features for character versus background classification and character recognition. They evaluate a single-layer Convolutional Neural Network (CNN) model on each possible window of the input image at multiple scales. Wang~\etal~\cite{Wang2012} and Jaderberg~\etal~\cite{Jaderberg2014} have also used CNNs for text detection in a similar manner, but using deeper CNN models. 

Other than CNNs, more traditional hand-crafted features and statistical models have been also used within this exhaustive search approach. Wang~\etal~\cite{Wang2011} propose an end-to-end recognition system based on a sliding window character classifier using Random Ferns, with features originating from a HOG descriptor. Other methods based on HOG features have been proposed by Mishra \etal~\cite{Mishra2012} and Minetto~\etal~\cite{minetto2013t} among others.

Methods based in sliding window yield good text localization results. Their main drawback compared to connected component based methods is their high computational cost, as sliding window approaches are confronted with a huge search space. Moreover, these methods are limited to detection of a single language and orientation for which they have been trained on.

Connected component based methods, on the other hand, are based on a typical bottom-up pipeline: first, they apply a segmentation algorithm to extract regions (connected components); then, they classify the resulting regions into character or background; and finally, the identified characters are grouped into longer sequences (i.e. words or text lines). 

Yao \etal~\cite{yao2012detecting,Yao2014} have proposed a method for detecting multi-script and arbitrarily oriented text by extracting regions (connected components) from the Stroke Width Transform (SWT) domain, a local image operator proposed earlier for text detection by Epshtein \etal~\cite{Epshtein2010}.  
Other methods have build on top of the SWT algorithm by combining it with specialized edge detectors~\cite{mosleh2012image} or deep belief networks~\cite{xu2014scene}.

On the other hand, another technique extensively used to extract character candidate connected components is the Maximally Stable Extremal Regions (MSER)~\cite{Matas2004} algorithm. Neumann and Matas~\cite{Neumann2010} have proposed a method for scene text detection and recognition that performs individual MSER classification using hand-crafted region-based features (e.g. aspect ratio, compactness, etc.), demonstrating the ability of MSER algorithm for detecting promising character candidates. They further extend their work in~\cite{Neumann2012} proposing a region representation derived from MSER where character/non-character classification is done for each possible Extremal Region (ER). 

The effectiveness of MSER for character candidates detection is also exploited by Chen \etal~\cite{Chen2011}, Novikova \etal~\cite{Novikova2012}, Shi~\etal~\cite{shi2013scene,shi2014end}, Alsharif~\etal~\cite{Alsharif2014}, and Yin \etal~\cite{Yin2013} among many others. Some of this works have proposed extensions of the MSER algorithm in order to filter or enhance the regions in the component tree. Yin \etal~\cite{Yin2013} method prunes the MSER tree using the strategy of minimizing regularized variations. Chen \etal~\cite{Chen2011} and Sun~\etal~\cite{sun2015robust} have proposed the edge-enhanced and color-enhanced contrasting extremal region (CER) algorithms respectively.

Huang~\etal~\cite{huang2014robust} make use of the MSER tree as a character proposals generator and apply a deep CNN text classifier to their locations. A similar approach is also used by Sun~\etal~\cite{sun2015robust} but using a fully connected network. This way, they take advantage of both texture-based and region-based text detection approaches. Another method that combines the advantages of sliding-window and region-based approaches is proposed by Neumann and Matas in~\cite{Neumann2013b}, where characters are detected as image regions that contain certain strokes with specific orientations in specific positions.

All the aforementioned mentioned methods, either region-based or texture-based, rely in generating individual character candidates and are complemented with a post-processing step where regions assessed to be characters are grouped together into words or text lines based on spatial, similarity, and/or collinearity constraints. This way, the hierarchical and recursive structure of text has been traditionally exploited in a post-processing stage with heuristic rules~\cite{Epshtein2010,Chen2011,Neumann2010,Neumann2012}, usually constrained to search for horizontally aligned text in order to avoid the combinatorial explosion of enumerating all possible text lines. 

Yao \etal~\cite{yao2012detecting} make use of a greedy agglomerative clustering for arbitrarily oriented text components grouping in which neighboring regions are grouped together if their average alignment is under a certain threshold. Yin \etal~\cite{Yin2013} propose a self-training distance metric learning algorithm that can learn distance weights and clustering thresholds simultaneously for character groups detection in a similarity feature space. A similar metric learning approach has been also explored by the authors if this paper in~\cite{Gomez2014}.

It is important to notice that all these grouping processes are assuming that their atomic elements are well-segmented individual characters. Either because they directly validate the arrangement of character sequences using a typographic model~\cite{Epshtein2010,Chen2011,Neumann2010,Neumann2012}, or because they learn an optimal grouping strategy from examples of well-segmented character groupings~\cite{Yin2013,Gomez2014}. In this paper we introduce a scene text detection methodology that takes inspiration from existing connected components based methods but does not make such an assumtion. Thus, we do not rely in individual character segmentation, neither in a rigid grouping model to describe the way individual characters are organized in well-organized sequences. 

Over and above the specific problem of scene text detection the use of object proposals methods to generate candidate class-independent object locations has become a popular trend in computer vision in recent times~\cite{hosang2015}. The main benefits are the reduction of the search space by providing a small set of high quality locations, thus allowing the use of more expensive and powerful recognition techniques, and the ability to naturally localize objects without a fixed aspect ratio. Object proposals algorithms are aligned with the object-level saliency detection paradigm~\cite{huo2016object,gonzalez2016perceptual} in modeling a selective process to guide the recognition analysis towards particular regions of interest in the image. 

In general terms we can distinguish between two major types of object proposals methods: the ones that make use of exhaustive search to evaluate a fast to compute objectness measure~\cite{cheng2014,zitnick2014}, and the ones where the search is driven by segmentation and grouping processes~\cite{Uijlings2013,manen2013,krahenbuhl2014}. 

Overall, generic object proposals algorithms are designed to target objects that can be isolated from its background as a single body: grouping-based methods do it by agglomerating adjacent regions; and most of the sliding window based methods do it intrinsically as they are actually trained with examples of such object type. Thus, in their majority these generic methods are not adequate for text detection, just because they are designed for a different task.

However, the use of generic object proposals techniques for scene text understanding has been exploited recently by Jaderberg~\etal~\cite{jaderberg2014reading} with impressive results. Their end-to-end pipeline combines object proposals from the EdgeBoxes~\cite{zitnick2014} algorithm and a trained aggregate channel features detector~\cite{dollar2014fast} with a powerful deep Convolutional Neural Network for holistic word recognition. Still, their method uses a CNN-based bounding box regression module on top of region proposals in order improve their quality. In this paper we design a text-specific selective search method that, contrary to existing generic algorithms, aims directly to the detection of text component groupings. Our method is similar to the generic selective search algorithm~\cite{Uijlings2013}, but differs from it in many aspects by taking into account the specificities of text regions, that are fundamentally different of the generic notion of object as normally used in Computer Vision research.

\section{TextProposals: a text-specific selective search algorithm}
\label{sec:method}



Our text-specific object proposals algorithm is grounded on a basic ``segmentation and grouping'' procedure: first we extract connected components from the input image, and then we group them by their similarity using the Single Linkage Clustering (SLC) method. This clustering analysis produces a dendrogram where each node corresponds to a group of connected components and defines a bounding box proposal. The main drawback of this simple ``segmentation and grouping'' approach is that we must found the optimal segmentation strategy and the optimal similarity metric in order to robustly deal with the extreme variability of scene text. The choice of a single segmentation and similarity metric, no matter which,  will often result in missing detections for some particular text instances.

\begin{figure*}
\centering
\tikzstyle{block} = [canvas is zy plane at x=0]
\tikzstyle{block_D} = [canvas is zy plane at x=0,text width=19.15em]
\tikzstyle{block_P} = [canvas is zy plane at x=0]
\tikzstyle{block_S} = [text width=9em]
\tikzstyle{block_DS} = [text width=5em]
\tikzstyle{line} = [draw]
\tikzstyle{edge} = [draw]
\subfloat[The basic ``segmentation and grouping'' procedure used in our method: first we extract connected components from the input image, then we group the connected components by their similarity using the SLC algorithm. Each node in the dendrogram generated this way corresponds to a bounding box proposal.]{
\begin{tikzpicture}[node distance = 0.5cm, auto]
    \node [block_S] (RGB) {\includegraphics[width=\linewidth]{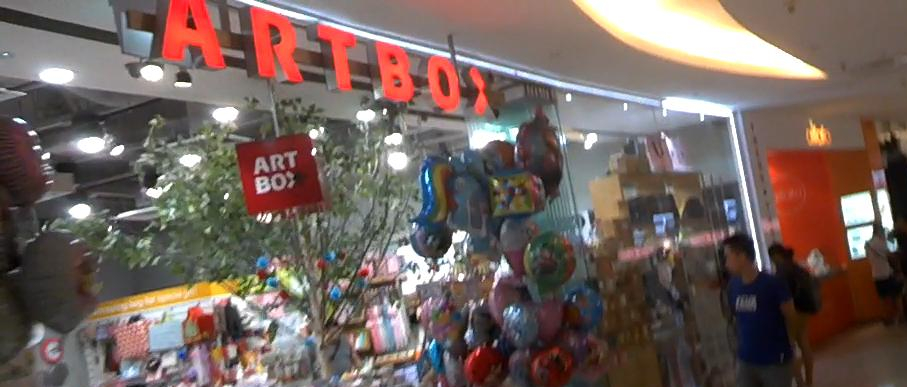}};
    \node [above=0.02cm of RGB,inner sep=0,outer sep=0] {\small{Input image}};
    \node [block_S, right=0.5cm of RGB] (mser) {\includegraphics[width=\linewidth]{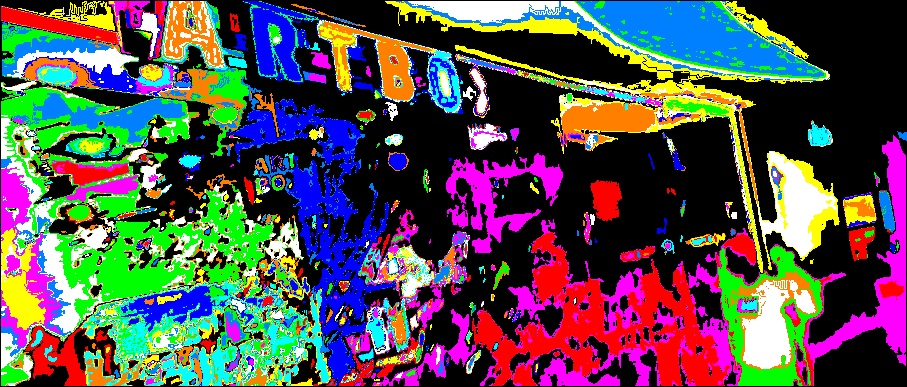}};
    \node [above=0.02cm of mser,inner sep=0,outer sep=0] {\small{Segmentation}};
    \draw [-{latex}] (RGB) -- (mser);
    \node [block_DS, right=0.5cm of mser] (D1) {\includegraphics[width=\linewidth]{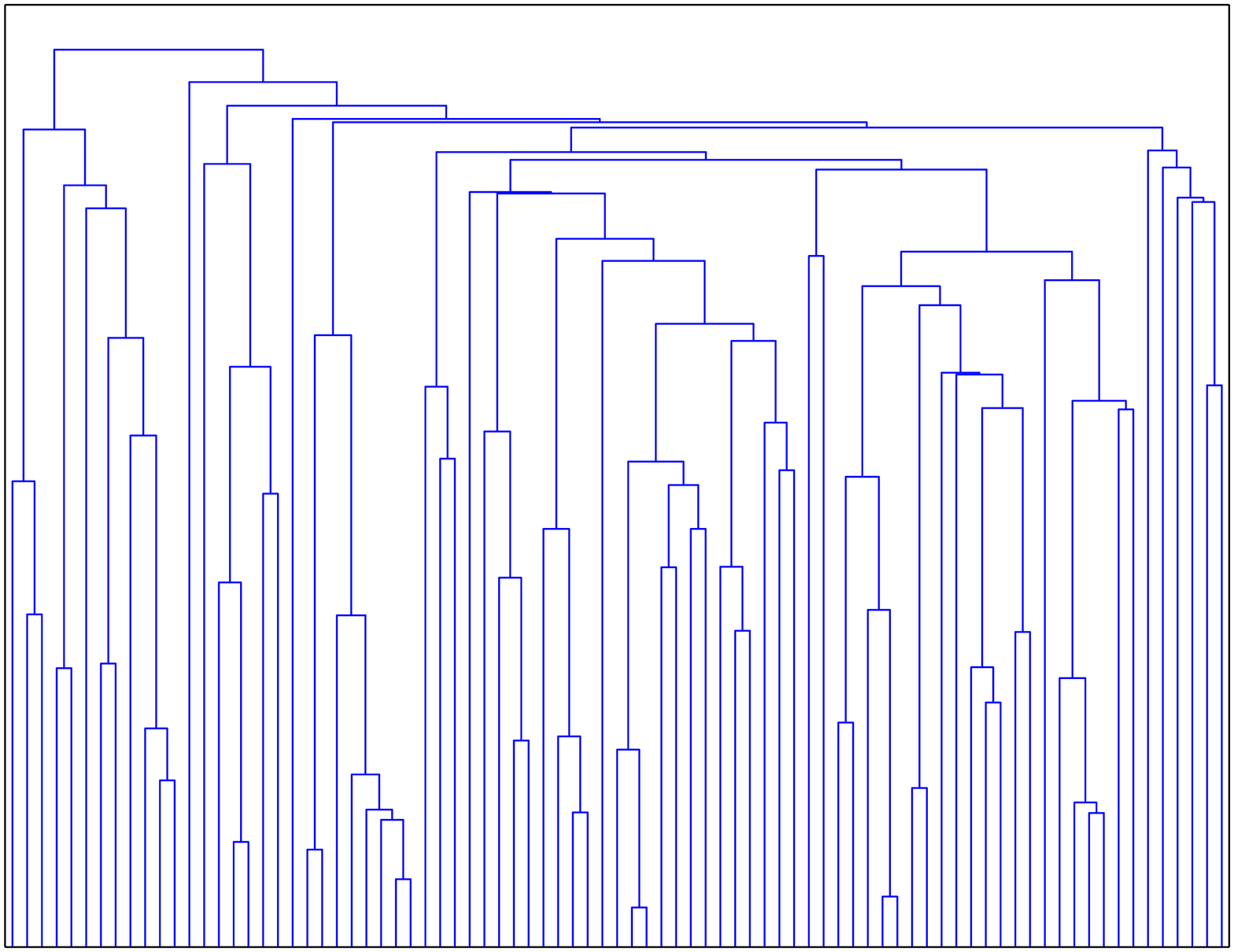}};
    \node [above=0.005cm of D1,inner sep=0,outer sep=0] {\small{Grouping}};
    \node [text width=4em,inner sep=0,outer sep=0] at (10.65,0.45) (W) {\includegraphics[width=\linewidth]{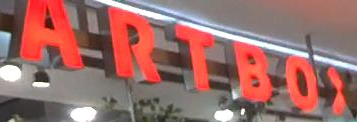}};
    \node [text width=2.55em,inner sep=0,outer sep=0] at (10.5,-0.55) (W2) {\includegraphics[width=\linewidth]{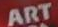}};
    \draw [-{latex}] (mser) -- (D1);
    \node[circle,draw=red,ultra thick] at (7.8,-0.5) (C) {};
    \node[circle,draw=red,ultra thick] at (8.65,-0.2) (C2) {};
    \path [-{latex},draw=red,fill=red, thick] (C) edge[bend left] node [left] {} (W);
    \path [-{latex},draw=red,fill=red, thick] (C2) edge[right] node [bend left] {} (W2);
\end{tikzpicture}
}
\\
\subfloat[The TextProposals algorithm increases the overall detection recall of the basic ``segmentation and grouping'' (a) by considering several input channels and scales, and several similarity measures. A ranking strategy prioritizes the best word proposals found.]{
\begin{tikzpicture}[x={(0cm,0.1cm)},y={(0.03cm,0.07cm)},z={(0.1cm,0.0cm)},scale=1.6, every node/.style={scale=1.6}]

    \node [canvas is zy plane at x=-22] (Ip2) {\includegraphics[width=0.5\linewidth]{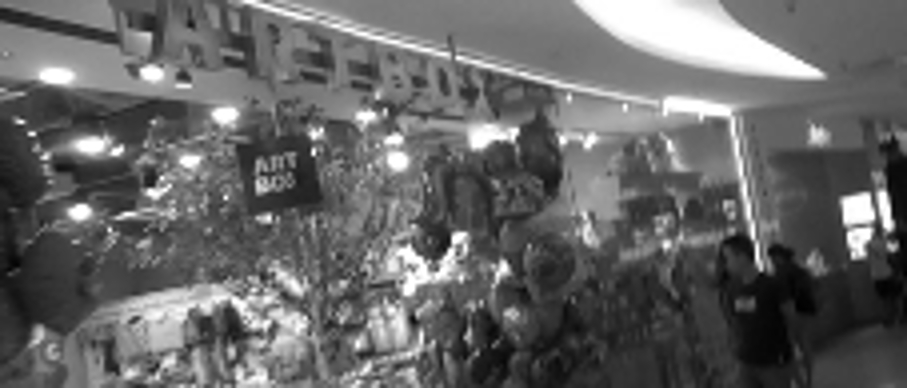}};
    \node [canvas is zy plane at x=-15] (Ip) {\includegraphics[width=0.75\linewidth]{method_015_IP2input}};
    \node [canvas is zy plane at x=-9] (I) {\includegraphics[width=\linewidth]{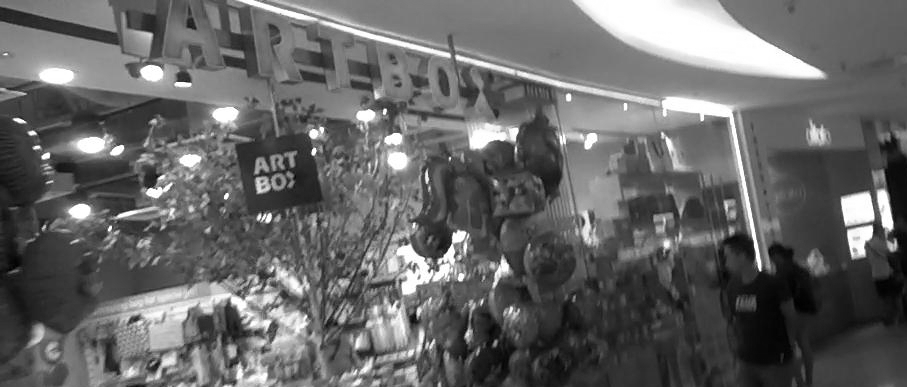}};
    \draw [dotted,thick] (I) -- (Ip);
    \draw [dotted,thick] (Ip) -- (Ip2);
    \node [canvas is zy plane at x=-6] (B) {\includegraphics[width=\linewidth]{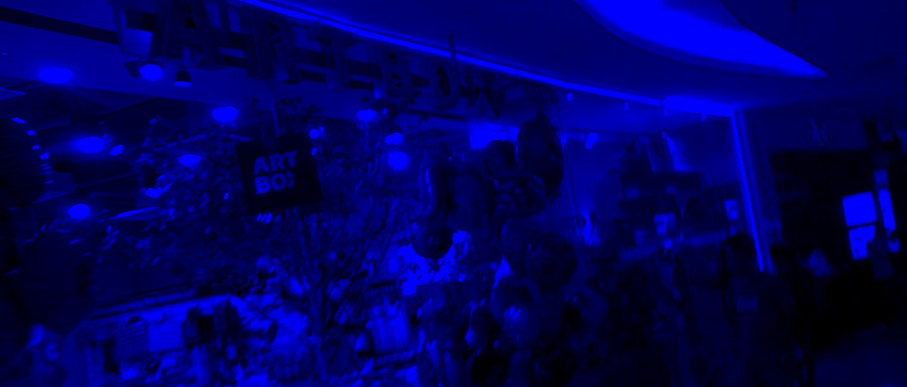}};
    \node [canvas is zy plane at x=-3] (G) {\includegraphics[width=\linewidth]{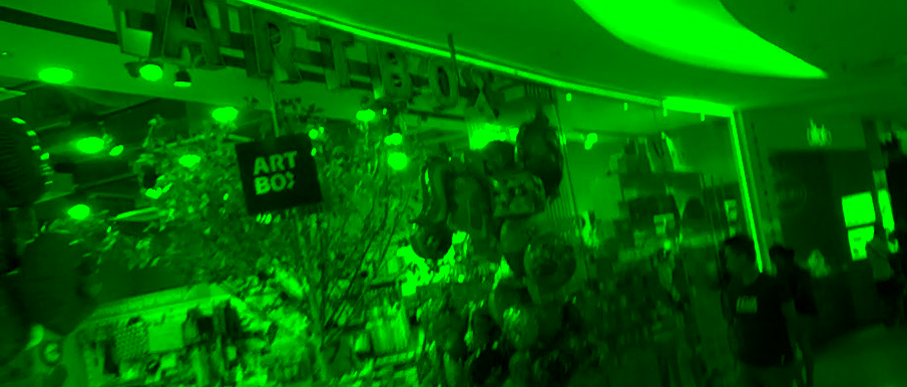}};
    \node [canvas is zy plane at x=0] (R) {\includegraphics[width=\linewidth]{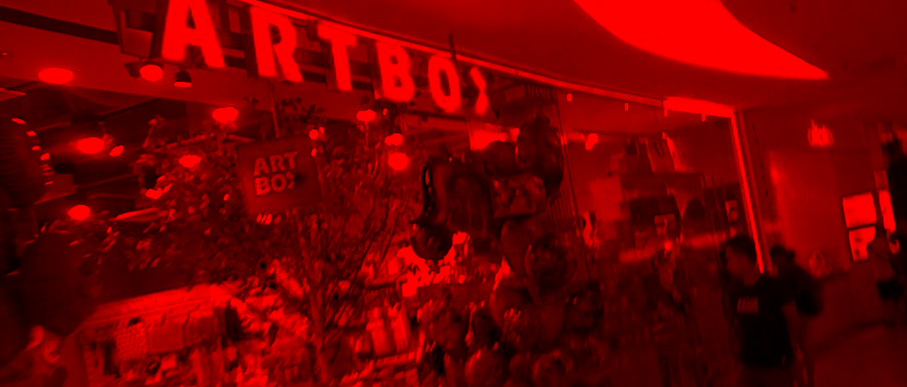}};
    \node [block, right=1.65cm of Ip2] (IP2mser) {\includegraphics[width=0.5\linewidth]{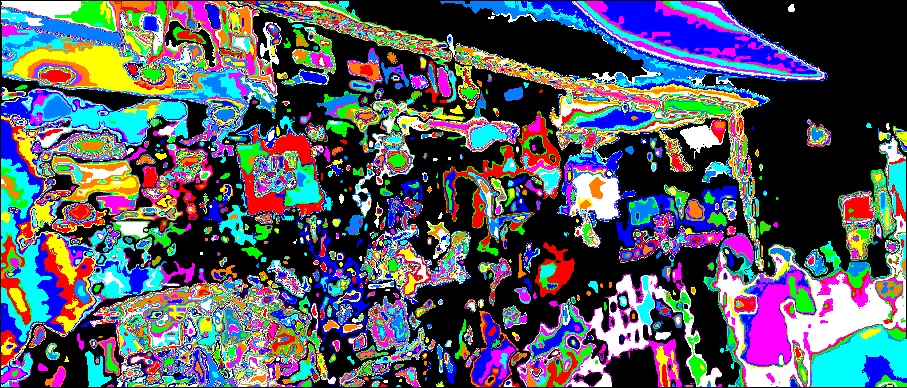}};
    \node [block, right=1.08cm of Ip] (IPmser) {\includegraphics[width=0.75\linewidth]{method_025_IP2mser}};
    \node [block, right=0.5cm of I] (Imser) {\includegraphics[width=\linewidth]{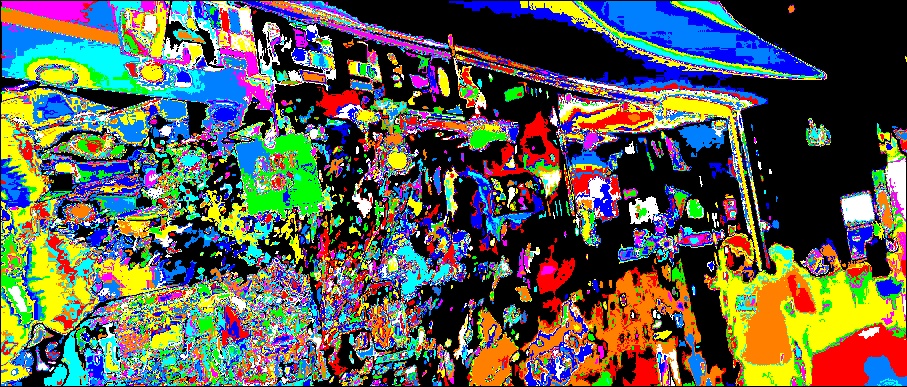}};
    \draw [dotted,thick] (Imser) -- (IPmser);
    \draw [dotted,thick] (IPmser) -- (IP2mser);
    \node [block, right=0.5cm of B] (Bmser) {\includegraphics[width=\linewidth]{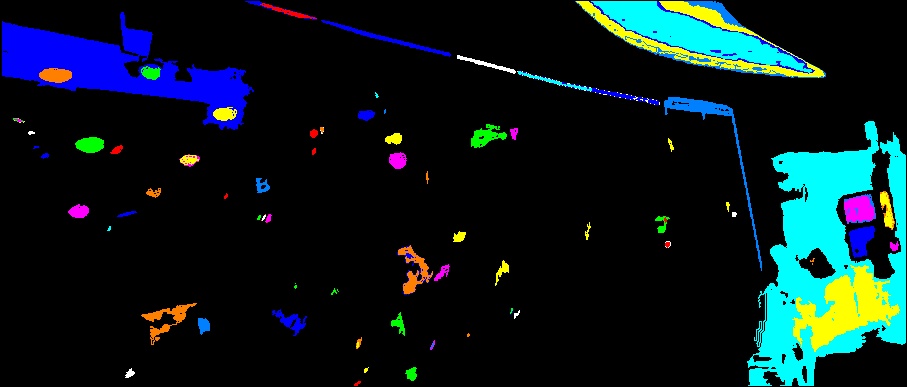}};
    \node [block, right=0.5cm of G] (Gmser) {\includegraphics[width=\linewidth]{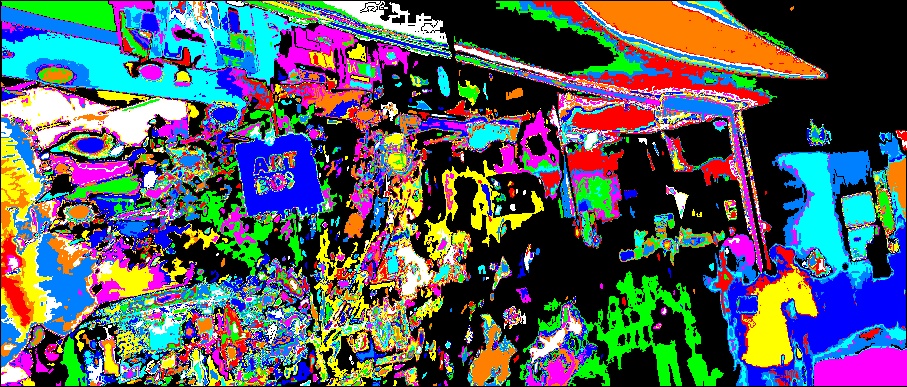}};
    \node [block, right=0.5cm of R] (Rmser) {\includegraphics[width=\linewidth]{method_021_Rmser}};
    \draw [-{latex}] (R) -- (Rmser);
    \draw [-{latex}] (G) -- (Gmser);
    \draw [-{latex}] (B) -- (Bmser);
    \draw [-{latex}] (I) -- (Imser);
    \draw [-{latex}] (Ip) -- (IPmser);
    \draw [-{latex}] (Ip2) -- (IP2mser);
    \node [block_D, right=1.0cm of IP2mser] (IP2D1) {\includegraphics[width=\linewidth]{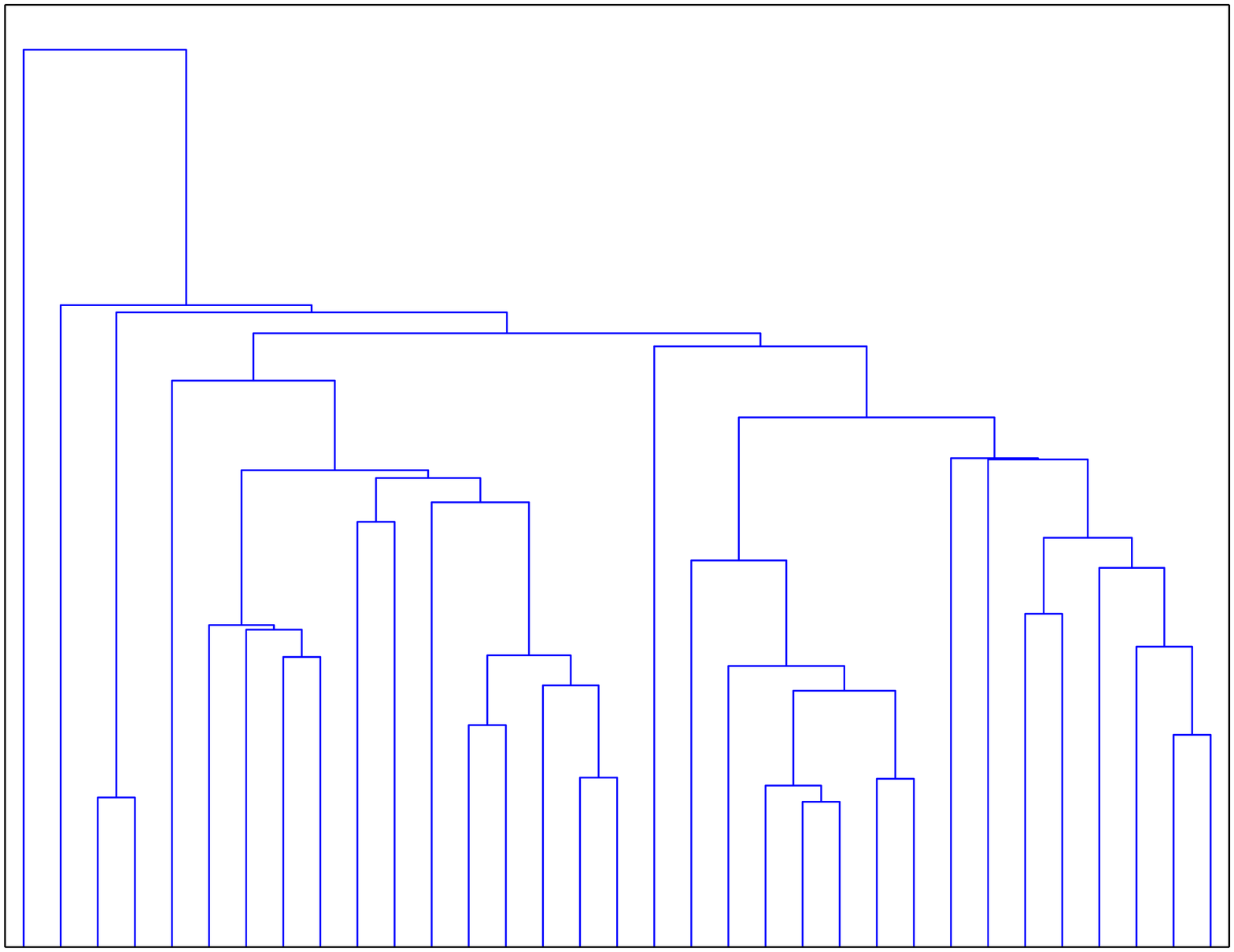}};
    \node [block_D, right=0.74cm of IPmser] (IPD1) {\includegraphics[width=\linewidth]{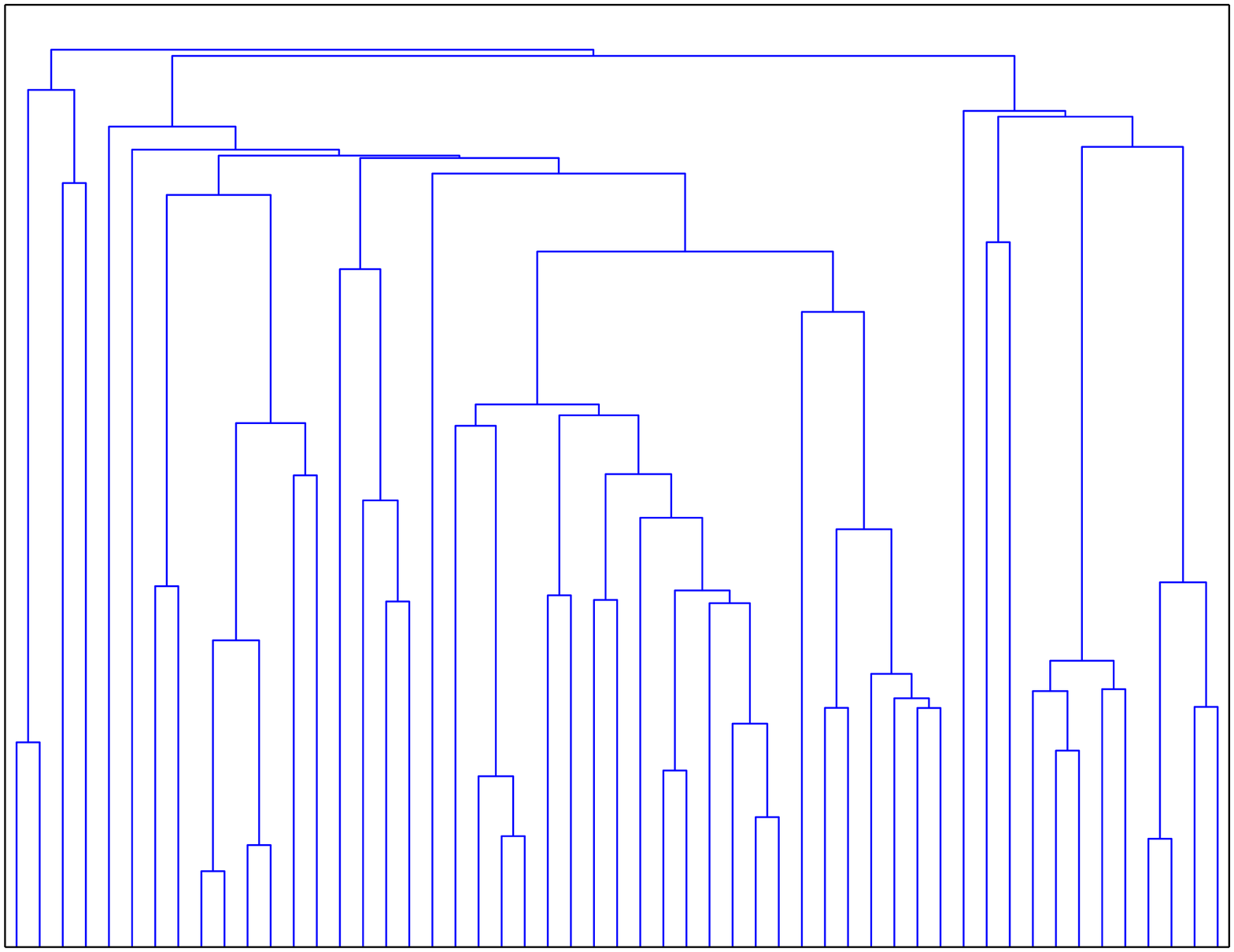}};
    \node [block_D, right=0.5cm of Imser] (ID1) {\includegraphics[width=\linewidth]{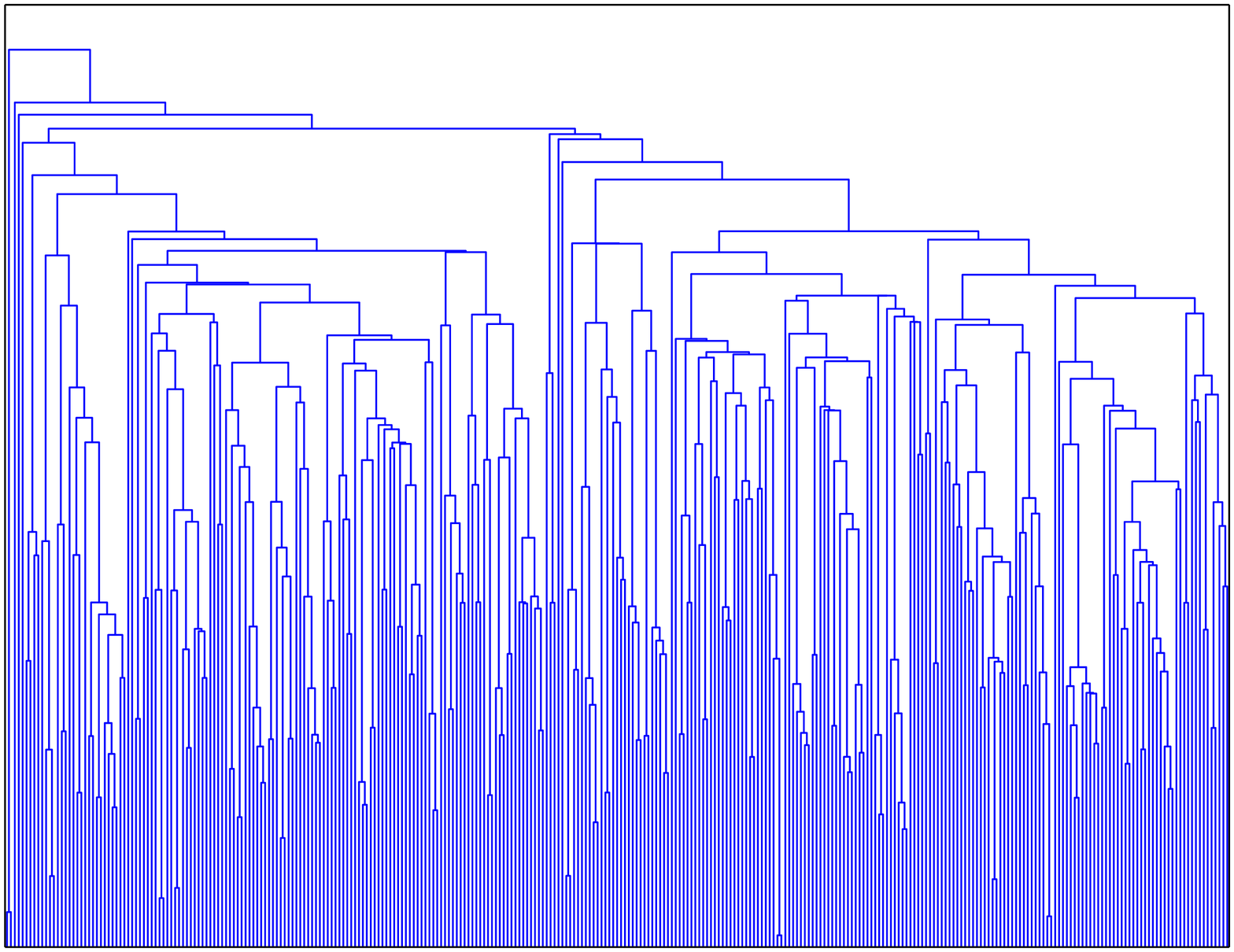}};
    \draw [dotted,thick] (ID1) -- (IPD1);
    \draw [dotted,thick] (IPD1) -- (IP2D1);
    \node [block_D, right=0.5cm of Bmser] (BD1) {\includegraphics[width=\linewidth]{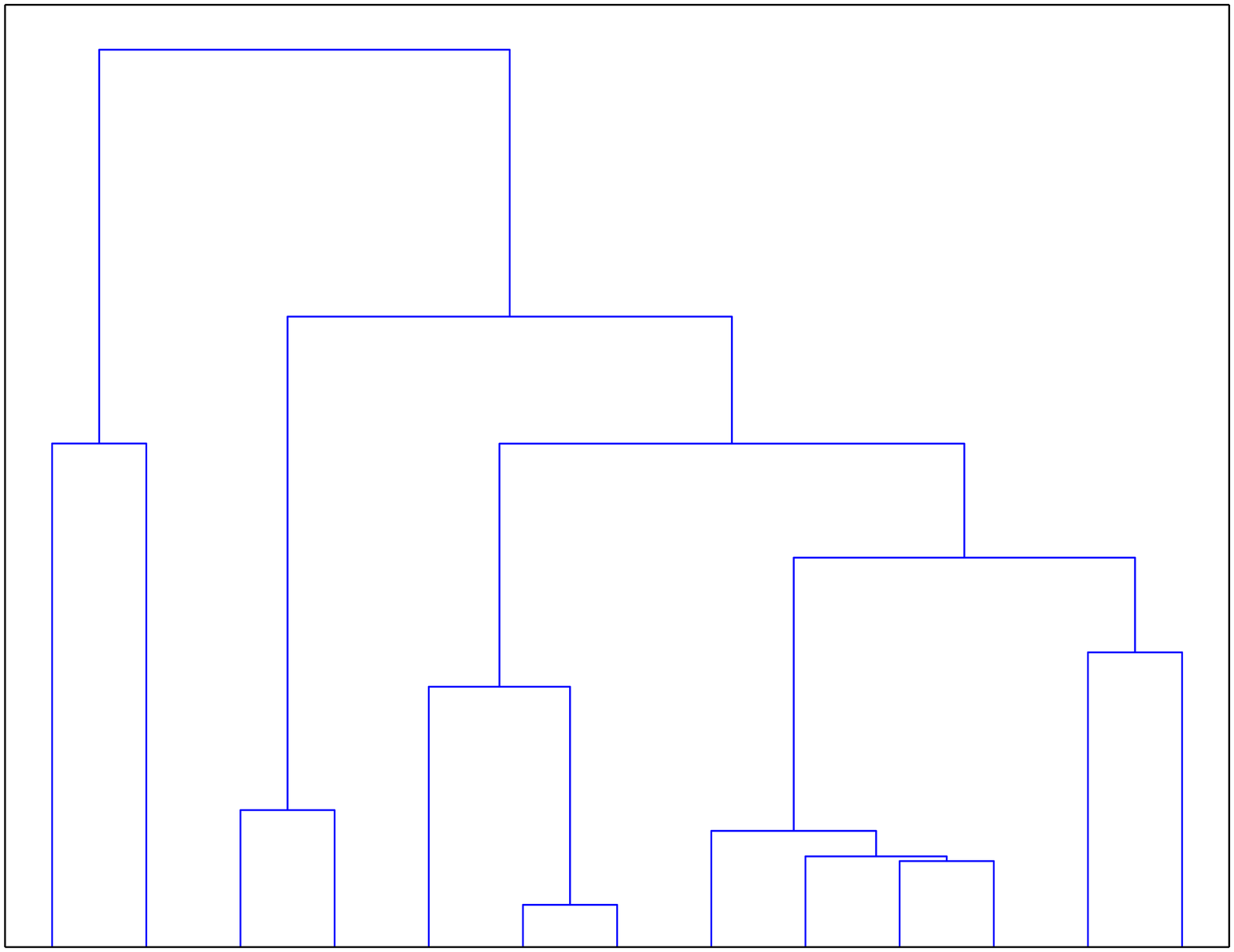}};
    \node [block_D, right=0.5cm of Gmser] (GD1) {\includegraphics[width=\linewidth]{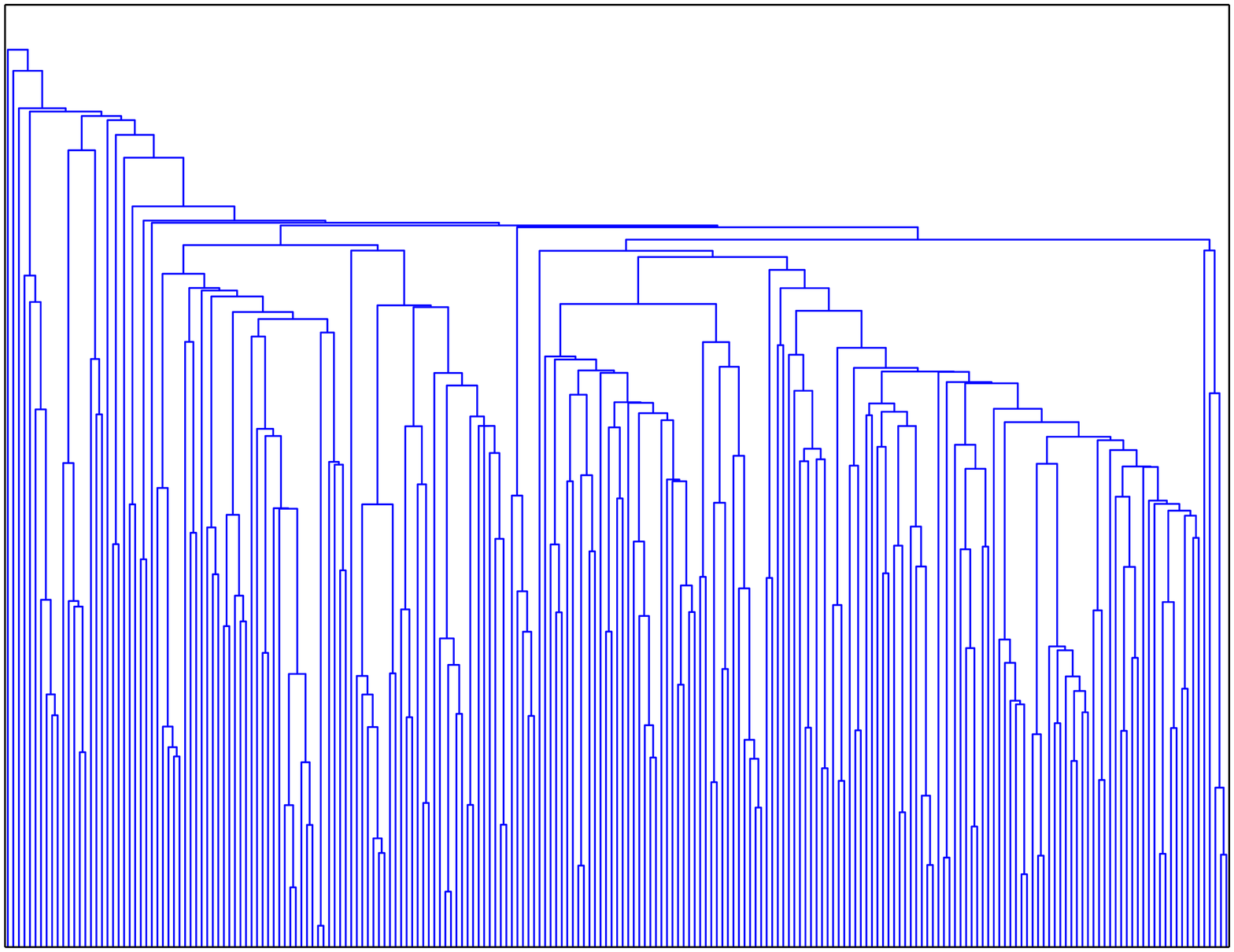}};
    \node [block_D, right=0.5cm of Rmser] (RD1) {\includegraphics[width=\linewidth]{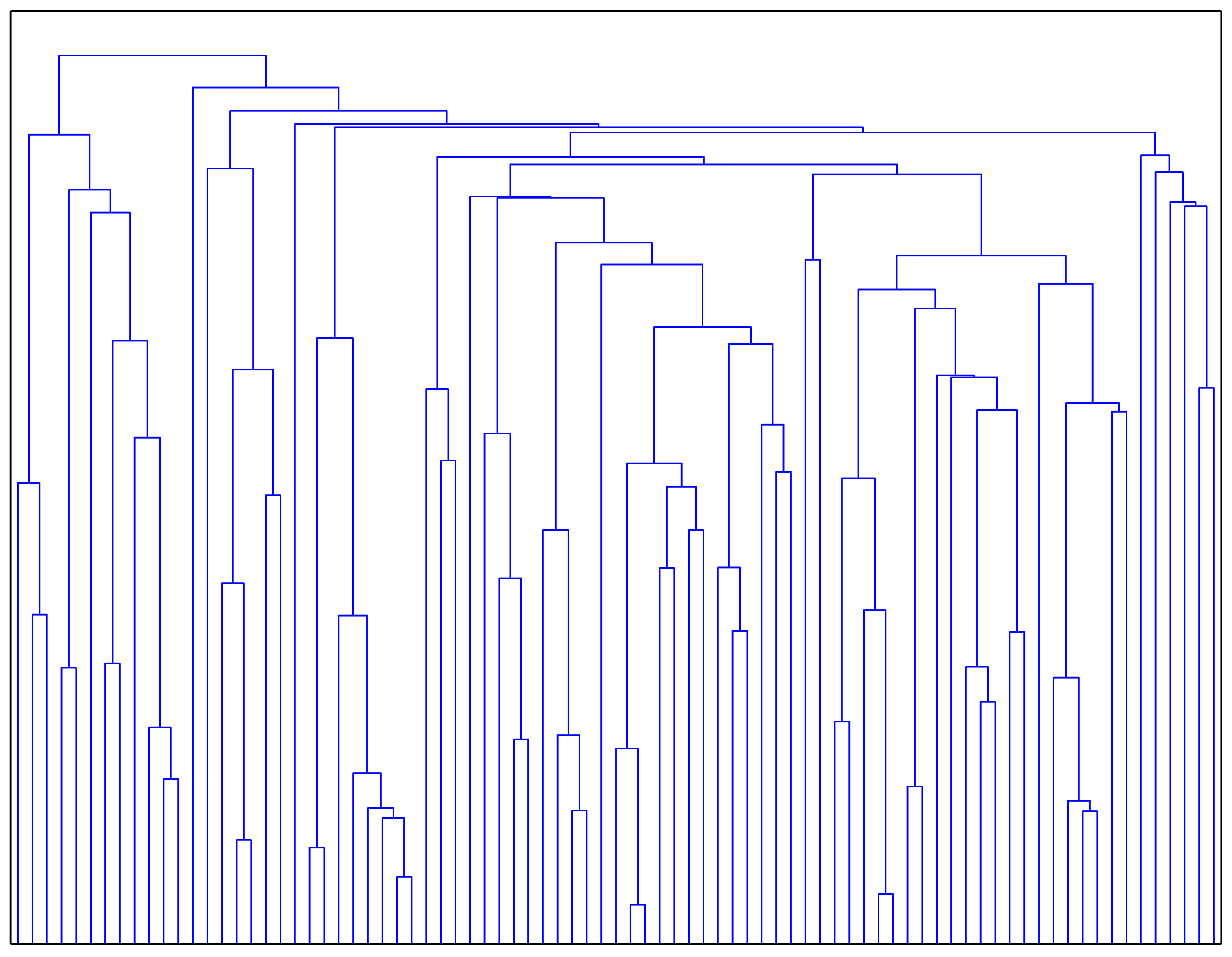}};
    \draw [-{latex}] (Rmser) -- (RD1);
    \draw [-{latex}] (Gmser) -- (GD1);
    \draw [-{latex}] (Bmser) -- (BD1);
    \draw [-{latex}] (Imser) -- (ID1);
    \draw [-{latex}] (IPmser) -- (IPD1);
    \draw [-{latex}] (IP2mser) -- (IP2D1);
    \node [block_D, right=0.05cm of IP2D1] (IP2D2) {\includegraphics[width=\linewidth]{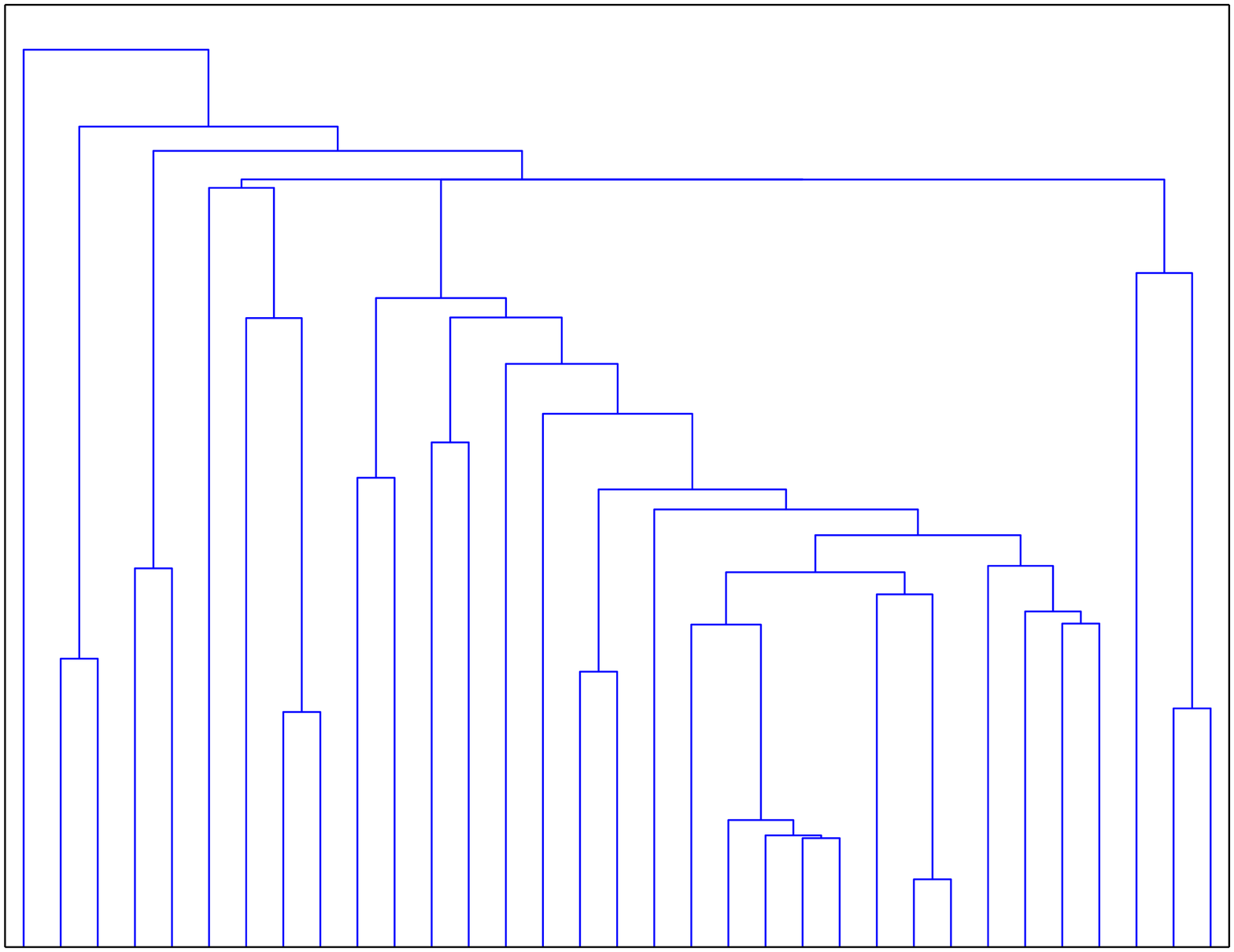}};
    \node [block_D, right=0.05cm of IPD1] (IPD2) {\includegraphics[width=\linewidth]{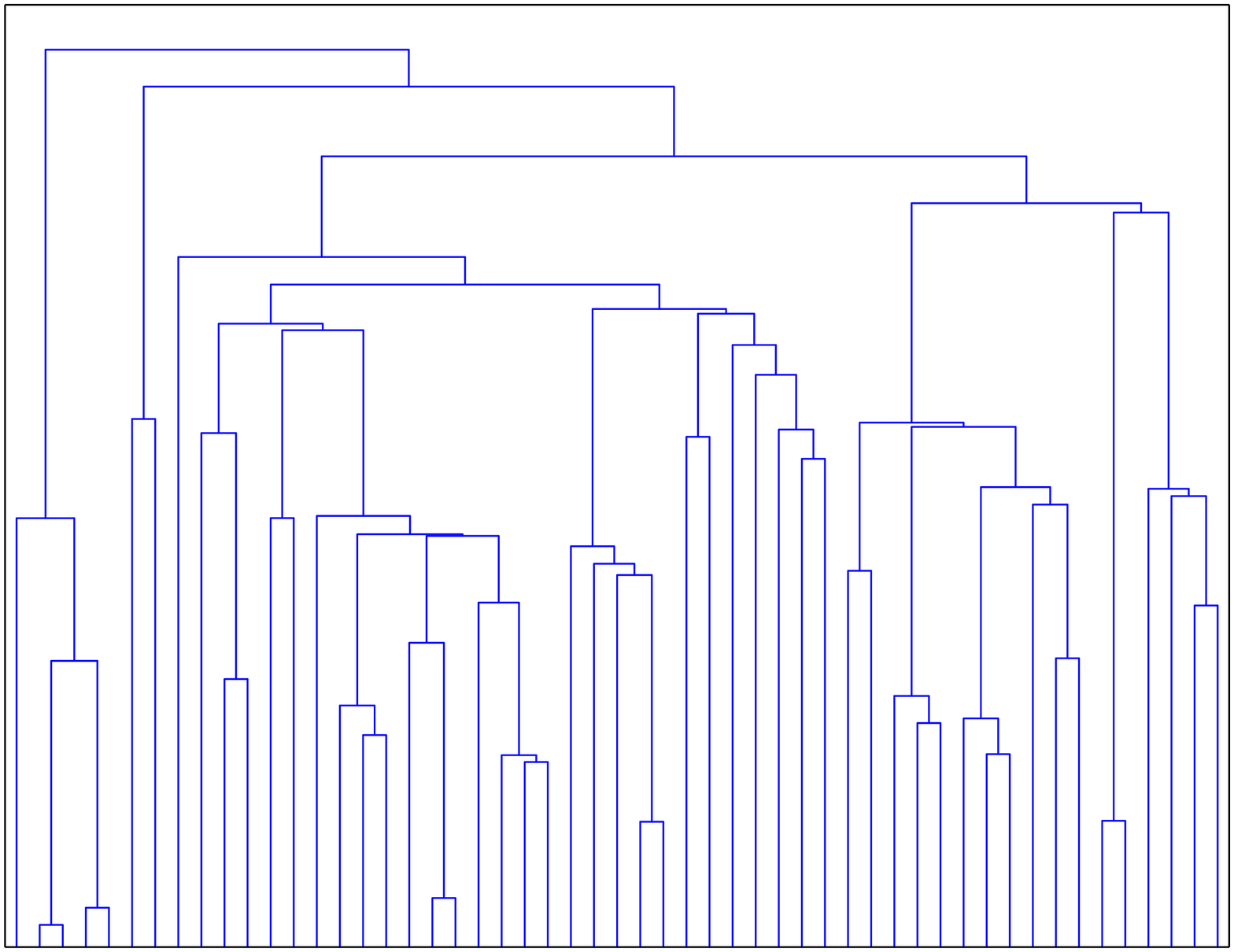}};
    \node [block_D, right=0.05cm of ID1] (ID2) {\includegraphics[width=\linewidth]{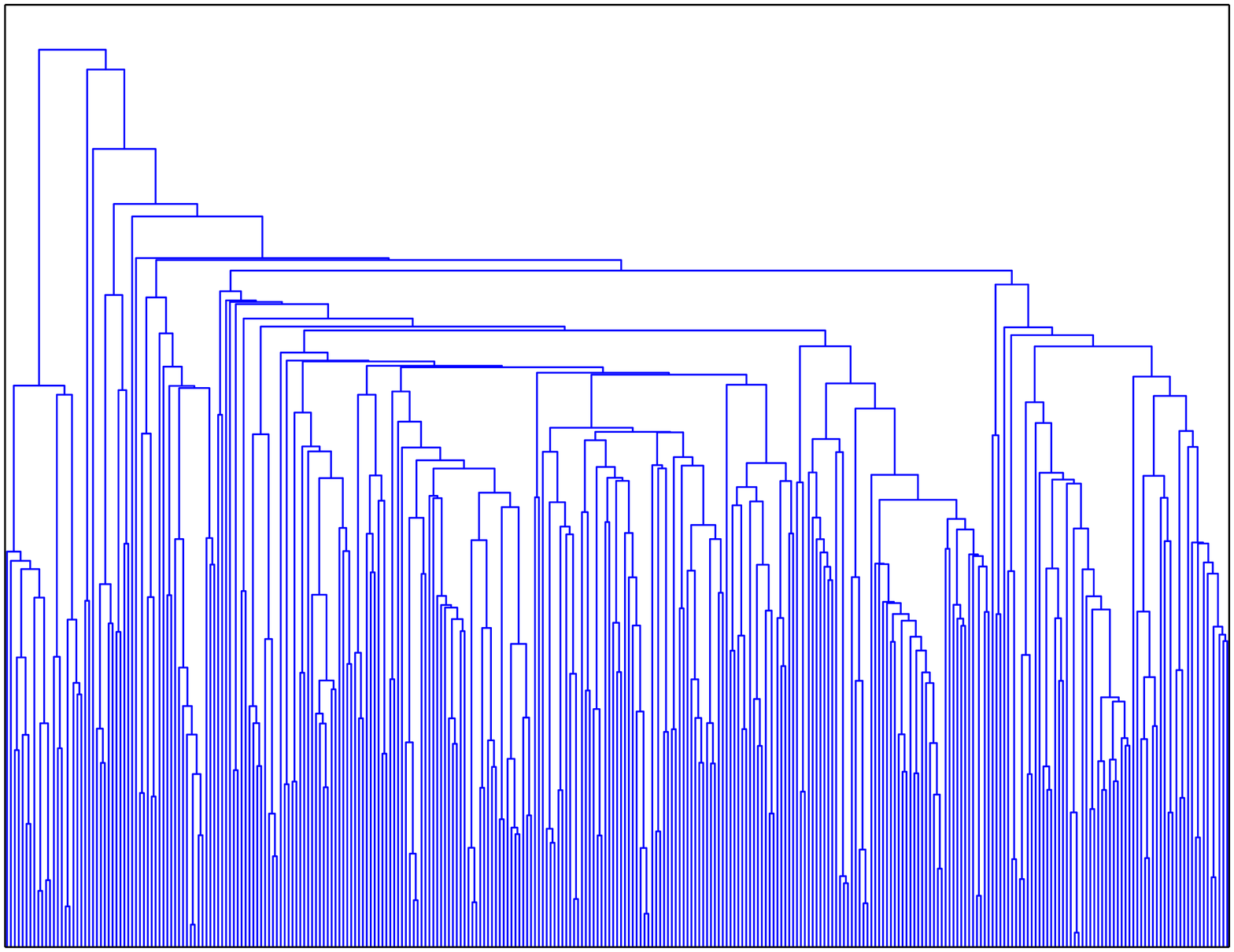}};
    \draw [dotted,thick] (ID2) -- (IPD2);
    \draw [dotted,thick] (IPD2) -- (IP2D2);
    \node [block_D, right=0.05cm of BD1] (BD2) {\includegraphics[width=\linewidth]{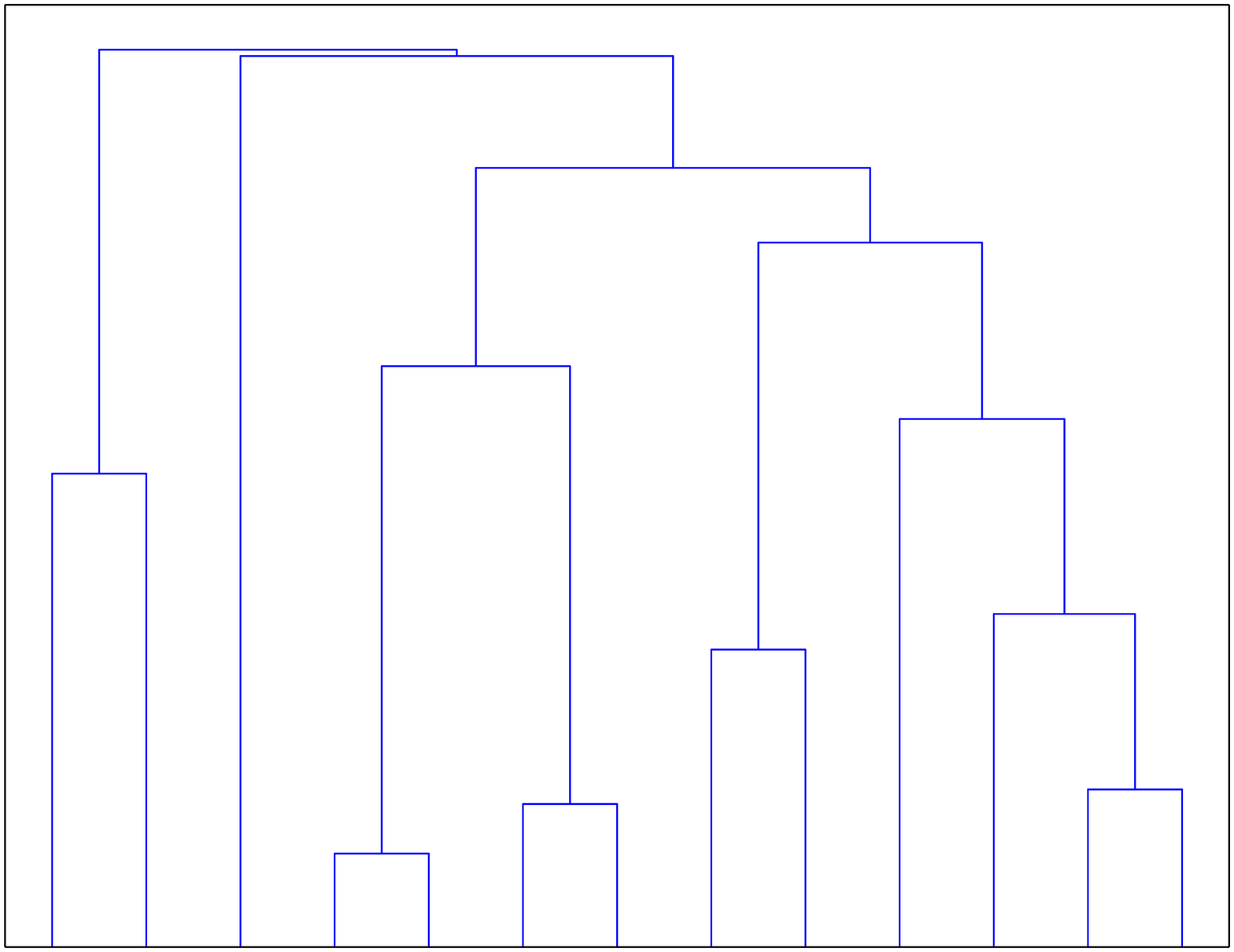}};
    \node [block_D, right=0.05cm of GD1] (GD2) {\includegraphics[width=\linewidth]{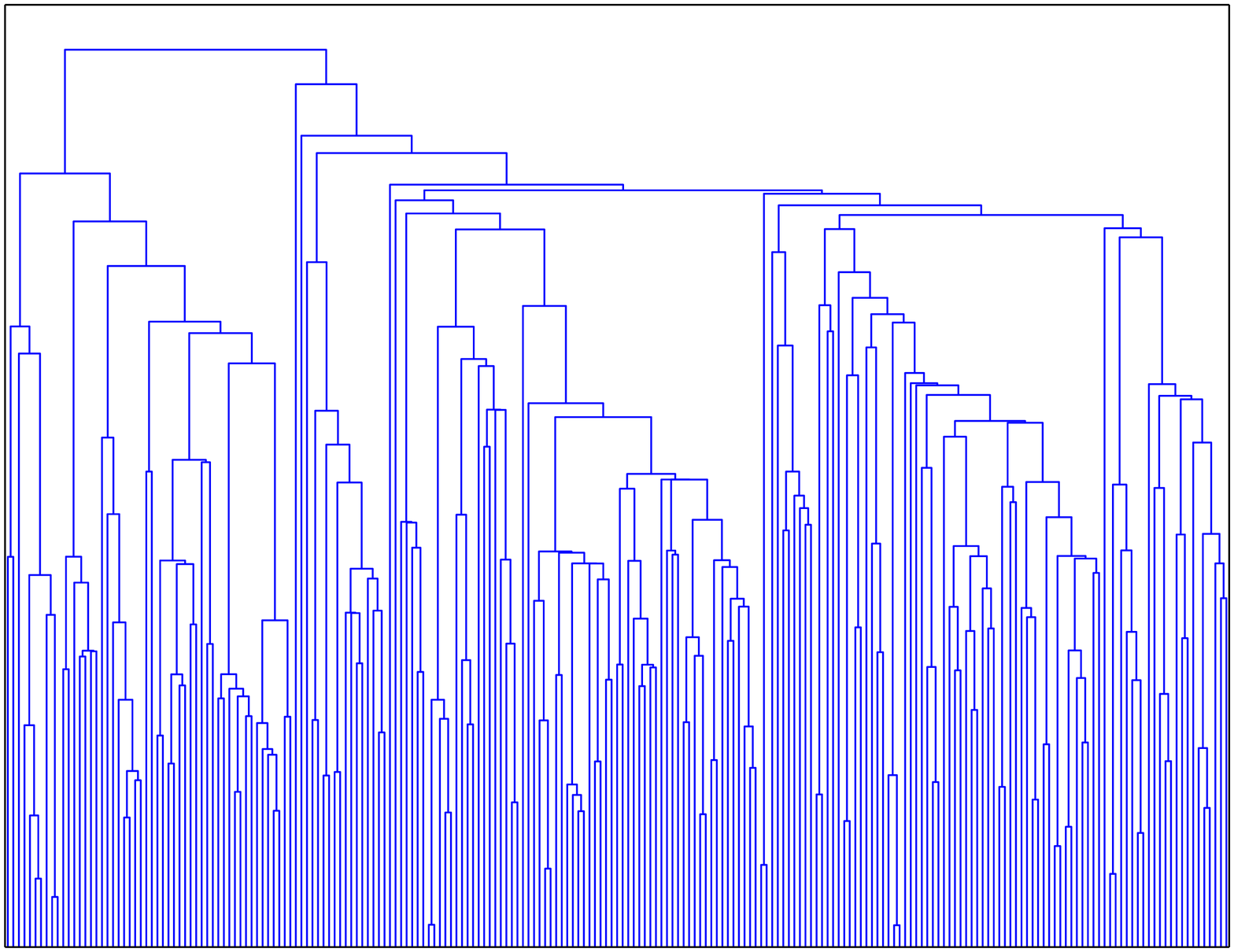}};
    \node [block_D, right=0.05cm of RD1] (RD2) {\includegraphics[width=\linewidth]{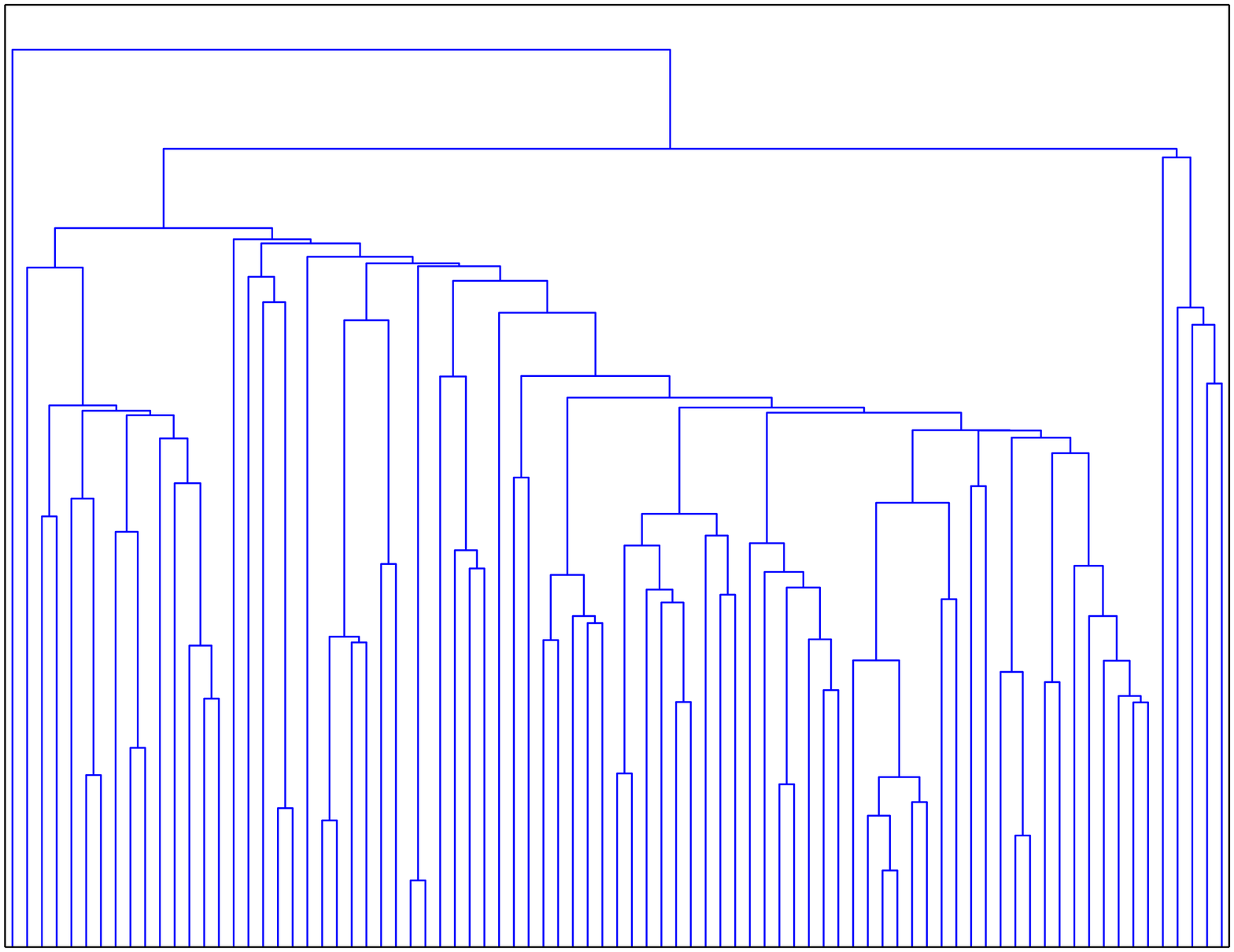}};
    \node [block_D, right=0.25cm of IP2D2] (IP2D3) {\includegraphics[width=\linewidth]{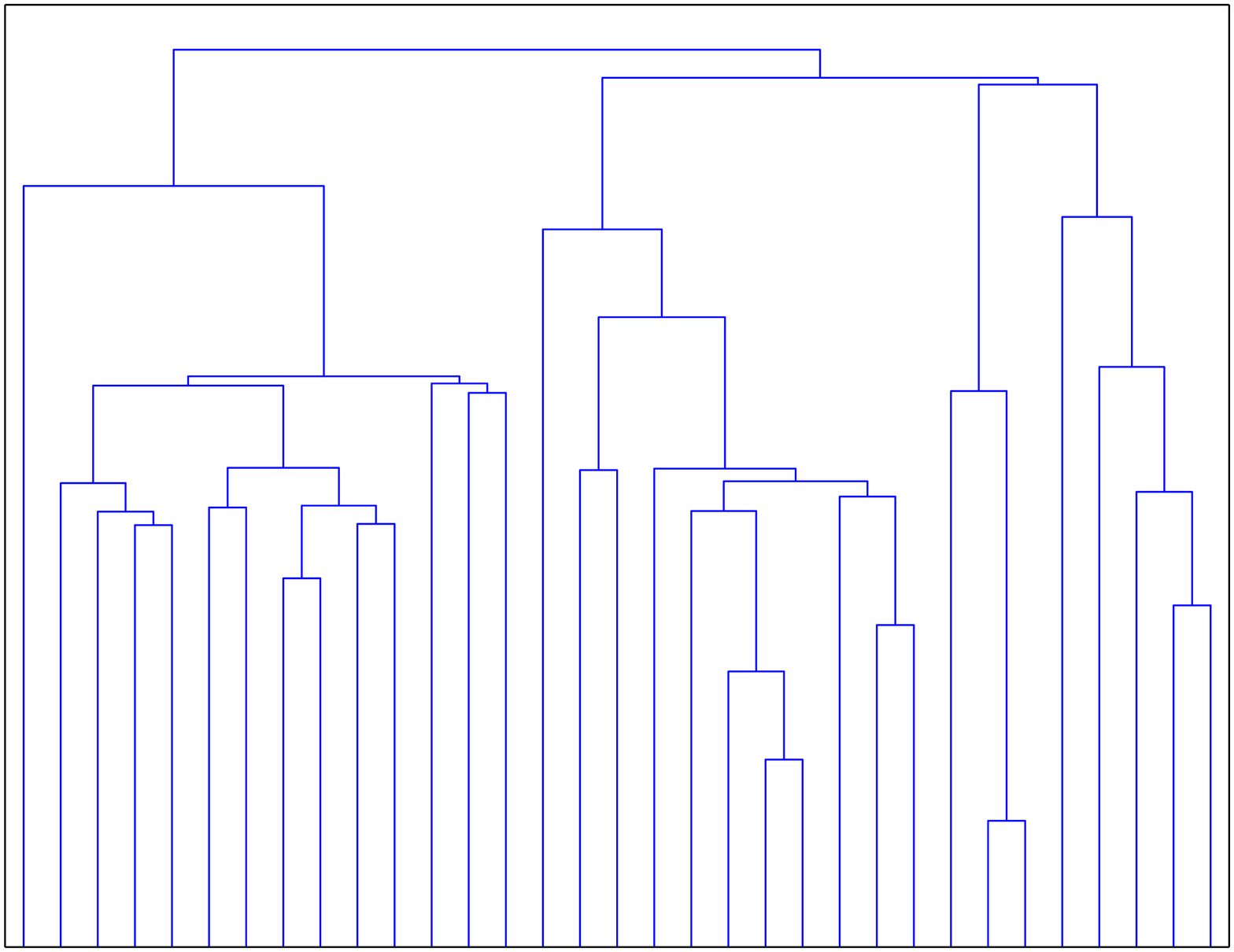}};
    \node [block_D, right=0.25cm of IPD2] (IPD3) {\includegraphics[width=\linewidth]{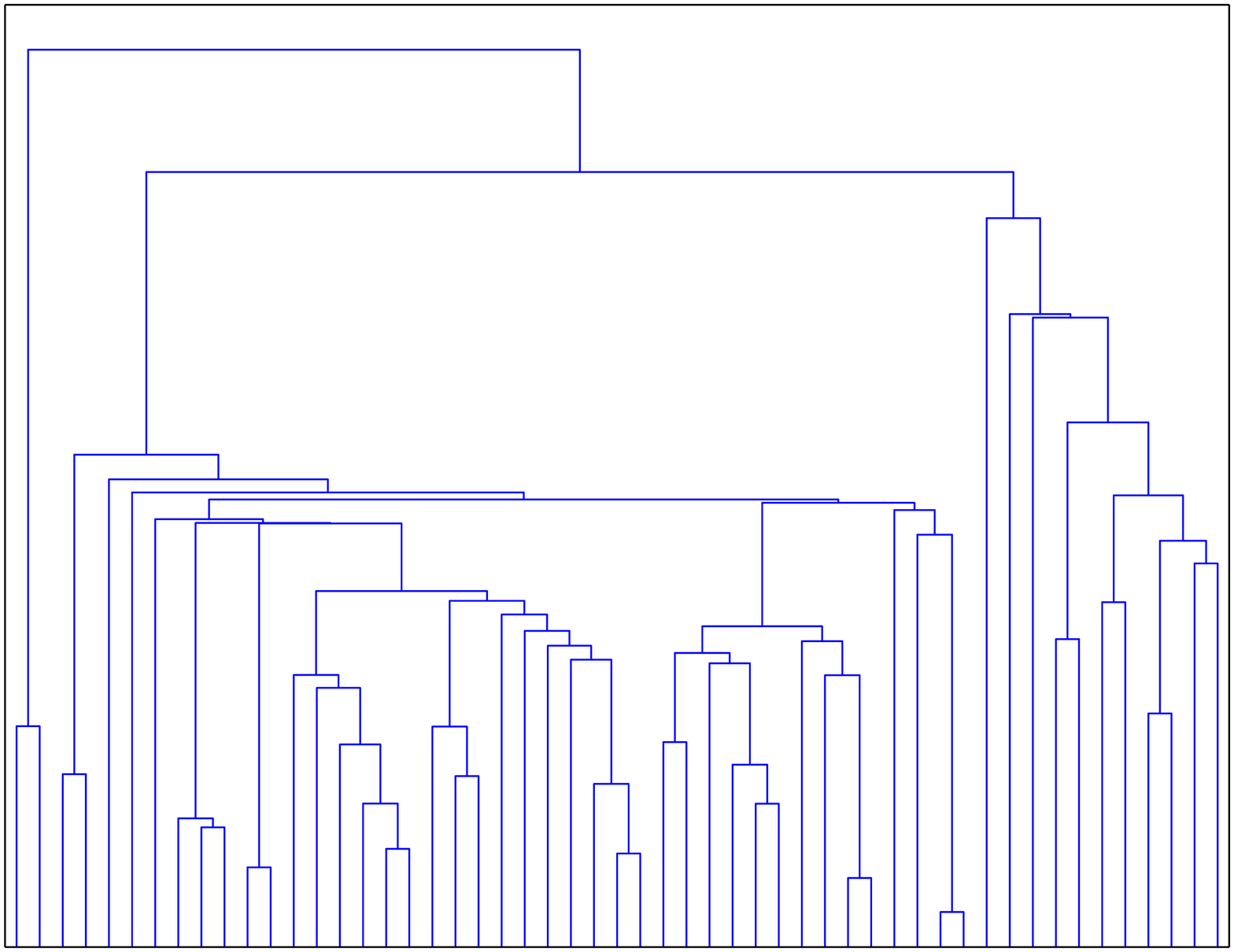}};
    \node [block_D, right=0.25cm of ID2] (ID3) {\includegraphics[width=\linewidth]{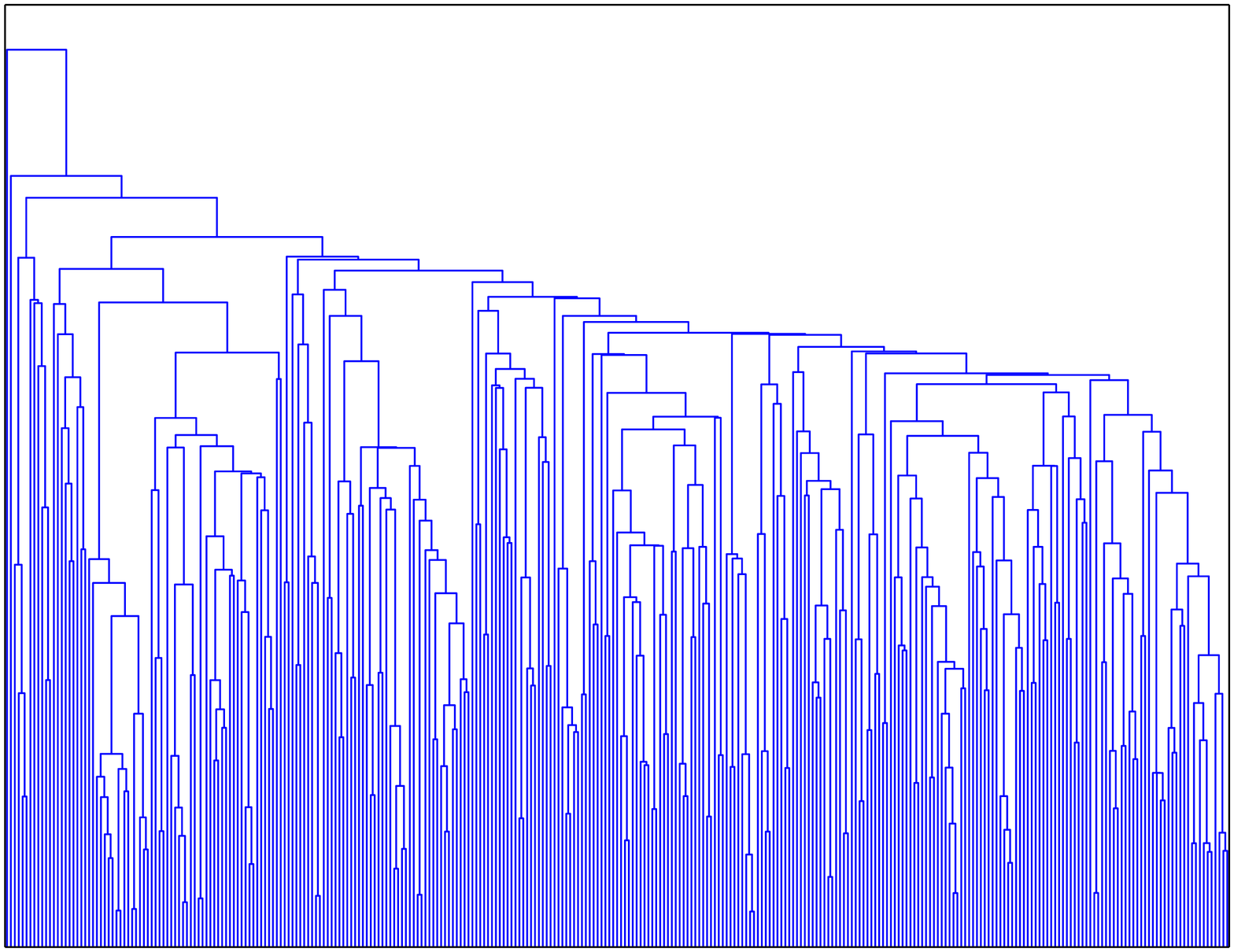}};
    \draw [dotted,thick] (ID3) -- (IPD3);
    \draw [dotted,thick] (IPD3) -- (IP2D3);
    \node [block_D, right=0.25cm of BD2] (BD3) {\includegraphics[width=\linewidth]{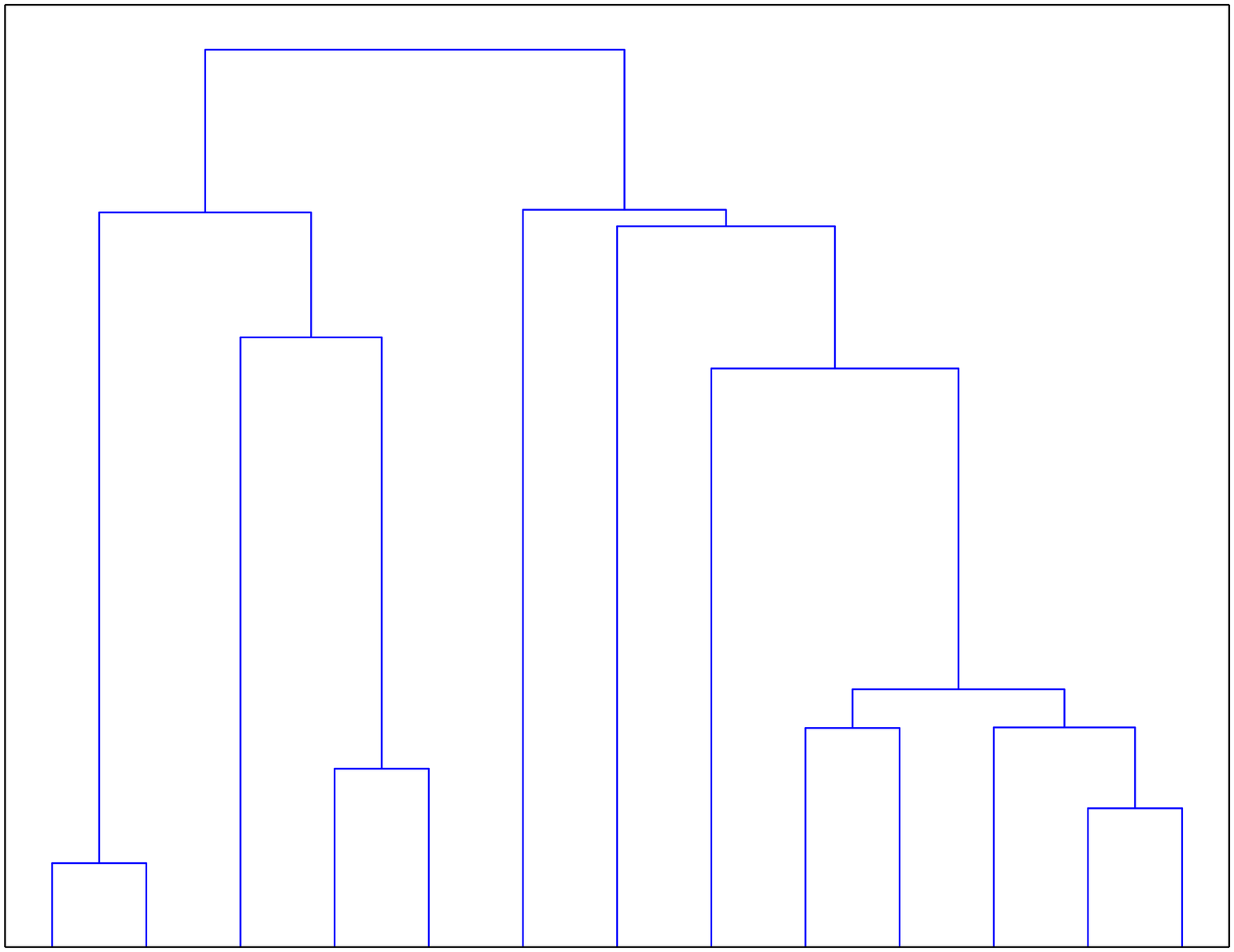}};
    \node [block_D, right=0.25cm of GD2] (GD3) {\includegraphics[width=\linewidth]{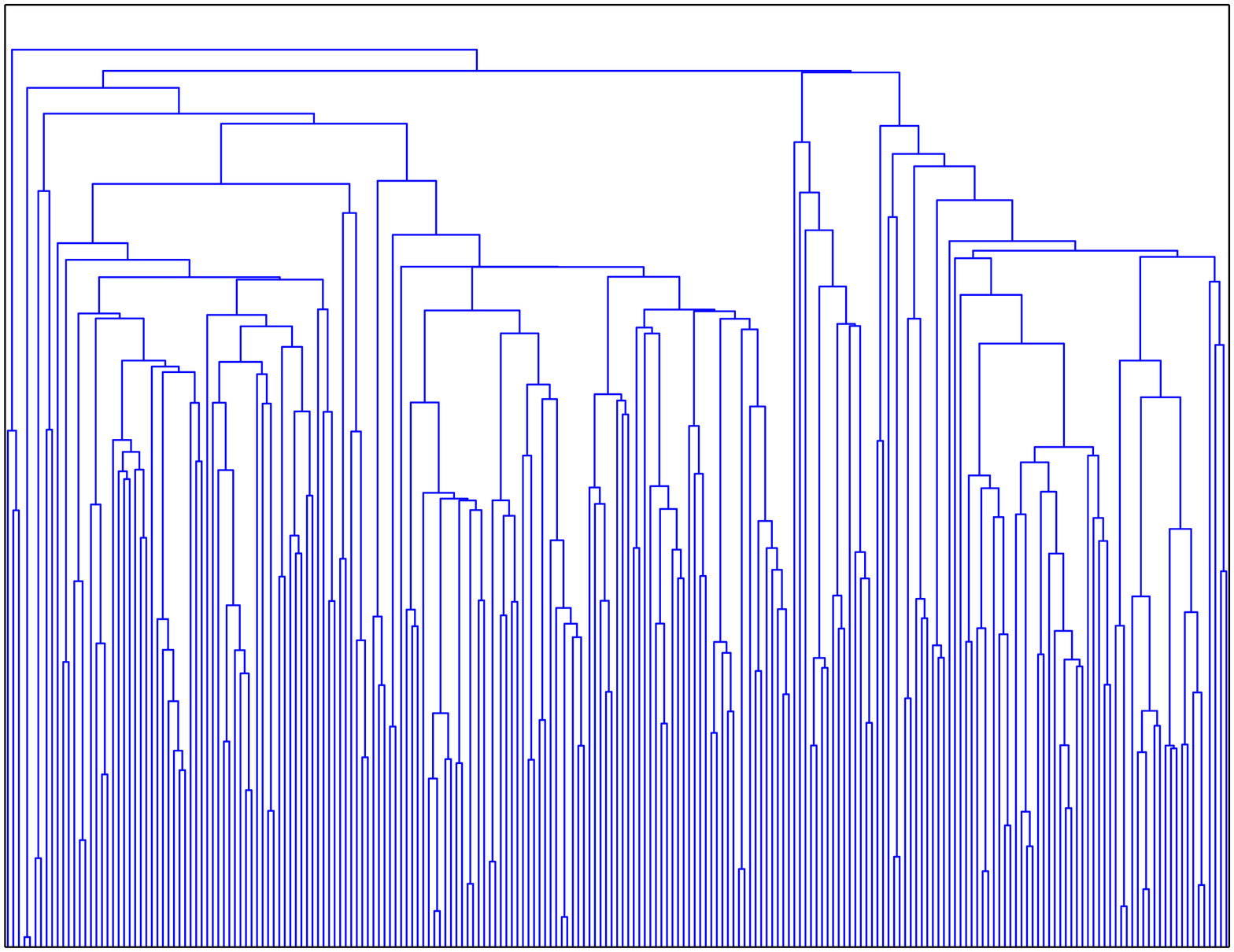}};
    \node [block_D, right=0.25cm of RD2] (RD3) {\includegraphics[width=\linewidth]{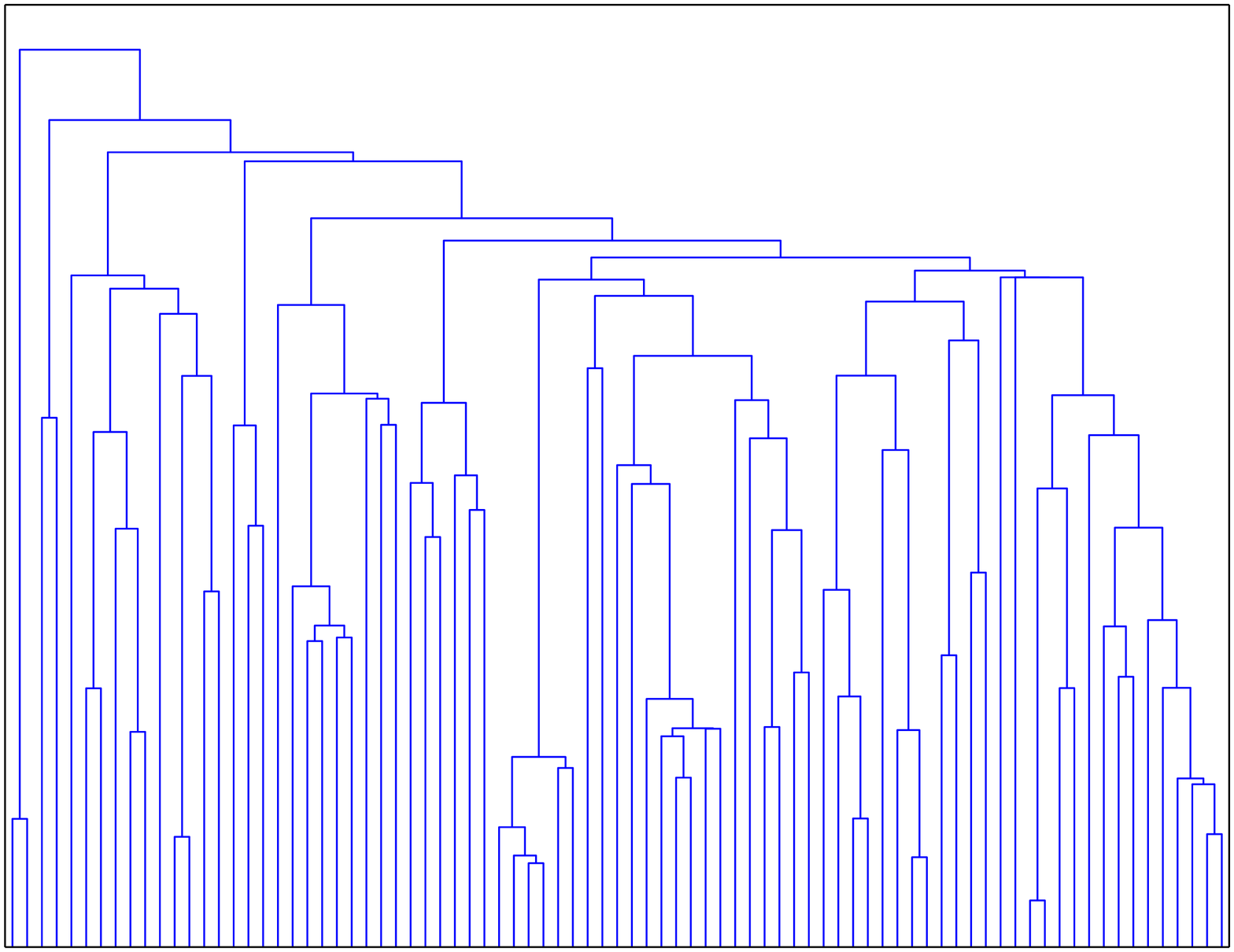}};
    \draw [dotted,thick] (RD2) -- (RD3);
    \draw [dotted,thick] (GD2) -- (GD3);
    \draw [dotted,thick] (BD2) -- (BD3);
    \draw [dotted,thick] (ID2) -- (ID3);
    \draw [dotted,thick] (IPD2) -- (IPD3);
    \draw [dotted,thick] (IP2D2) -- (IP2D3);

    \node [text width=2em,inner sep=0,outer sep=0] at (0,0,69) (w1) {\includegraphics[width=\linewidth]{method_041_word}};
    \node [text width=2em,inner sep=0,outer sep=0] at (-3.5,0,69) (w2) {\includegraphics[width=0.6\linewidth]{method_042_word}};
    \node [text width=2em,inner sep=0,outer sep=0] at (-7,0,69) (w3) {\includegraphics[width=0.7\linewidth]{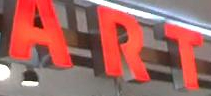}};
    \node [text width=2em,inner sep=0,outer sep=0] at (-10.5,0,69) (w4) {\includegraphics[width=0.6\linewidth]{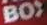}};
    \node [text width=2em,inner sep=0,outer sep=0] at (-19.2,0,69) (w5) {\includegraphics[width=\linewidth]{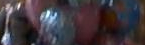}};
    \node [text width=2em,inner sep=0,outer sep=0] at (-22.5,0,69) (w6) {\includegraphics[width=0.6\linewidth]{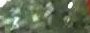}};
    \draw [dotted, ultra thick] (-13.5,0,65) -- (-16.2,0,65);
    \node [rotate=90,text width=6em,inner sep=0,outer sep=0] at (-9.5,0,-11) {\tiny{$N$ Input channels}};
    \node [text width=7em,inner sep=0,outer sep=0, above=0.25cm of RD2] {\tiny{~$M$ Similarity measures}};
    \node [rotate=90,text width=6em,inner sep=0,outer sep=0] at (-9.5,0,62) {\tiny{Ranked proposals}};
    \draw [thin,gray!50] (2,0,58) -- (-24,0,58);
    \draw [thick,gray!50] (2,0,58) -- (3,0,57);
    \draw [thick,gray!50] (-24,0,58) -- (-25,0,57);
    \draw [-{latex}] (-11,0,58) -- (-11,0,60);
\end{tikzpicture}
}
\caption{Diagrams of the ``segmentation-grouping'' (a) and TextProposals (b) algorithms.}
\label{fig:method_diagrams}
\end{figure*}

The TextProposals algorithm solves this problem and increases the overall detection recall of the basic ``segmentation and grouping'' by considering several input channels, and several complementary similarity measures. Moreover, our method includes an efficient ranking strategy that prioritizes the best word proposals found. Figure~\ref{fig:method_diagrams} presents the diagrams of the ``segmentation-grouping'' and TextProposals algorithms.

\begin{figure}
\centering
\subfloat[]{
\begin{minipage}[b][4.5cm][t]{.31\textwidth}
\includegraphics[width=\linewidth]{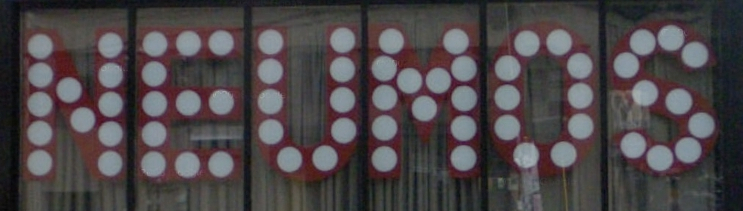}
\vfill
\includegraphics[width=\linewidth]{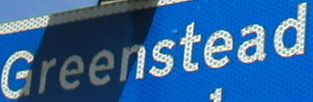}
\vfill
\includegraphics[width=\linewidth]{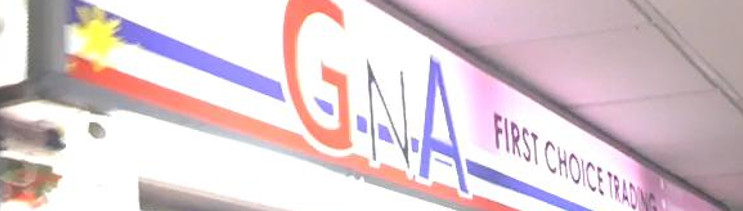}
\end{minipage}
}
\hfill
\subfloat[]{
\begin{minipage}[b][4.5cm][t]{.31\textwidth}
\includegraphics[width=\linewidth]{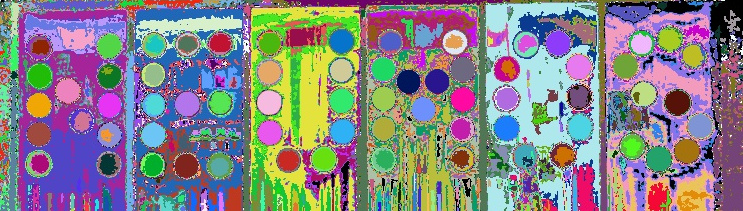}
\vfill
\includegraphics[width=\linewidth]{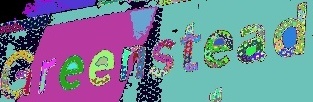}
\vfill
\includegraphics[width=\linewidth]{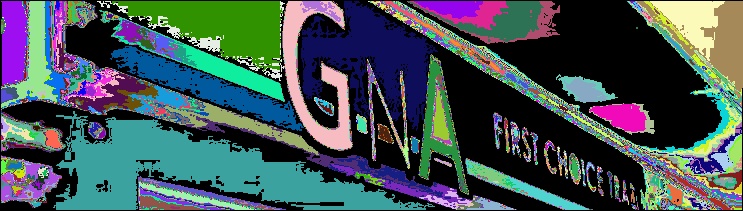}
\end{minipage}
}
\hfill
\subfloat[]{
\begin{minipage}[b][4.5cm][t]{.31\textwidth}
\includegraphics[width=\linewidth]{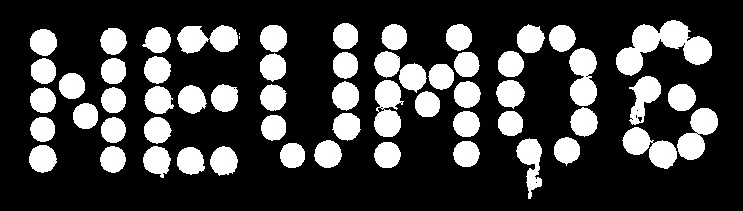}
\vfill
\includegraphics[width=\linewidth]{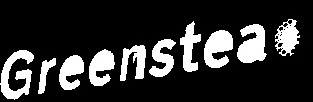}
\vfill
\includegraphics[width=\linewidth]{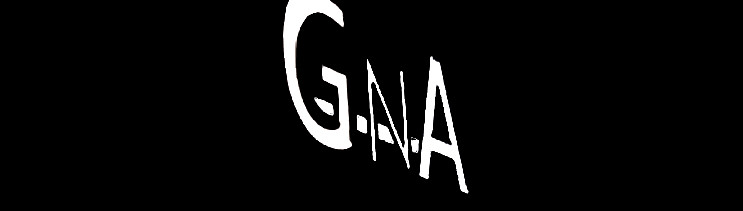}
\end{minipage}
}

\caption{Detailed steps of our method giving rise to good proposals for three different scene text words. From left to right: the detail of a single word from the original input image, the initial over-segmentation (each region in a different color), and the group of regions that generates the word proposal with better Intersection over Union with its corresponding ground truth bounding box.}
\label{fig:mser_grouping}
\end{figure}

As illustrated in Figure~\ref{fig:mser_grouping} our method is able to produce good quality word proposals in different cases for which a single best segmentation strategy and similarity metric does not exists. From left to right we show: the detail of a single word from the original input image, the initial over-segmentation (each connected component in a different color), and the group of connected components that generates the word proposal with better Intersection over Union with the ground truth word bounding box. In the case of the word in the top row, the best proposal has been generated by applying the Single Linkage Clustering (SLC) algorithm using the Euclidean distance in a three-dimensional space defined by the average intensity value of the connected components and the $x,y$ coordinates of their centers. In the other rows the same strategy does not produce a good word proposal, either because the shadowing effect breaks the color similarity of text regions or because they have different colors by design. Instead the best proposal is generated respectively in the spaces defined by the diameter of the regions and the $x,y$ coordinates of their centers (middle row), or by the mean background color and the $x,y$ coordinates of their centers (bottom row).

These examples make clear that our grouping can not rely in a single similarity measure of text-parts, as due to design, scene layout, and environment effects different similarity cues might be active in each case. As a result a flexible approach is proposed, where various weak similarity cues are explored independently in parallel. It is also important to notice how in these examples we make use of many overlapping regions (connected components) that are not filtered in any way by their shapes or any other attribute. Also notice how the best grouping in the top row example would be rejected by any discriminative rule/classifier based on regions collinearity. Contrary to traditional text detection methods in here we avoid the use of such filters.

\subsection{Region decomposition}
\label{sec:mser}

The first step in our method is the initial pixel-level segmentation where the atomic parts, that will give rise to text groupings, are identified. For this we make use of the Maximally Stable Extremal Regions (MSER) algorithm~\cite{Matas2004} as in many existing text detection methods. However, since the regions that are of our interest are not only well-segmented characters, we can relax the parametrization of the MSER algorithm in order to produce a richer over-segmentation with many overlapping regions. The obtained regions are not filtered in any way.

In fact, the proposed method is not strictly dependent on the MSER algorithm and would be able to produce similar results with any other over-segmentation technique as far it is able to extract small level sets (connected components) corresponding to text parts. Thus, the use of the MSER algorithm here can be seen as a way to optimize the whole object proposals method by reducing the number of regions to analyze compared with using the whole component tree of the image.

\subsection{Group Hypothesis Creation}
\label{sec:grouping}

The grouping process starts with a set of regions $\mathcal{R}_c$ extracted with the MSER algorithm. Initially each region $r\in\mathcal{R}_c$ starts in its own cluster and then the closest pair of clusters ($A,B$) is merged iteratively, using the single linkage criterion (SLC) ($ \min \, \{\, \mathrm{d}(r_a,r_b) : r_a \in A,\, r_b \in B \,\} $), until all regions are clustered together ($C \equiv \mathcal{R}_c$). 

For defining the distance metric $\mathrm{d}(r_a,r_b)$ in order to describe similarity relations between text-parts of a higher-level text grouping (e.g. words) we aim to use a set of complementary features with low computational cost. In here we use a set of seven weak similarity cues for which the corresponding region features can be easily computed in a sequence as illustrated in Figure~\ref{fig:features}. The list of features is as follows:

\begin{figure}
\centering
\subfloat[]{\includegraphics[width=0.31\linewidth]{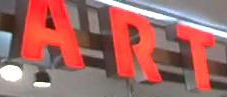}}
\hfill
\subfloat[]{\includegraphics[width=0.31\linewidth]{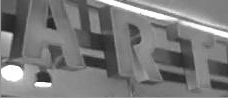}}
\hfill
\subfloat[]{\includegraphics[width=0.31\linewidth]{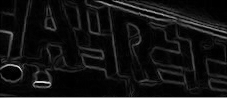}}
\\
\subfloat[]{
\label{fig:features_d}
\begin{minipage}[b][7.3cm][t]{.12\textwidth}
\begin{tikzpicture}[node distance = 0.1cm, auto]
\node[minimum width=\linewidth, inner sep=0,outer sep=0] (F1) at (0,0) {\includegraphics[width=\linewidth]{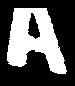}};
\node[text=black, rectangle, minimum width=\linewidth,minimum height=1em,below=0.0003cm of F1, inner sep=2.5,outer sep=0] {};
\end{tikzpicture}
\vfill
\begin{tikzpicture}[node distance = 0.1cm, auto]
\node[minimum width=\linewidth, inner sep=0,outer sep=0] (F1) at (0,0) {\includegraphics[width=\linewidth]{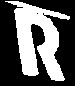}};
\node[text=black, rectangle, minimum width=\linewidth,minimum height=1em,below=0.0003cm of F1, inner sep=2.5,outer sep=0] {};
\end{tikzpicture}
\vfill
\begin{tikzpicture}[node distance = 0.1cm, auto]
\node[minimum width=\linewidth, inner sep=0,outer sep=0] (F1) at (0,0) {\includegraphics[width=\linewidth]{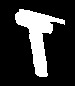}};
\node[text=black, rectangle, minimum width=\linewidth,minimum height=1em,below=0.0003cm of F1, inner sep=2.5,outer sep=0] {};
\end{tikzpicture}
\end{minipage}
}
\hfill
\subfloat[]{
\label{fig:features_e}
\begin{minipage}[b][7.3cm][t]{.12\textwidth}
\begin{tikzpicture}[node distance = 0.1cm, auto]
\node[minimum width=\linewidth, inner sep=0,outer sep=0] (F1) at (0,0) {\includegraphics[width=\linewidth]{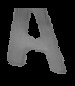}};
\node[draw=black!80,fill=gray!50, text=black, rectangle, minimum width=\linewidth,minimum height=1em,below=0.0003cm of F1, inner sep=2.5,outer sep=0] { \scriptsize{$\mu_I=122$}};
\end{tikzpicture}
\vfill
\begin{tikzpicture}[node distance = 0.1cm, auto]
\node[minimum width=\linewidth, inner sep=0,outer sep=0] (F1) at (0,0) {\includegraphics[width=\linewidth]{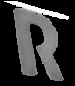}};
\node[draw=black!80,fill=gray!50, text=black, rectangle, minimum width=\linewidth,minimum height=1em,below=0.0003cm of F1, inner sep=2.5,outer sep=0] { \scriptsize{$\mu_I=145$}};
\end{tikzpicture}
\vfill
\begin{tikzpicture}[node distance = 0.1cm, auto]
\node[minimum width=\linewidth, inner sep=0,outer sep=0] (F1) at (0,0) {\includegraphics[width=\linewidth]{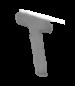}};
\node[draw=black!80,fill=gray!50, text=black, rectangle, minimum width=\linewidth,minimum height=1em,below=0.0003cm of F1, inner sep=2.5,outer sep=0] { \scriptsize{$\mu_I=151$}};
\end{tikzpicture}
\end{minipage}
}
\hfill
\subfloat[]{
\label{fig:features_f}
\begin{minipage}[b][7.3cm][t]{.12\textwidth}
\begin{tikzpicture}[node distance = 0.1cm, auto]
\node[minimum width=\linewidth, inner sep=0,outer sep=0] (F1) at (0,0) {\includegraphics[width=\linewidth]{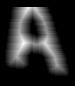}};
\node[draw=black!80,fill=gray!50, text=black, rectangle, minimum width=\linewidth,minimum height=1em,below=0.0003cm of F1, inner sep=2.5,outer sep=0] { \scriptsize{$\mu_{S}=18.0$}};
\end{tikzpicture}
\vfill
\begin{tikzpicture}[node distance = 0.1cm, auto]
\node[minimum width=\linewidth, inner sep=0,outer sep=0] (F1) at (0,0) {\includegraphics[width=\linewidth]{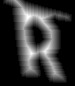}};
\node[draw=black!80,fill=gray!50, text=black, rectangle, minimum width=\linewidth,minimum height=1em,below=0.0003cm of F1, inner sep=2.5,outer sep=0] { \scriptsize{$\mu_{S}=16.5$}};
\end{tikzpicture}
\vfill
\begin{tikzpicture}[node distance = 0.1cm, auto]
\node[minimum width=\linewidth, inner sep=0,outer sep=0] (F1) at (0,0) {\includegraphics[width=\linewidth]{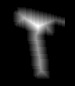}};
\node[draw=black!80,fill=gray!50, text=black, rectangle, minimum width=\linewidth,minimum height=1em,below=0.0003cm of F1, inner sep=2.5,outer sep=0] { \scriptsize{$\mu_{S}=15.1$}};
\end{tikzpicture}
\end{minipage}
}
\hfill
\subfloat[]{
\label{fig:features_g}
\begin{minipage}[b][7.3cm][t]{.12\textwidth}
\begin{tikzpicture}[node distance = 0.1cm, auto]
\node[minimum width=\linewidth, inner sep=0,outer sep=0] (F1) at (0,0) {\includegraphics[width=\linewidth]{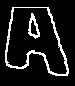}};
\node[text=black, rectangle, minimum width=\linewidth,minimum height=1em,below=0.0003cm of F1, inner sep=2.5,outer sep=0] {};
\end{tikzpicture}
\vfill
\begin{tikzpicture}[node distance = 0.1cm, auto]
\node[minimum width=\linewidth, inner sep=0,outer sep=0] (F1) at (0,0) {\includegraphics[width=\linewidth]{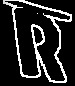}};
\node[text=black, rectangle, minimum width=\linewidth,minimum height=1em,below=0.0003cm of F1, inner sep=2.5,outer sep=0] {};
\end{tikzpicture}
\vfill
\begin{tikzpicture}[node distance = 0.1cm, auto]
\node[minimum width=\linewidth, inner sep=0,outer sep=0] (F1) at (0,0) {\includegraphics[width=\linewidth]{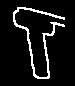}};
\node[text=black, rectangle, minimum width=\linewidth,minimum height=1em,below=0.0003cm of F1, inner sep=2.5,outer sep=0] {};
\end{tikzpicture}
\end{minipage}
}
\hfill
\subfloat[]{
\label{fig:features_h}
\begin{minipage}[b][7.3cm][t]{.12\textwidth}
\begin{tikzpicture}[node distance = 0.1cm, auto]
\node[minimum width=\linewidth, inner sep=0,outer sep=0] (F1) at (0,0) {\includegraphics[width=\linewidth]{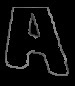}};
\node[draw=black!80,fill=gray!50, text=black, rectangle, minimum width=\linewidth,minimum height=1em,below=0.0003cm of F1, inner sep=2.5,outer sep=0] { \scriptsize{$\mu_I=109$}};
\end{tikzpicture}
\vfill
\begin{tikzpicture}[node distance = 0.1cm, auto]
\node[minimum width=\linewidth, inner sep=0,outer sep=0] (F1) at (0,0) {\includegraphics[width=\linewidth]{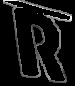}};
\node[draw=black!80,fill=gray!50, text=black, rectangle, minimum width=\linewidth,minimum height=1em,below=0.0003cm of F1, inner sep=2.5,outer sep=0] { \scriptsize{$\mu_I=118$}};
\end{tikzpicture}
\vfill
\begin{tikzpicture}[node distance = 0.1cm, auto]
\node[minimum width=\linewidth, inner sep=0,outer sep=0] (F1) at (0,0) {\includegraphics[width=\linewidth]{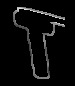}};
\node[draw=black!80,fill=gray!50, text=black, rectangle, minimum width=\linewidth,minimum height=1em,below=0.0003cm of F1, inner sep=2.5,outer sep=0] { \scriptsize{$\mu_I=122$}};
\end{tikzpicture}
\end{minipage}
}
\hfill
\subfloat[]{
\label{fig:features_i}
\begin{minipage}[b][7.3cm][t]{.12\textwidth}
\begin{tikzpicture}[node distance = 0.1cm, auto]
\node[minimum width=\linewidth, inner sep=0,outer sep=0] (F1) at (0,0) {\includegraphics[width=\linewidth]{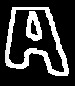}};
\node[text=black, rectangle, minimum width=\linewidth,minimum height=1em,below=0.0003cm of F1, inner sep=2.5,outer sep=0] {};
\end{tikzpicture}
\vfill
\begin{tikzpicture}[node distance = 0.1cm, auto]
\node[minimum width=\linewidth, inner sep=0,outer sep=0] (F1) at (0,0) {\includegraphics[width=\linewidth]{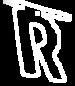}};
\node[text=black, rectangle, minimum width=\linewidth,minimum height=1em,below=0.0003cm of F1, inner sep=2.5,outer sep=0] {};
\end{tikzpicture}
\vfill
\begin{tikzpicture}[node distance = 0.1cm, auto]
\node[minimum width=\linewidth, inner sep=0,outer sep=0] (F1) at (0,0) {\includegraphics[width=\linewidth]{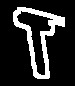}};
\node[text=black, rectangle, minimum width=\linewidth,minimum height=1em,below=0.0003cm of F1, inner sep=2.5,outer sep=0] {};
\end{tikzpicture}
\end{minipage}
}
\hfill
\subfloat[]{
\label{fig:features_j}
\begin{minipage}[b][7.3cm][t]{.12\textwidth}
\begin{tikzpicture}[node distance = 0.1cm, auto]
\node[minimum width=\linewidth, inner sep=0,outer sep=0] (F1) at (0,0) {\includegraphics[width=\linewidth]{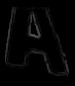}};
\node[draw=black!80,fill=gray!50, text=black, rectangle, minimum width=\linewidth,minimum height=1em,below=0.0003cm of F1, inner sep=2.5,outer sep=0] { \scriptsize{$\mu_{\Delta}=20$}};
\end{tikzpicture}
\vfill
\begin{tikzpicture}[node distance = 0.1cm, auto]
\node[minimum width=\linewidth, inner sep=0,outer sep=0] (F1) at (0,0) {\includegraphics[width=\linewidth]{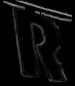}};
\node[draw=black!80,fill=gray!50, text=black, rectangle, minimum width=\linewidth,minimum height=1em,below=0.0003cm of F1, inner sep=2.5,outer sep=0] { \scriptsize{$\mu_{\Delta}=31$}};
\end{tikzpicture}
\vfill
\begin{tikzpicture}[node distance = 0.1cm, auto]
\node[minimum width=\linewidth, inner sep=0,outer sep=0] (F1) at (0,0) {\includegraphics[width=\linewidth]{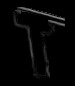}};
\node[draw=black!80,fill=gray!50, text=black, rectangle, minimum width=\linewidth,minimum height=1em,below=0.0003cm of F1, inner sep=2.5,outer sep=0] { \scriptsize{$\mu_{\Delta}=31$}};
\end{tikzpicture}
\end{minipage}
}

\caption{Computation sequence of the region features comprising the similarity measures used in the SLC grouping analysis. Different masks (d,g,i) are applied to the input color image (a), its gray scale version (b), or its gradient magnitude values (c) to calculate simple features: e.g. mean intensity value of the region (e), stroke width (f), mean intensity value on the outer boundary (h), or mean gradient magnitude on the region border. Feature computation details are provided on the main text.}
\label{fig:features}
\end{figure}
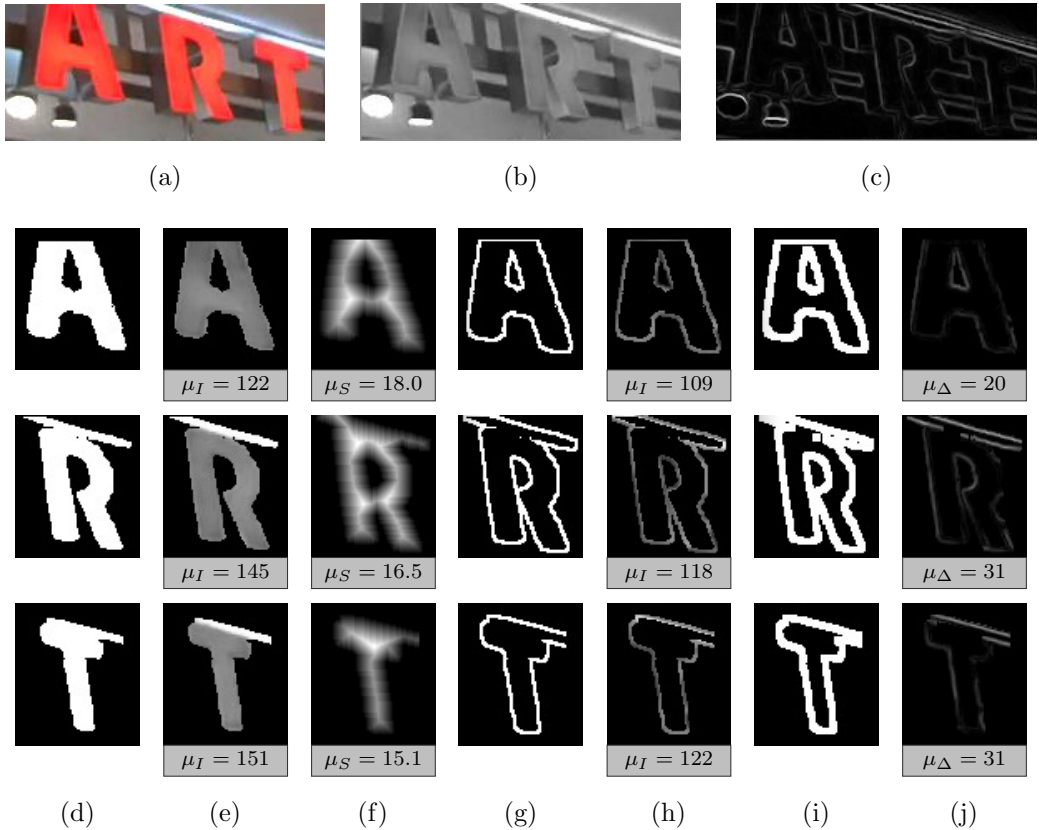
%
%

\noindent {\bf Intensity and color mean of the region.} We calculate the mean intensity value and the mean color, in L*a*b* colorspace, of the pixels that belong to the connected component. For this we make use of the region pixels as a mask (see Figure~\ref{fig:features_d}) on the grayscale or color image as shown in Figure~\ref{fig:features_e}.

\noindent {\bf Stroke width.} We approximate the stroke width using a Distance Transformed version of the connected component mask. Concretely, for each pixel of the connected component mask (\ref{fig:features_d}) we calculate the $L_1$ distance to the closest zero pixel using the algorithm described in~\cite{borgefors1986distance} with default parameters. From the distance transformed image (see Figure~\ref{fig:features_f}) we take the average of the row-wise maximum distance values as an approximation of the stroke width of the region.

\noindent {\bf Intensity and color mean of the outer boundary.} Similarly as for the intensity and color mean of the region, we calculate the mean intensity value and the mean color, in L*a*b* colorspace, of the pixels that belong to the outer boundary of the connected component. For this we first create a mask with the outer boundary pixels (see Figure~\ref{fig:features_g}) by dilation of the original region mask with a $3 \times 3$ rectangular kernel, and subtraction of the original region pixels. Then, we apply this mask on the grayscale, as in Figure~\ref{fig:features_h}, or color image.

\noindent {\bf Gradient magnitude mean on the border.} We calculate the mean of the gradient magnitude on the border of the region. Here we make use of a mask covering the pixels of both the inner and outer boundaries of the region (see Figure~\ref{fig:features_i}). We proceed similarly as before but combining a dilation and an erosion operation with the same $3 \times 3$ rectangular kernel. In this case the mask is applied to the gradient magnitude values of the input image as shown in Figure~\ref{fig:features_j}. 

\noindent {\bf Diameter of the region.} We fit an ellipse to the original region pixels using the algorithm described in~\cite{fitzgibbon1995buyer} and take the length of the major axis of the ellipse as the diameter of the region.

Each of these similarity features is used independently but coupled with spatial information, i.e. the $x,y$ coordinates of the regions' centers. So, independently of the similarity feature, we restrict the groups of regions that are of our interest to those that comprise spatially close regions. Thus, in the end we run the SLC analysis with seven different complementary distance metrics $\mathrm{d}^{(i)}$ with $i = \{1,\dots,7\}$:

\begin{equation}
\label{eq:dist1}
 \mathrm{d}^{(i)}(r_a,r_b) = (f^{i}(r_a) - f^{i}(r_b))^2 + (x_a - x_b)^2 + (y_a - y_b)^2
\end{equation}

where $f^{i}$ is one of the simple similarity features listed above, and $\{x_a, y_a\}$, $\{x_b,y_b\}$ are the coordinates of the centers of regions $r_a$ and $r_b$ respectively. Using $\mathrm{d}^{(i)}(r_a,r_b)$ as defined equation~\ref{eq:dist1} our clustering analysis remains rotation invariant by using the squared Euclidean distance between the regions' centers. And thus our method is able to generate proposals for arbitrarily oriented text instances. Optionally it is possible to prioritize horizontally aligned regions to merge first by adding a small factor $\lambda \in [0,1]$ to the $x$ coordinates term:

\begin{equation}
\label{eq:dist2}
 \mathrm{d}^{(i)}(r_a,r_b) = (f^{i}(r_a) - f^{i}(r_b))^2 + \lambda (x_a - x_b)^2 + (y_a - y_b)^2
\end{equation}

At this point it is obvious that the more we diversify our grouping strategies, the more chances we have to find a good proposal for a given target word, but this is at the cost of increasing the total number of word proposals. In the following we list a number of possible diversification strategies that can be combined in different ways. In the experimental section~\ref{sec:exp_diversification} we will analyze the performance of different combinations of these strategies in order to find an optimal configuration as trade-off between detection recall and number of generated proposals.

\noindent \textbf {Diversification by complementary similarity cues.} We use SLC clustering as explained before with different similarity measures.

\noindent \textbf {Diversification by use of different color channels.} We extract regions using the MSER algorithm on different color channels separately.

\noindent \textbf {Diversification by use of different spatial pyramid levels.} We extract MSER regions from a three-level spatial pyramid.

\begin{algorithm}[h]
\SetAlgoLined
 \caption{TextProposals}
 \label{alg:textproposals}
 \KwIn{RGB image $I$}
 \KwOut{Set of bounding box proposals $B$ and their scores}
 Initialize $B = \emptyset$\;
 Initialize $F = $ list of similarity features\;
 Extract input channels and scales $C = \{c_1, \dots , c_n\}$ from $I$\;
 \ForEach{channel $c \in C$}{
  Obtain MSER~\cite{Matas2004} regions $R_c = \{r_1, \dots, r_m\}$ from $c$\;
  Obtain coordinates of regions' centers $\{x_1, \dots, x_m\}$, $\{y_1, \dots, y_m\}$\;
  \ForEach{feature $f \in F$}{
  Calculate feature set $S_{fc} = \{s_1, \dots, s_m\}$ with $s_i = (f(r_i),x_i,y_i)$\;
  Build dendrogram $D_{fc}$ applying SLC clustering over $S_{fc}$\;
  Extract bounding boxes $B_{fc}$ and scores for each $node \in D_{fc}$\;
  $B = B \cup B_{fc}$\;
  }
 }
 Sort by score and deduplicate bounding boxes in $B$\; 
\end{algorithm}

Using all mentioned diversification strategies would generate a total of $84$ similarity hierarchies ($3$ pyramid levels $\times$ $4$ color channels $\times$ $7$ similarity cues). Accordingly, the complete TextProposals method is detailed in Algorithm~\ref{alg:textproposals}, where the procedure used to score and rank bounding boxes will be explained in the next section.

\subsection{Text Proposals Ranking}
\label{sec:rank}

Once we have created a similarity hierarchy using SLC, with each of its nodes representing a text proposal, we need an efficient way to sort the nodes in a ranked list of proposals that prioritizes the most promising ones.

For this we propose the use of a weak classifier to obtain a text-likeliness measure of proposals. The idea is thus to train a classifier to discriminate between text and non-text hypotheses, and to produce a confidence value that can be used to rank them. 
Since the classifier is going to be evaluated on every node of our hierarchies, we aim to use a fast classifier and features with low computational cost. We train a Real AdaBoost classifier with decision stumps using as features $F^{i}(G)$ the coefficients of variation of the individual region features $f^{i}$ described in the previous section~\ref{sec:grouping} (e.g. stroke width, diameter, foreground mean intensity value, etc.): 

\begin{equation}
F^{i}(G) =  {\sigma^{i}}/{\mu^{i}}
\end{equation}

\noindent
where $\mu^{i}$ and $\sigma^{i}$ are respectively the mean and standard deviation of the i'th feature $f^{i}$ in a particular group $G$,  $\{f^{i}(r) : r \in G\}$.

In addition we also use another set of simple bounding box based features that proved to improve the weak classifier performance. First we obtain the bounding box of the group constituent regions, and the bounding box enclosing only the regions' centers. A set of simple features originates from calculating: (1) the ratio between the areas of both bounding boxes, (2) the ratio between their widths, (3) the ratio between their heights, (4) the ratio between the difference of their left-most $x$ coordinates and the difference of their right-most $x$ coordinates, and (5) the same as in 4 but for the top/bottom $y$ coordinates differences.

All the used group-level features can be computed efficiently in an incremental way along the SLC hierarchies, and all $f^{i}$ individual region features have been previously computed for the cluster analysis.

To train the classifier we mine for positive and negative samples in the ICDAR2013 training set images with the following procedure: first, for each training image we generate similarity hierarchies using all the diversification strategies described in section~\ref{sec:grouping}; second, for each regions grouping (node) in each of the $84$ generated hierarchies we find the best matching ground-truth annotation in terms of their bounding box Intersection over Union (IoU); then we take the group as a positive sample if IoU $> 0.7$, otherwise we take it as a negative sample if IoU $< 0.2$ and it does not fully overlap with a ground truth bounding box. This way we obtain approximately 200k positive samples and 1 million negative samples, then we balance the training data by randomly selecting 200k of the negative samples. 

At test time we obtain a list of proposals ranked with scores provided by evaluation of the AdaBoost classifier on every node of our hierarchies.

\subsubsection{Hierarchy-based inference and optimization}
\label{sec:hierarchy_optimization}

With the ranked list of word proposals we can build an end-to-end pipeline straightforward by evaluating a holistic word recognition method in all of them (or in the $N$ best ranked) and then performing a Non-Maximal Suppression (NMS) strategy as in~\cite{jaderberg2014reading}. However, being our word proposals organized in a hierarchy where each node (i.e. proposal) has an inclusion relation with its respective childs allows us to do a much more efficient inference.

First of all, object proposals algorithms normally need to remove duplicated detections in order to not waste computation resources by evaluating the final classification model more than once in the same bounding box. 
In our case the number of duplicated detections may account for large numbers, because in many cases our agglomerative clustering merges overlapping regions that produce no change in the merged grouping bounding box. At the time of building our similarity hierarchies we take this into account and set a flag for whether the classifier response (both the word transcription and the classification score/probability) has to be calculated in a particular node or it can be just propagated from it's childs. This process of deduplication within a single hierarchy has no cost for us. Since we have several independent hierarchies, we also maintain a hash table of evaluated bounding boxes.

Similarly, we also take advantage of the inclusion relation between nodes in the hierarchy to do an implicit Non-Maximal Suppression (NMS) of the end-to-end system outputs. For this, we walk the nodes of the hierarchy and evaluate the holistic word recognizer in all nodes where it is worth (because they are among the $N$ best proposals and their bounding box has not been already evaluated), and then we select only the nodes for which the model recognition score is better than for any of its descendants and any of its ancestors in the hierarchy. This is, a given node $A$ is selected as an output of the end-to-end system if its recognition score $\mathcal{R_s}$ is larger than a given classification threshold $\mathcal{R_s}(A) > \tau$ and the following inequalities hold:

\begin{equation}
  \mathcal{R_s}(A) > \mathcal{R_s}(B), \forall B \in suc(A) 
\end{equation}
\begin{equation}
  \mathcal{R_s}(A) \geq \mathcal{R_s}(C), \forall C \in anc(A) 
\end{equation}

\noindent
where $suc(A)$ is the set of all successor nodes of $A$ and $anc(A)$ is the set of its ancestors.

While a final NMS procedure is needed to find an agreement between the different hierarchies outputs, at this point the number of boxes to be processed with NMS is minimal.


\section{Experiments and Results}
\label{sec:experiments}
In this section we conduct exhaustive experimentation of our text-specific object proposals algorithm. We basically do two different kind of experiments: in Section~\ref{sec:exp_proposals} we analyze the quality of the word proposals generated by our method; in Section~\ref{sec:exp_end2end} we integrate our method with two well known holistic word classifiers~\cite{Almazan2014,jaderberg2014reading} and evaluate the end-to-end word spotting performance of the system. 

In our experiments we make use of the following scene text datasets: the ICDAR Robust Reading Competitions datasets (ICDAR2003~\cite{Lucas2003}, ICDAR2013~\cite{karatzas2013icdar}, and ICDAR2015~\cite{karatzas2015}), the Street View Text dataset (SVT)~\cite{Wang2010}, and the Multi-Language end-to-end (MLe2e) dataset~\cite{gomez2016}. In all cases we provide results for their official test sets. The ICDAR2013 train set has been used to train the proposals ranking model described in Section~\ref{sec:rank}.

\subsection{Quality of object proposals}
\label{sec:exp_proposals}
The evaluation framework used in all this section is the standard for object proposals methods~\cite{hosang2015} and is based on the analysis of the detection recall achieved by a given method under certain conditions. Recall is calculated as the ratio of ground truth bounding boxes that have been predicted among the word proposals with an intersection over union (IoU) larger than a given threshold. This way, we evaluate the recall at a given IoU threshold as a function of the number of proposals, and the quality of the first ranked $N$ proposals by calculating their recall at different IoU thresholds. 


\subsubsection{Evaluation of diversification strategies}
\label{sec:exp_diversification}
 
We analyze the performance of different variants of our method by evaluating different combinations of diversification strategies presented in Section~\ref{sec:method}. Table~\ref{tab:diversity_comparison} shows the average number of proposals per image, recall rates at various IoU thresholds, and average time performance obtained with some of the possible combinations. 

\begin{table}[h]
\setlength{\tabcolsep}{3pt}
\begin{center}
\scriptsize
    \begin{tabularx}{\linewidth}{ X c c c c c c c}
\toprule
    Method & \# prop. & 0.5 IoU & 0.6 IoU & 0.7 IoU & 0.8 IoU & 0.9 IoU & Avg. time(s)\\ 
\midrule
P0+I+D  		&1614  & 0.86 & 0.74 & 0.53 & 0.26 & 0.07 & 0.36\\
P0+I+F  		&1455  & 0.88 & 0.81 & 0.63 & 0.31 & 0.09 & 0.36\\
P0+I+B  		&1488  & 0.84 & 0.73 & 0.52 & 0.23 & 0.06 & 0.36\\
P0+I+S  		&1596  & 0.80 & 0.71 & 0.50 & 0.20 & 0.06 & 0.36\\
P0+I+DFBGS 		&4588  & 0.94 & 0.88 & 0.71 & 0.41 & 0.11 & 0.87\\
P0+I+DFBGS$F_{lab}$$B_{lab}$&5441  & 0.94 & 0.88 & 0.71 & 0.41 & 0.11 & 1.16\\
P0+RGB+DFBGS		&12996 & 0.94 & 0.91 & 0.82 & 0.52 & 0.19 & 2.42\\
\bf{P0P1+RGB+DFBGS}	&16795 & 0.95 & \bf{0.94} & \bf{0.88} & 0.59 & 0.22 & 2.74\\
P0P1P2+RGB+DFBGS 	&18297 & 0.95 & 0.94 & 0.88 & 0.61 & 0.25 & 2.91\\
P0P1P2+RGBI+DFBGS 	&21663 & 0.96 & 0.94 & 0.88 & 0.61 & 0.26 & 3.66\\
\bottomrule 
    \end{tabularx}
\end{center}
\caption{Detection recall at different IoU thresholds and running time comparison using different diversification strategies in validation data. Color channels: (R), (G), (B), and (I). Spatial pyramid levels: (P0) $1:1$ scale, (P1) $1:2$ scale, (P2) $1:4$ scale. Similarity cues: (D) Diameter, (F) Foreground intensity, (B) Background intensity, (G) Gradient, (S) Stroke width, ($F_{lab}$) Foreground Lab color, and ($B_{lab}$) Background Lab color. }
\label{tab:diversity_comparison}
\end{table}

\begin{figure}[h]
\begin{center}
   \subfloat[``EYES'' IoU=0.56]{\includegraphics[width=0.33\linewidth]{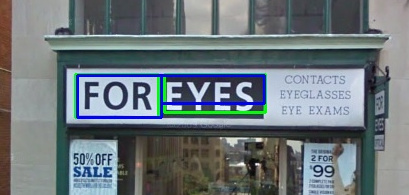}\label{iou_disagreement_1}}
   \subfloat[``Donald'' IoU=0.53]{\includegraphics[width=0.33\linewidth]{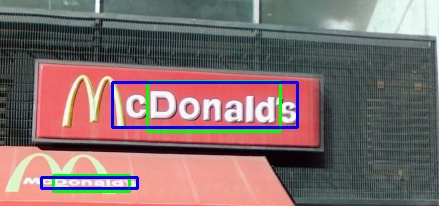}\label{iou_disagreement_2}}
   \subfloat[``STARBUCKS'' IoU=0.52]{\includegraphics[width=0.33\linewidth]{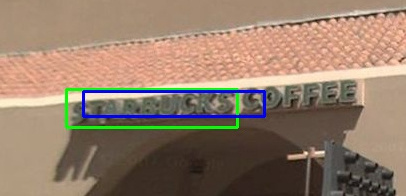}\label{iou_disagreement_3}}\\
   \subfloat[``Martin'' IoU=0.50]{\includegraphics[width=0.33\linewidth]{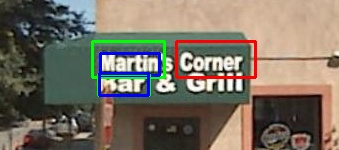}\label{iou_disagreement_4}}
   \subfloat[``Colorado'' IoU=0.65]{\includegraphics[width=0.33\linewidth]{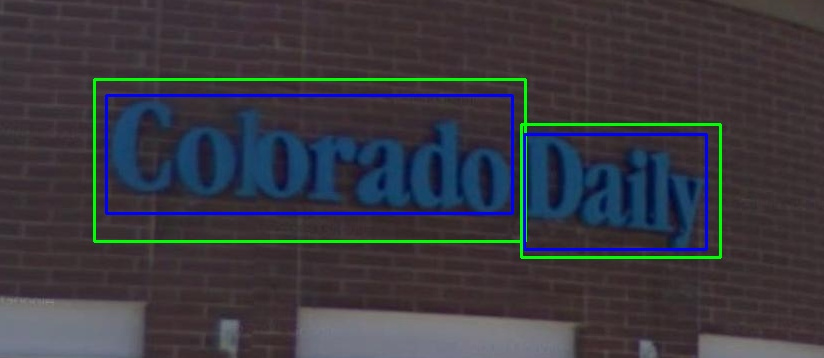}\label{iou_disagreement_5}}
   \subfloat[``FOOD'' IoU=0.54]{\includegraphics[width=0.33\linewidth]{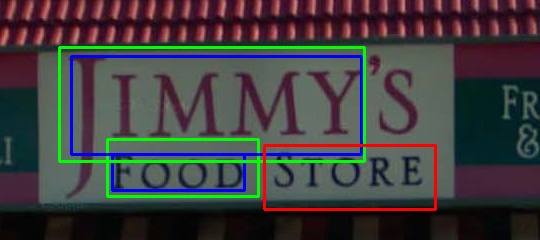}\label{iou_disagreement_6}}
\end{center}
   \caption{Examples of word proposals (blue) matching ground-truth annotations (green) and their IoU scores. IoU scores around $0.5$ may correspond to bad localizations covering only part of the word~\protect\subref{iou_disagreement_1} or neighboring characters~\protect\subref{iou_disagreement_2}~\protect\subref{iou_disagreement_3}. However in some cases also good localizations are scored low~\protect\subref{iou_disagreement_4}~\protect\subref{iou_disagreement_5}~\protect\subref{iou_disagreement_6} due to human annotation inconsistency. In the extreme case missing detections (in red) correspond to correctly detected words with IoU scores under $0.5$~\protect\subref{iou_disagreement_4}~\protect\subref{iou_disagreement_6}.}
\label{fig:iou_disagreement}
\end{figure}

Notice that while an IoU score of $0.5$ is normally accepted in generic object detection tasks, for scene text detection we would rather prefer better quality proposals, e.g. with IoU around $0.7$, because intuitively proposals with a $0.5$ IoU are likely to contain only part of the ground-truth word and thus may complicate the final recognition. On the other hand good word proposals do not necessary need to reach large IoU scores because in some cases the ground-truth information is quite ambiguous in describing the bounding box padding allowed for a ``word object'' annotation. To illustrate this issues we show in Figure~\ref{fig:iou_disagreement} some examples of word proposals generated by our method and the IoU with their matching ground truth bounding boxes. Ultimately, we note that IoU-based analysis of object proposals is not only task dependent but also very sensible to the dataset bounding boxes annotations quality/consistency. By manual inspection in the validation dataset we decided to focus our baseline analysis at the $0.6$ and $0.7$ IoU thresholds.
 
As a result of this analysis we have selected for further evaluation a particular combination of diversification strategies (see bold text in Table~\ref{tab:diversity_comparison}) as a trade-off between detection recall and number of proposals. This combination will be used in the rest of the experiments in this paper as ``TextProposals''. 

It can be appreciated that the selected combination produces significant gains at $0.6$ and $0.7$ IoU thresholds compared to other less diverse options. In particular, the difference of using three color channels (RGB) instead of one (I) is notable, and adding the second level (P1) while produces a rather small gain has practically no cost in number of proposals. We also observe redundancy in using the intensity channel in combination with RGB channels, as well as in adding the Lab color similarity cues ($F_{lab}$,$B_{lab}$).

\subsubsection{Evaluation of proposals' rankings}

\begin{figure}
\begin{center}
   \includegraphics[width=0.32\linewidth]{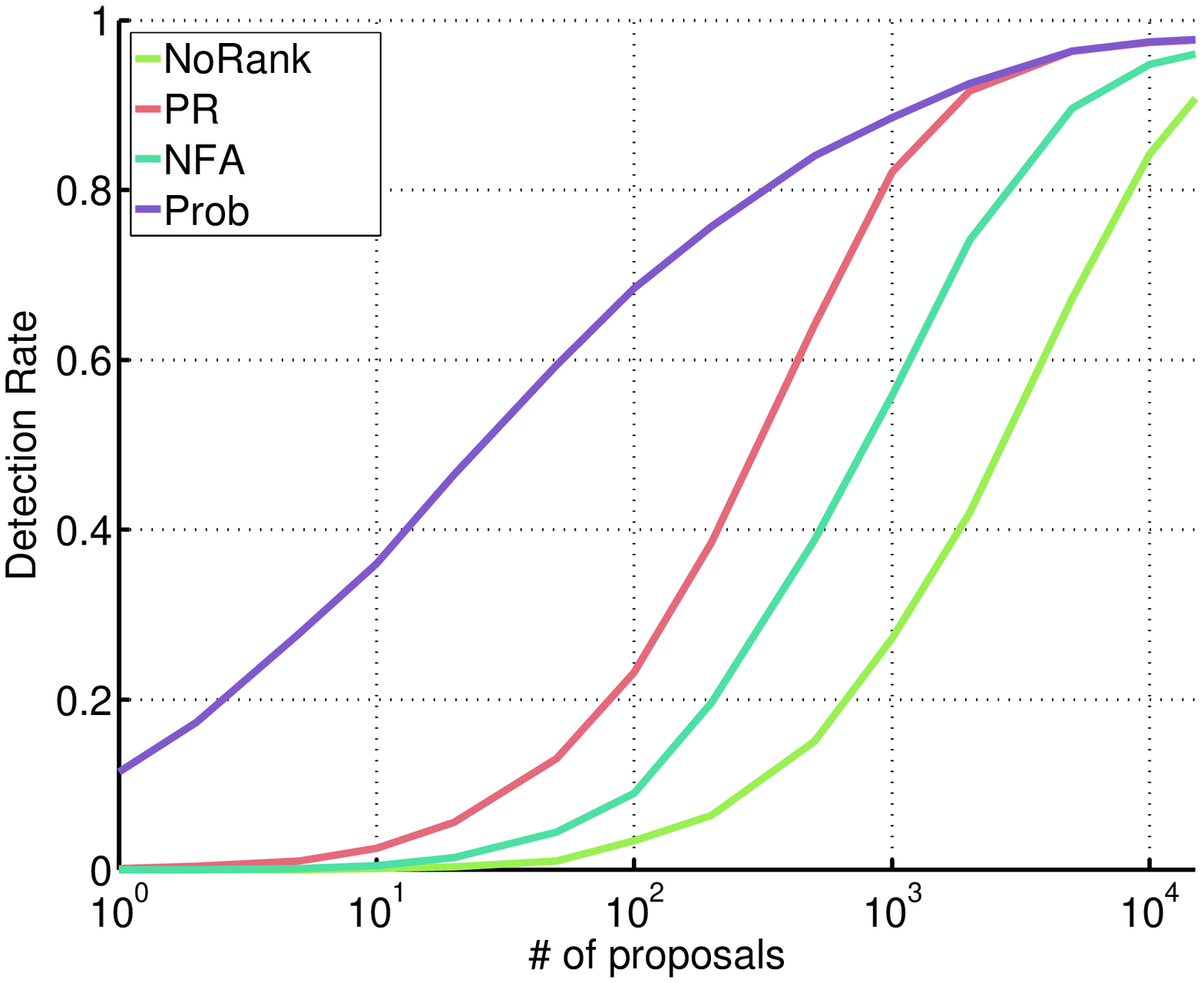}
   \includegraphics[width=0.32\linewidth]{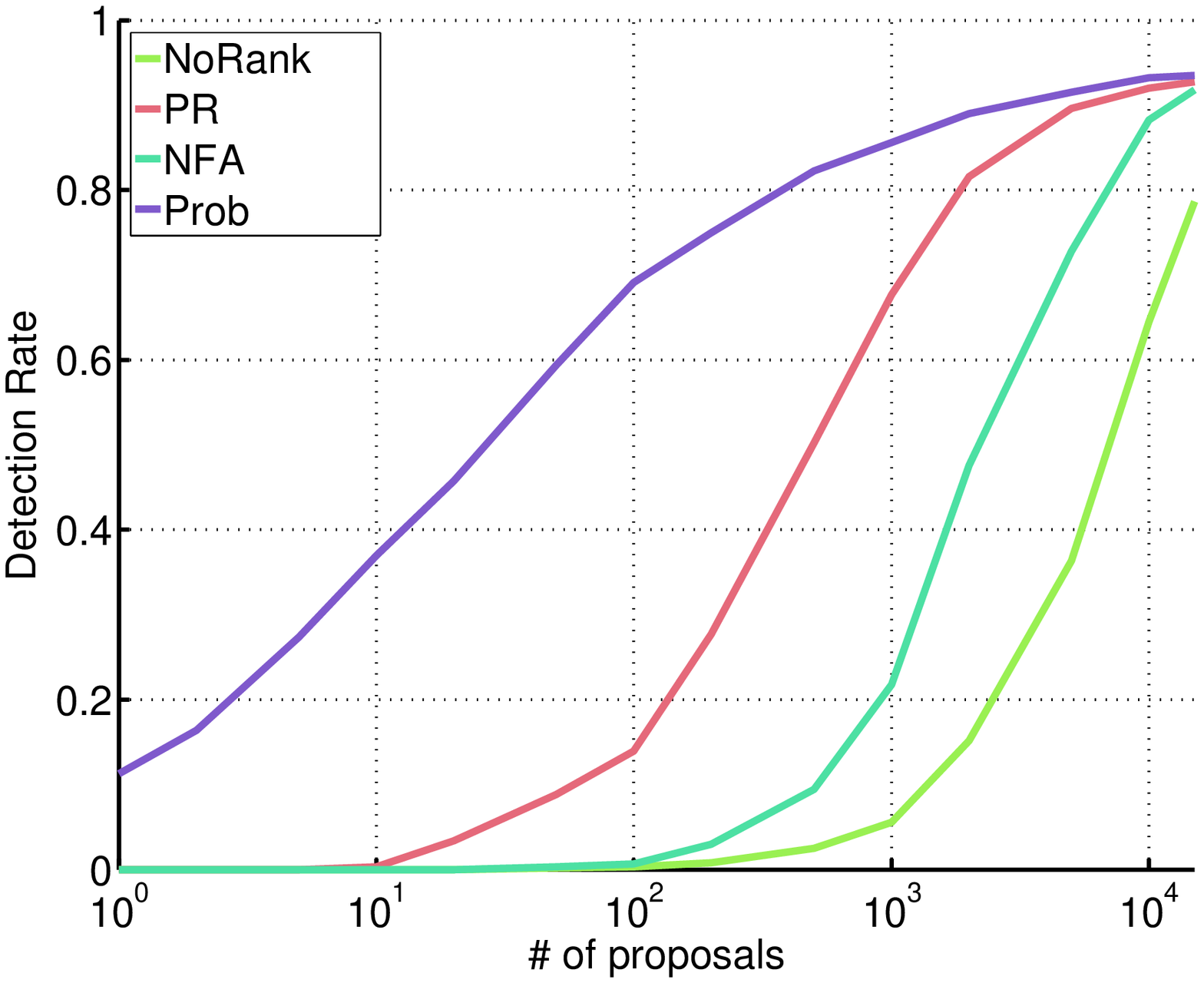}
   \includegraphics[width=0.32\linewidth]{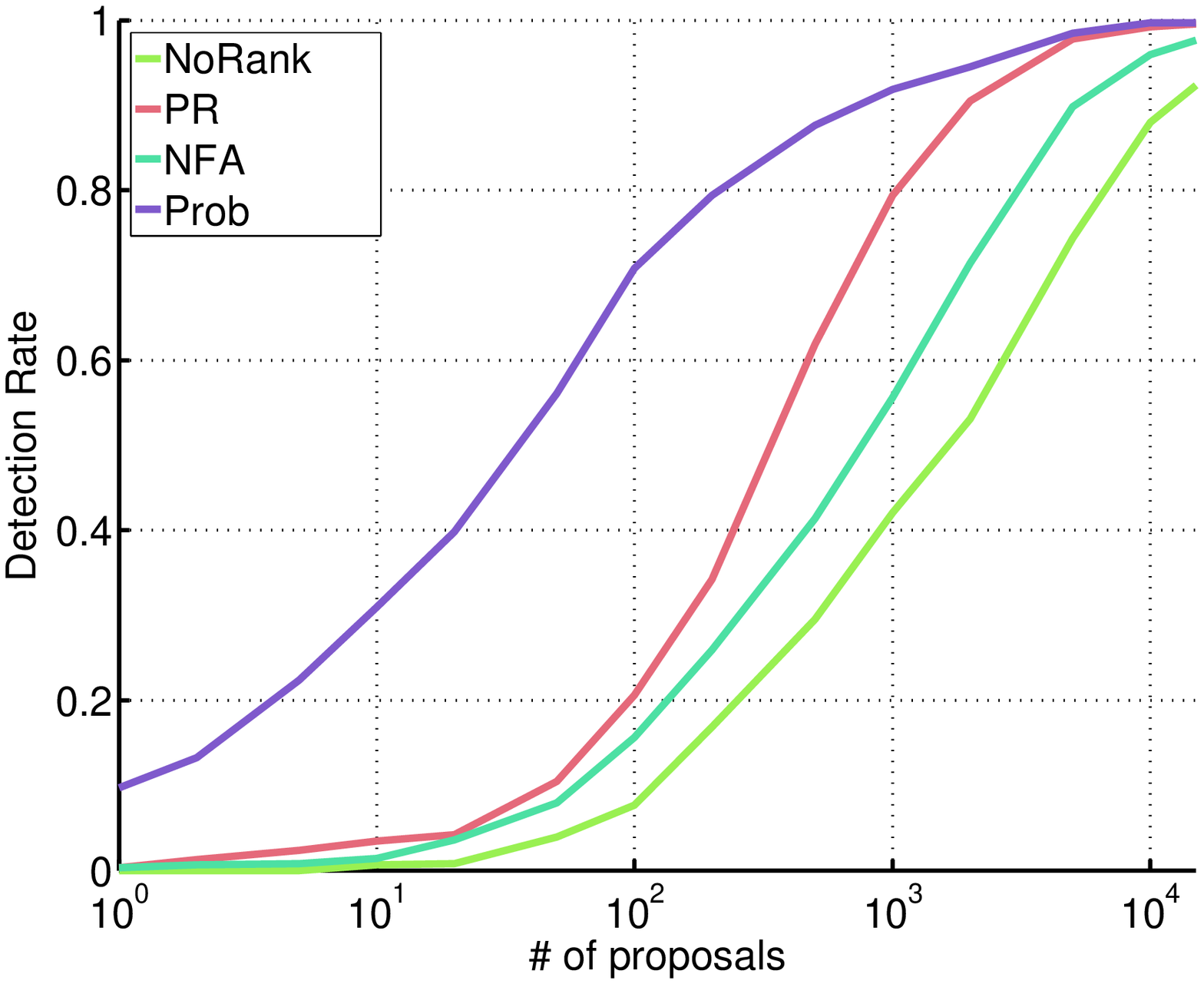}
\end{center}
   \caption{TextProposals performance at $0.5$ IoU using various ranking strategies in ICDAR2013(left), SVT(middle), and MLe2e(right) datasets: (PR) Pseudo-random ranking, (NFA) Meaningfulness ranking, (Prob) the ranking provided by the weak classifier.}
\label{fig:rank_comparison}
\end{figure}

Figure~\ref{fig:rank_comparison} shows the effect that different ranking strategies have in our TextProposals in three different datasets. The provided plots illustrate how the recall varies by taking into account only a certain number of proposals, prioritizing the ones with higher rank. Apart of the ranking provided by the weak text classifier described in section~\ref{sec:rank} we analyze the performance of three other ranking strategies: a pseudo-random ranking, a cluster meaningfulness ranking, and a totally flat ranking (i.e. every proposal has the same priority and thus the first generated is the first evaluated).
 
The pseudo-random ranking is the one proposed in~\cite{Uijlings2013}.
 For the meaningfulness ranking (NFA), we make use of the cluster quality measure detailed in~\cite{cao2004}. Intuitively this value is small for groups comprising a set of very similar regions, that are densely concentrated in small volumes of the feature space, and Thus seems well indicated for measuring text-likeliness of our proposals. 

As can be appreciated the area under the curve (AUC) provided by the weak text classifier is better than the rest of strategies analyzed. Particularly important is the observation that with this ranking using only the best $100$ proposals (in average) we reach around $70\%$ of the maximum attainable recall. Since the overhead of using the classifier is negligible we use this ranking strategy for the rest of the experiments in this paper.

\subsubsection{Comparison with state of the art generic methods} 

In this section we analyze the performance of our TextProposals in comparison with the following state-of-the-art generic object proposals methods: BING~\cite{cheng2014}, EdgeBoxes~\cite{zitnick2014}, Randomized Prim's~\cite{manen2013} (RP), and Geodesic object proposals~\cite{krahenbuhl2014} (GOP). We use publicly available code of these methods with default parameters. 

\begin{figure}
\begin{center}
\includegraphics[width=0.32\linewidth]{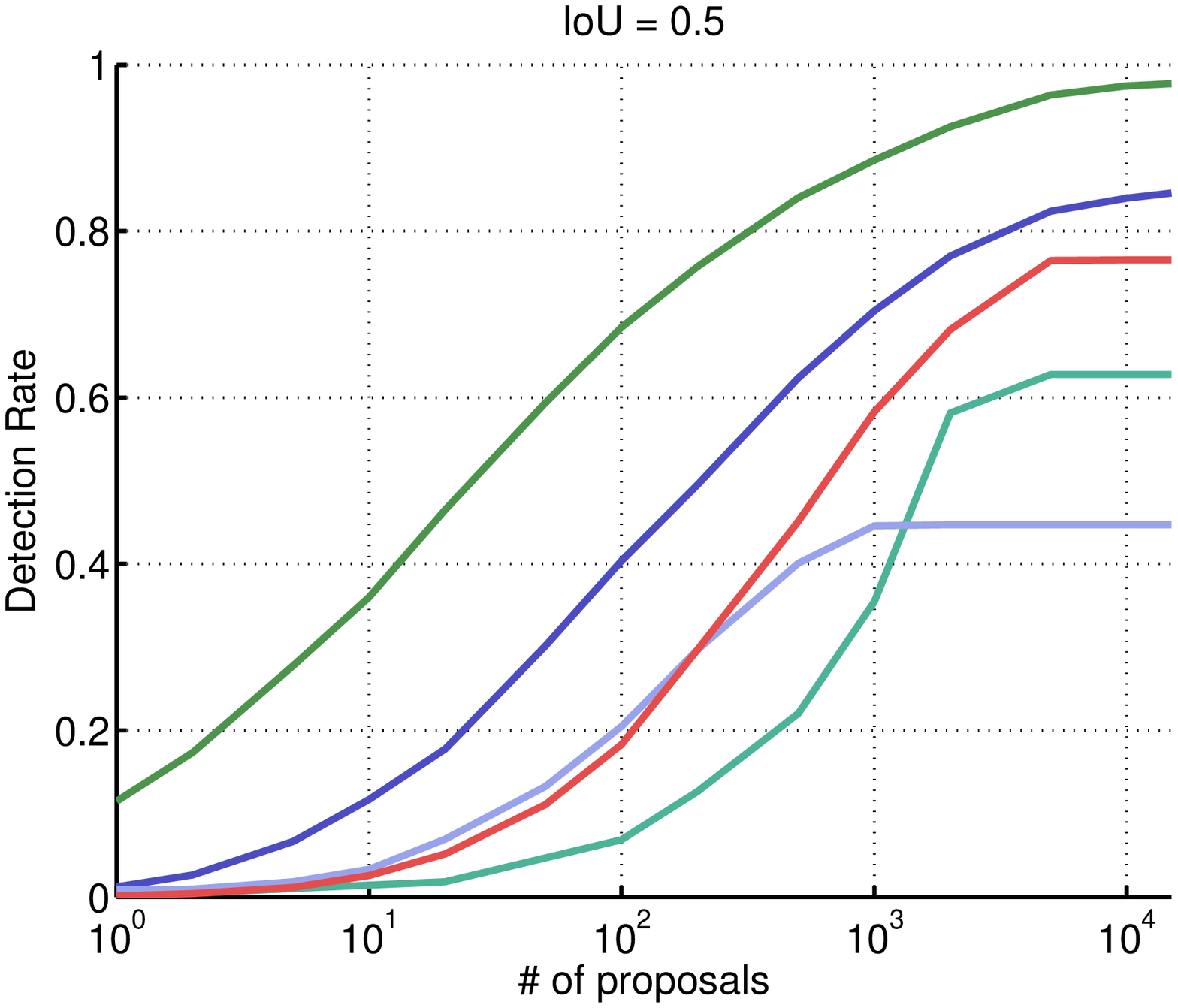} \includegraphics[width=0.32\linewidth]{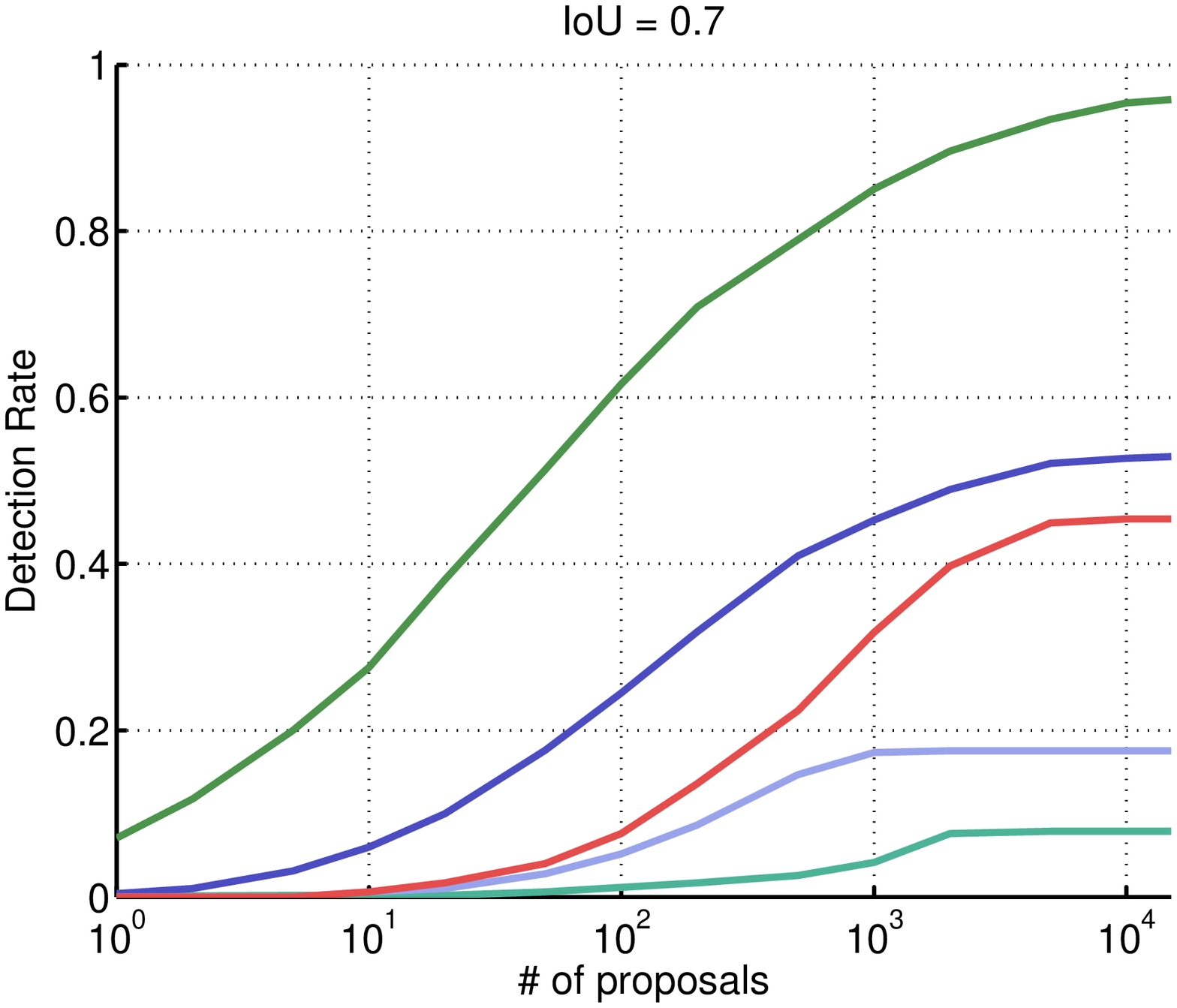} \includegraphics[width=0.32\linewidth]{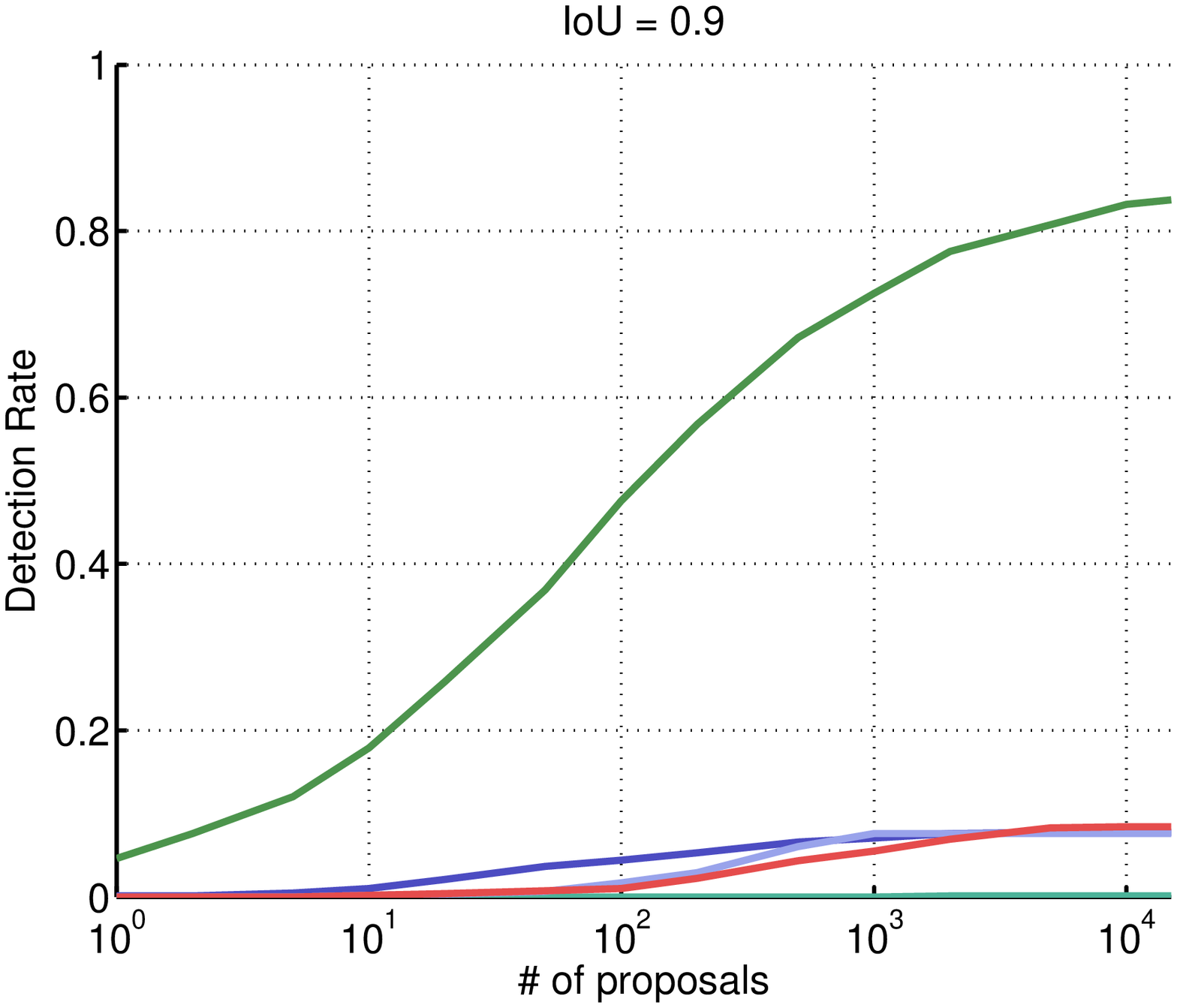}\\
\fbox{\includegraphics[width=0.60\linewidth]{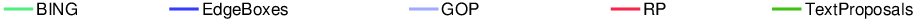}}\\
\includegraphics[width=0.32\linewidth]{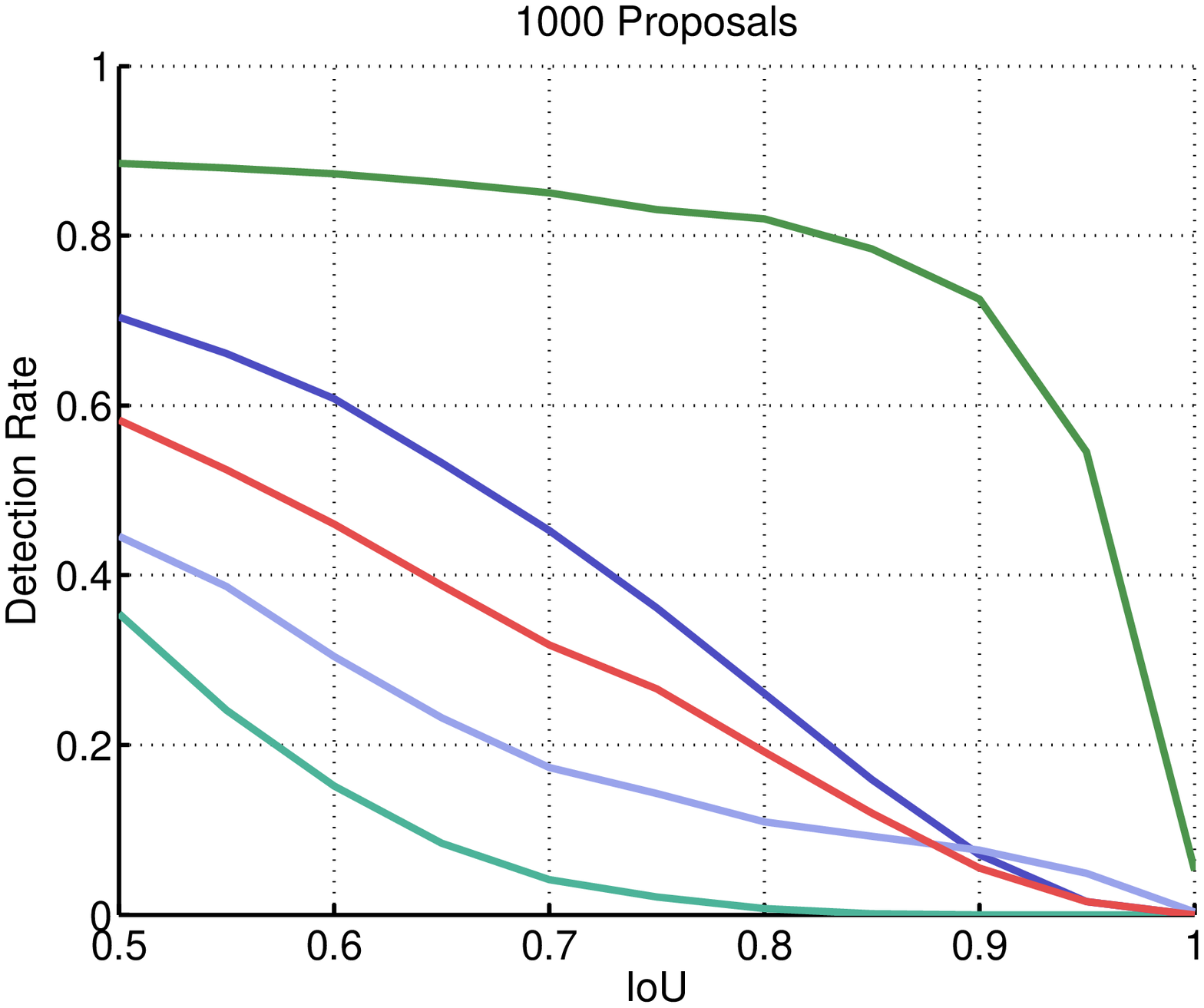} \includegraphics[width=0.32\linewidth]{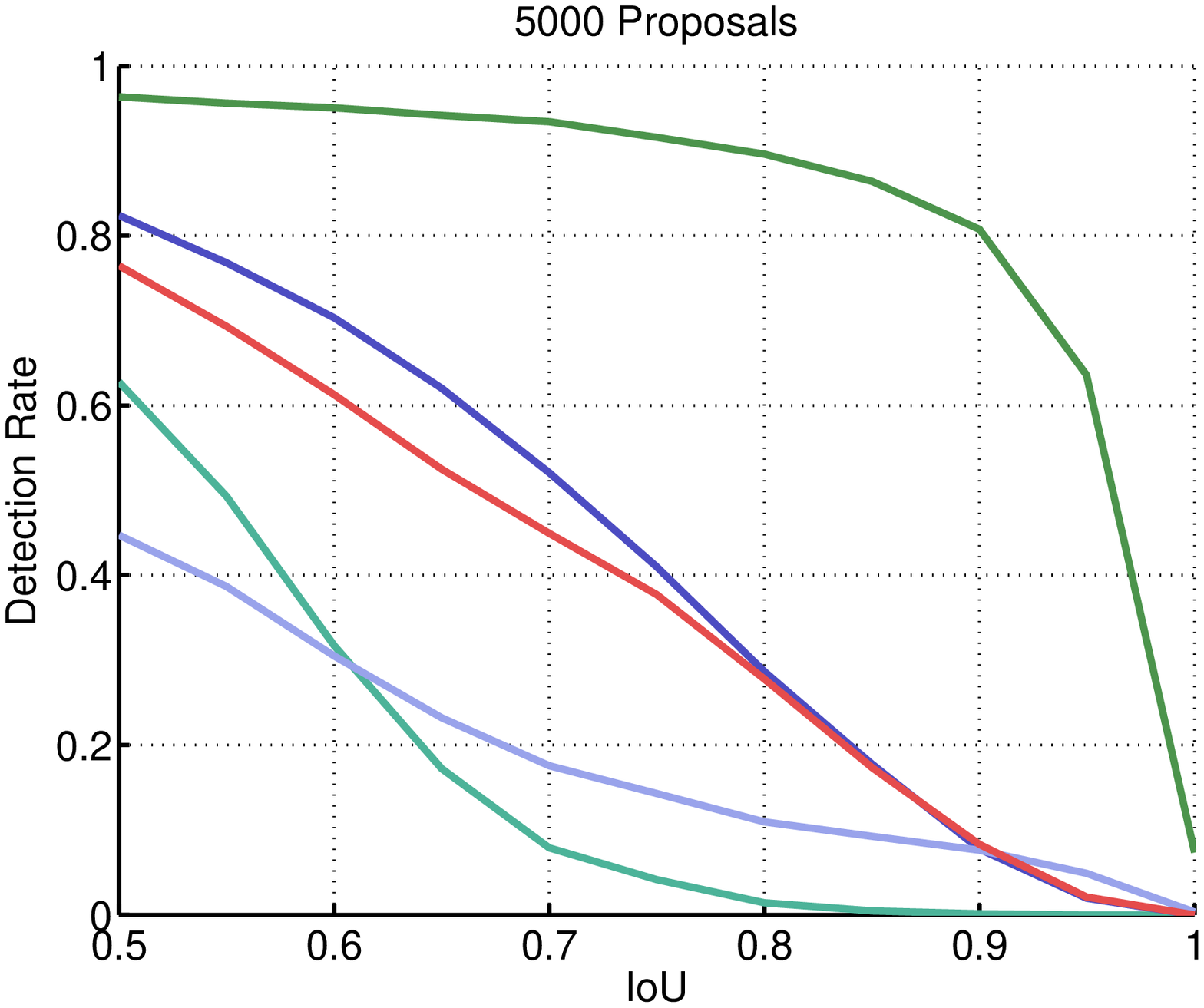} \includegraphics[width=0.32\linewidth]{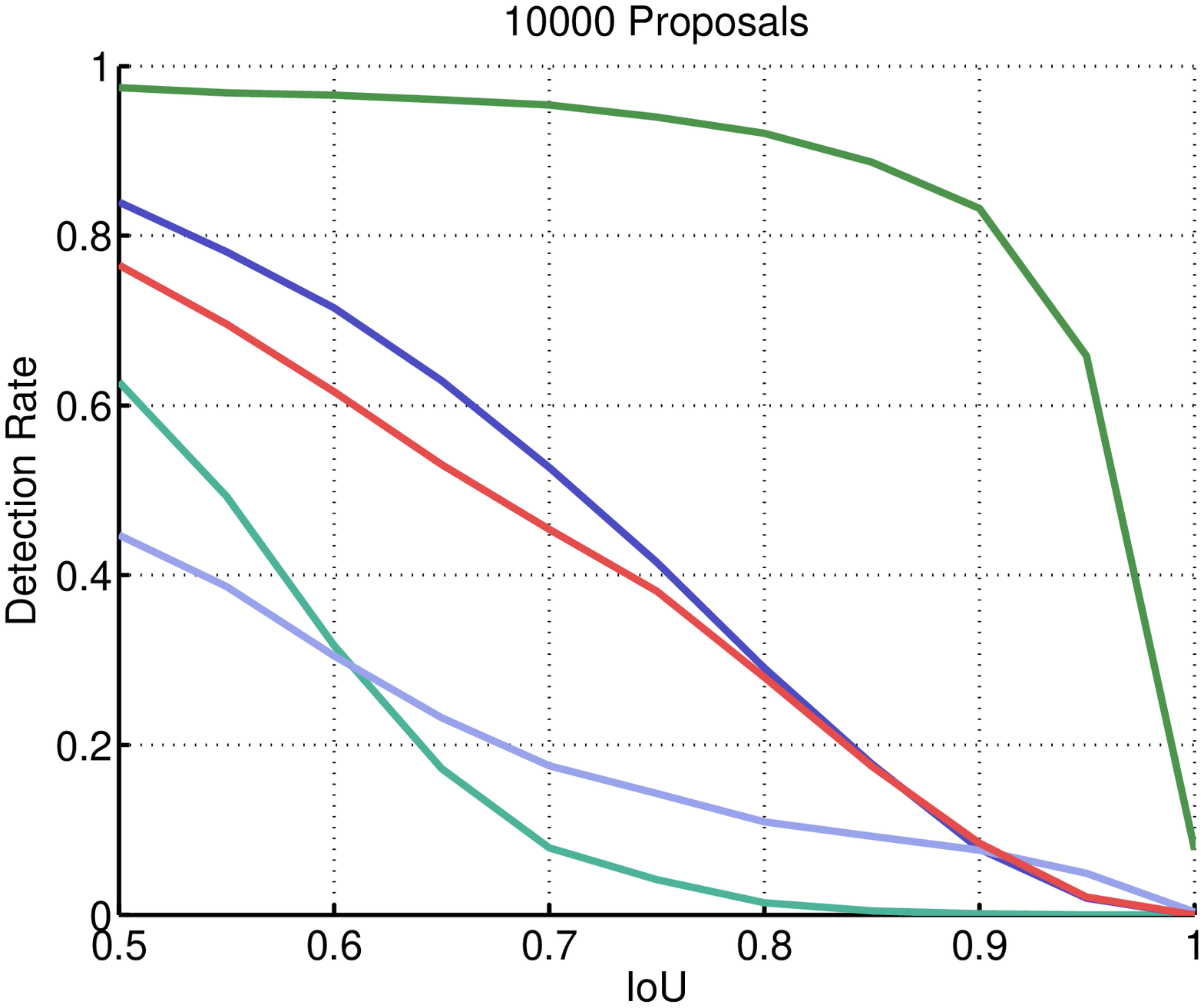}
\end{center}
   \caption{A comparison of various state-of-the-art object proposals methods in the ICDAR2013 dataset. (top) Detection rate versus number of proposals for various intersection over union thresholds. (bottom) Detection rate versus intersection over union threshold for various fixed numbers of proposals.}
\label{fig:plot_icdar}
\end{figure}

Table~\ref{tab:icdar_results} shows the performance comparison of all the evaluated methods in the ICDAR2013 dataset, while a more detailed comparison is provided in Figure~\ref{fig:plot_icdar}. All time measurements have been calculated by executing code in a single thread on the same i7 CPU. 

\begin{table}
\begin{center}
\scriptsize
    \begin{tabularx}{\linewidth}{ X c c c c c}
\toprule
    Method & \# prop. & 0.5 IoU & 0.7 IoU & 0.9 IoU & time(s)\\  
\midrule
    BING~\cite{cheng2014} & 2716 & 0.63 & 0.08 & 0.00 & 1.21\\  
    EdgeBoxes~\cite{zitnick2014} & 9554 & 0.85 & 0.53 & 0.08 & 2.24\\  
    RP~\cite{manen2013} & 3393 & 0.77 & 0.45 & 0.08 & 12.80\\  
    GOP~\cite{krahenbuhl2014} & 855 & 0.45 & 0.18 & 0.08 & 4.76\\  
\midrule
 \textbf{TextProposals}& 13719 & \textbf{0.98} & \textbf{0.96} & \textbf{0.84} & 2.85\\ 
\bottomrule 
    \end{tabularx}
\end{center}
\caption{Average number of proposals, recall at different IoU thresholds, and running time comparison with Object Proposals state of the art algorithms in the ICDAR2013 dataset.}
\label{tab:icdar_results}
\end{table}

As can be appreciated our method outperforms all evaluated algorithms in terms of detection recall on this dataset. Moreover, it is important to notice that detection rates of all the generic Object Proposals heavily deteriorate for large IoU thresholds while our method provides much more stable rates, indicating a better coverage of text objects (see the high AUC difference in Figure~\ref{fig:plot_icdar} bottom plots).

The low recall rates of grouping-based object proposals algorithms were foreseeable, since as discussed before these methods are designed to detect generic objects by agglomerating adjacent regions. Similarly, the BING proposals algorithm is trained to detect single body objects with compact shapes. In the case of the EdgeBoxes algorithm the provided comparison makes more sense, because it has been already integrated in a scene text end-to-end recognition pipeline~\cite{jaderberg2014reading} with good results. Thus, a direct comparison can be established, and the better performance of our TextProposals in Table~\ref{tab:icdar_results} and Figure~\ref{fig:plot_icdar} allows us to hypothesize a consequent improvement on the end-to-end results of~\cite{jaderberg2014reading} by exchanging the proposals generation module. This claim is supported with experimental evidence in section~\ref{sec:exp_end2end}.

\begin{figure}
\begin{center}
\includegraphics[width=0.32\linewidth]{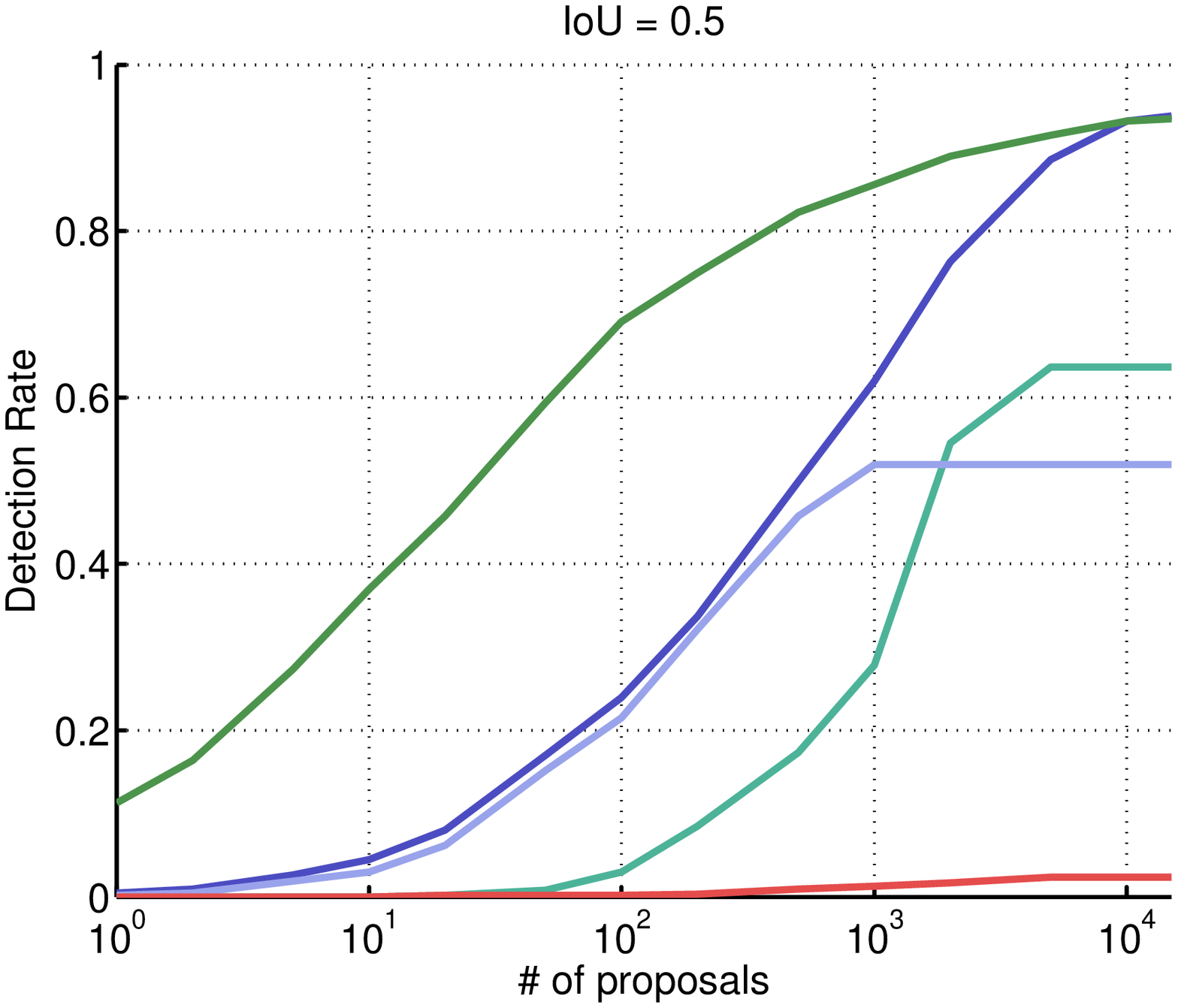} \includegraphics[width=0.32\linewidth]{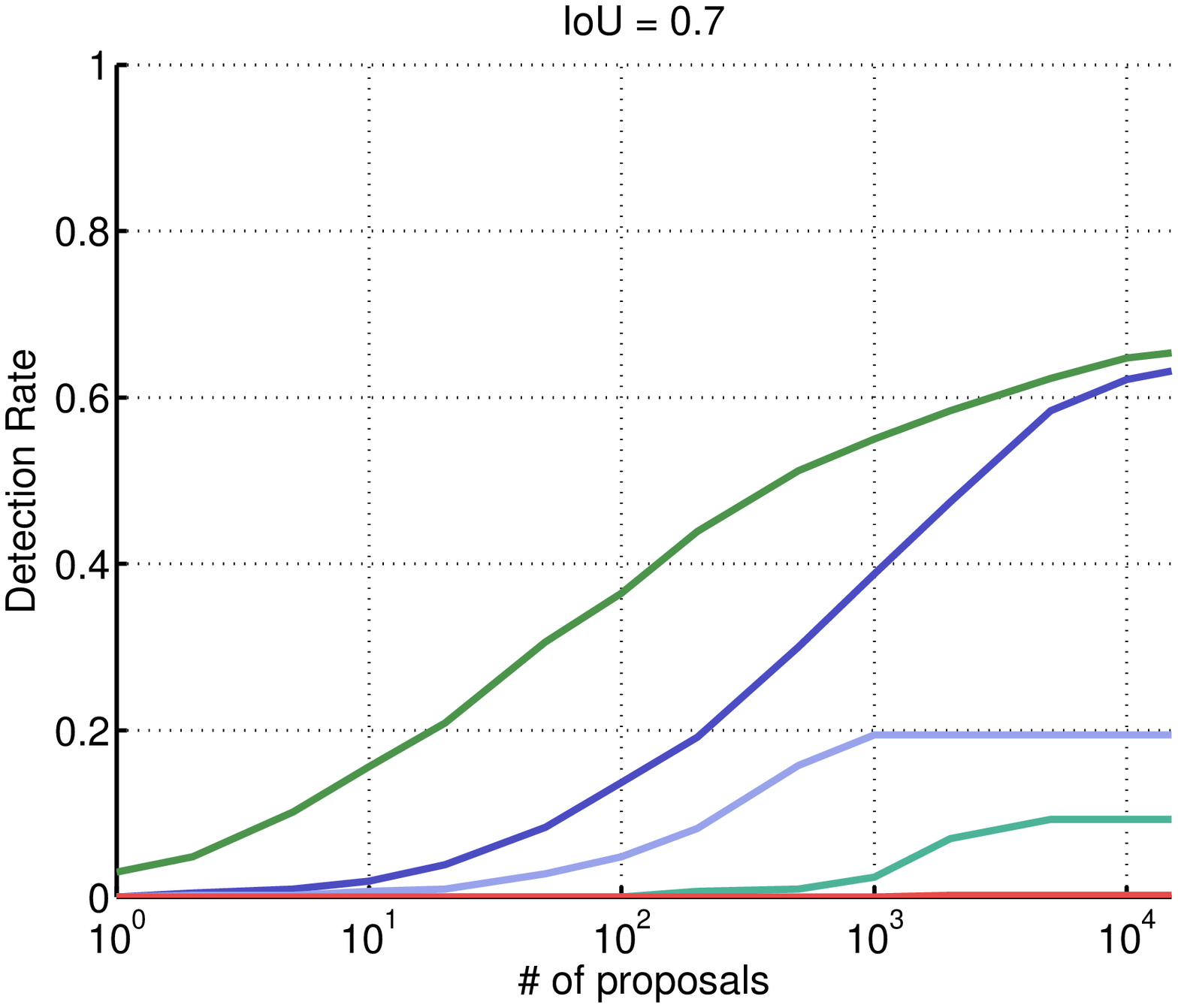} \includegraphics[width=0.32\linewidth]{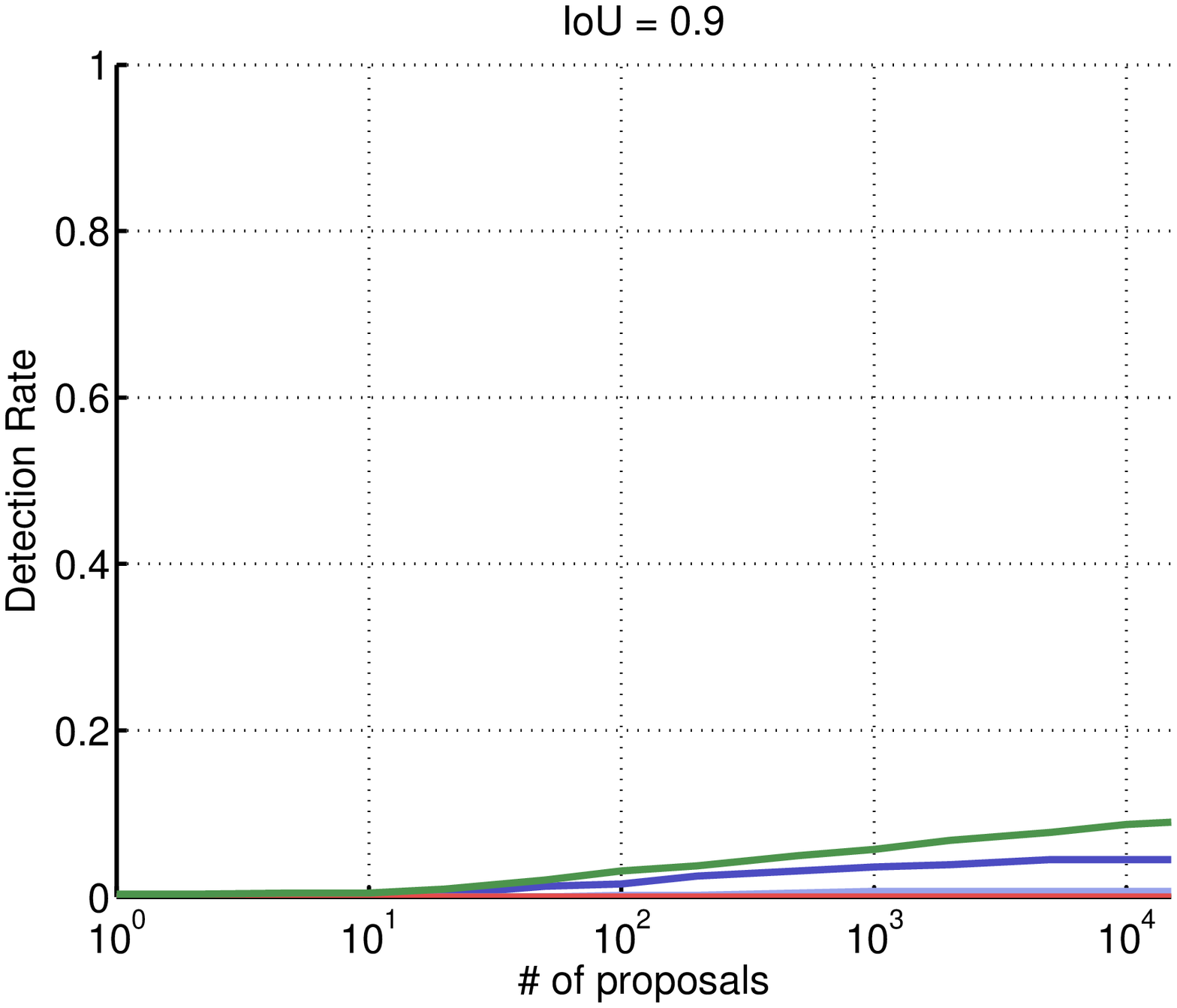}\\
\fbox{\includegraphics[width=0.60\linewidth]{legend}}\\
\includegraphics[width=0.32\linewidth]{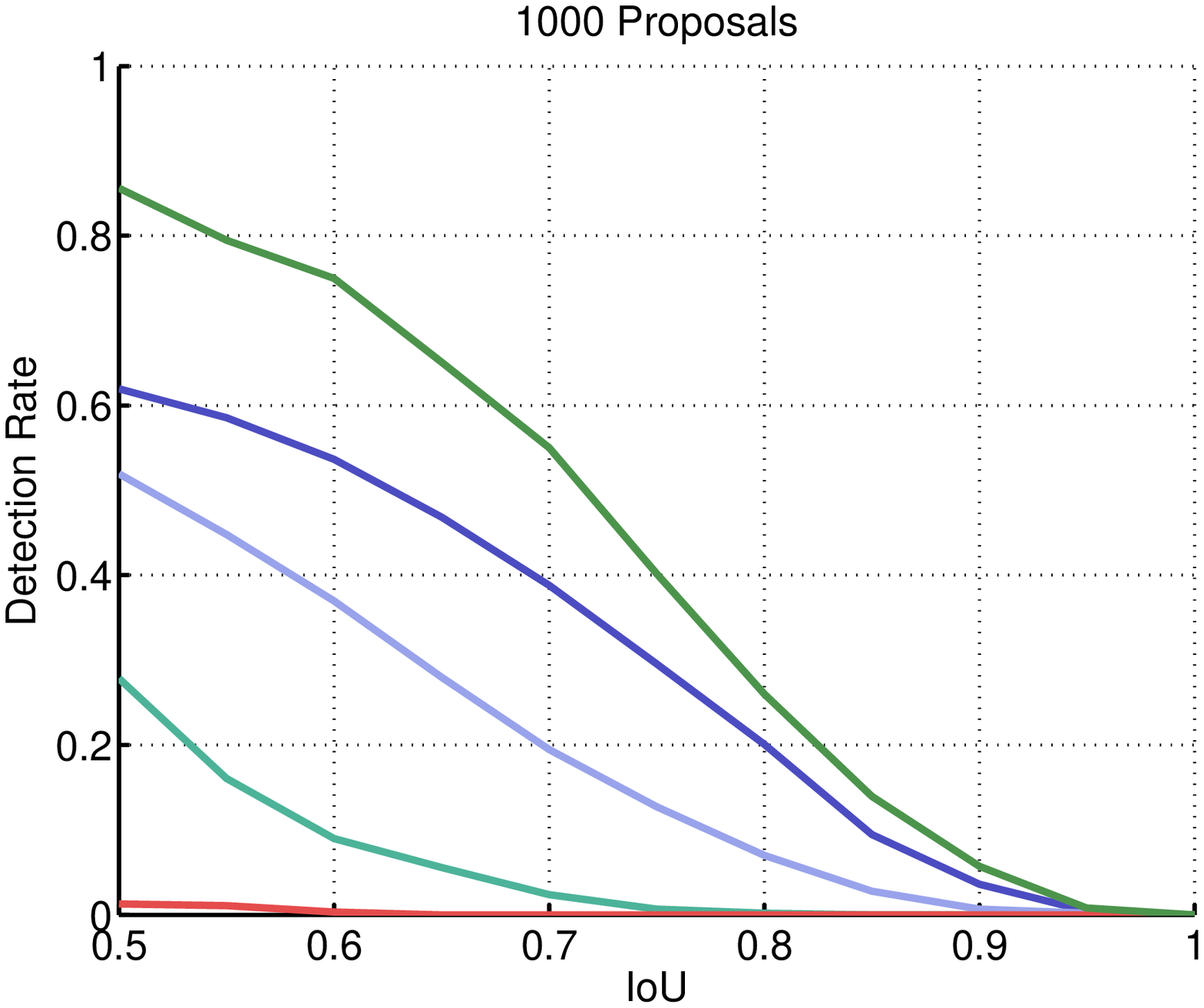} \includegraphics[width=0.32\linewidth]{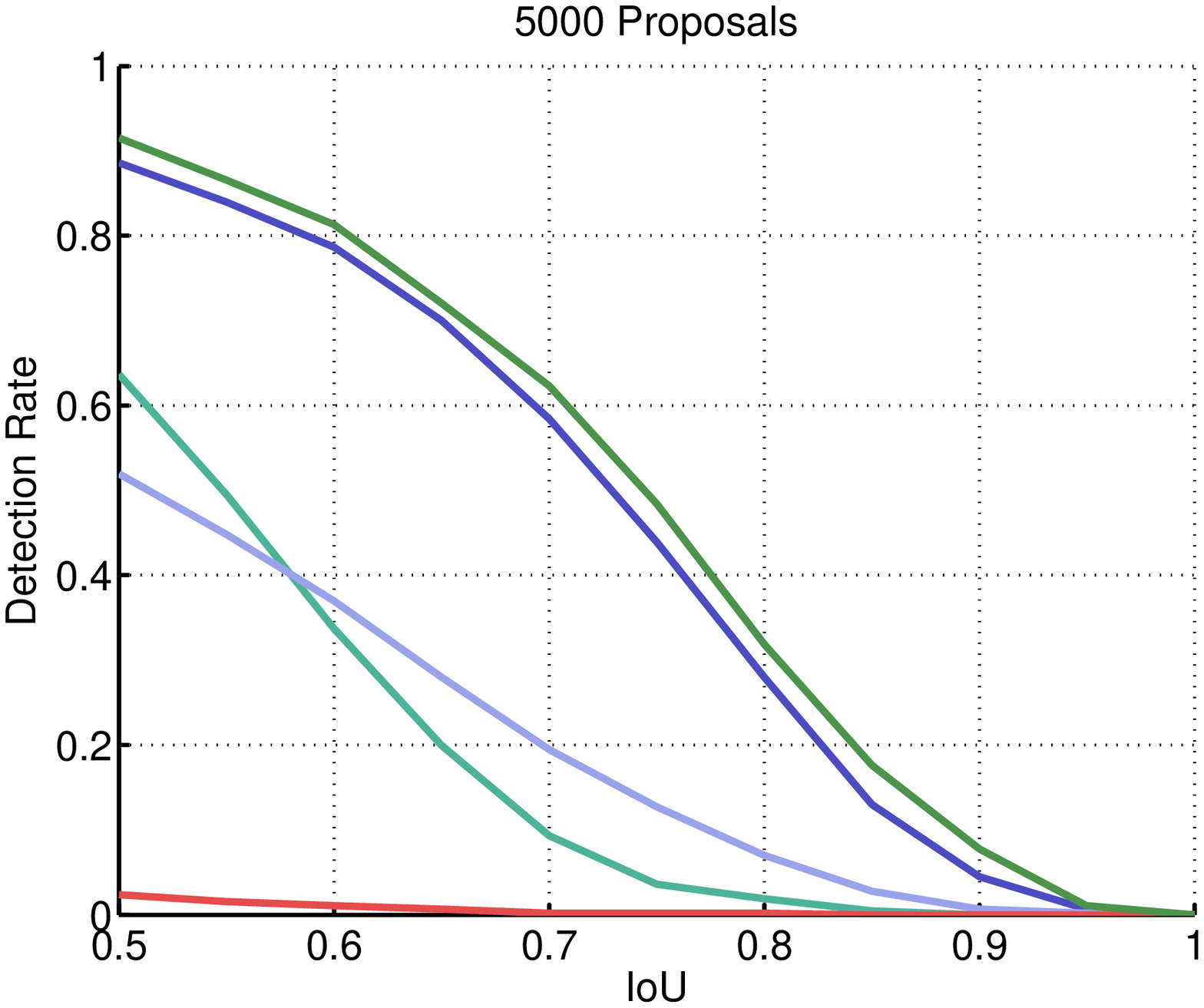} \includegraphics[width=0.32\linewidth]{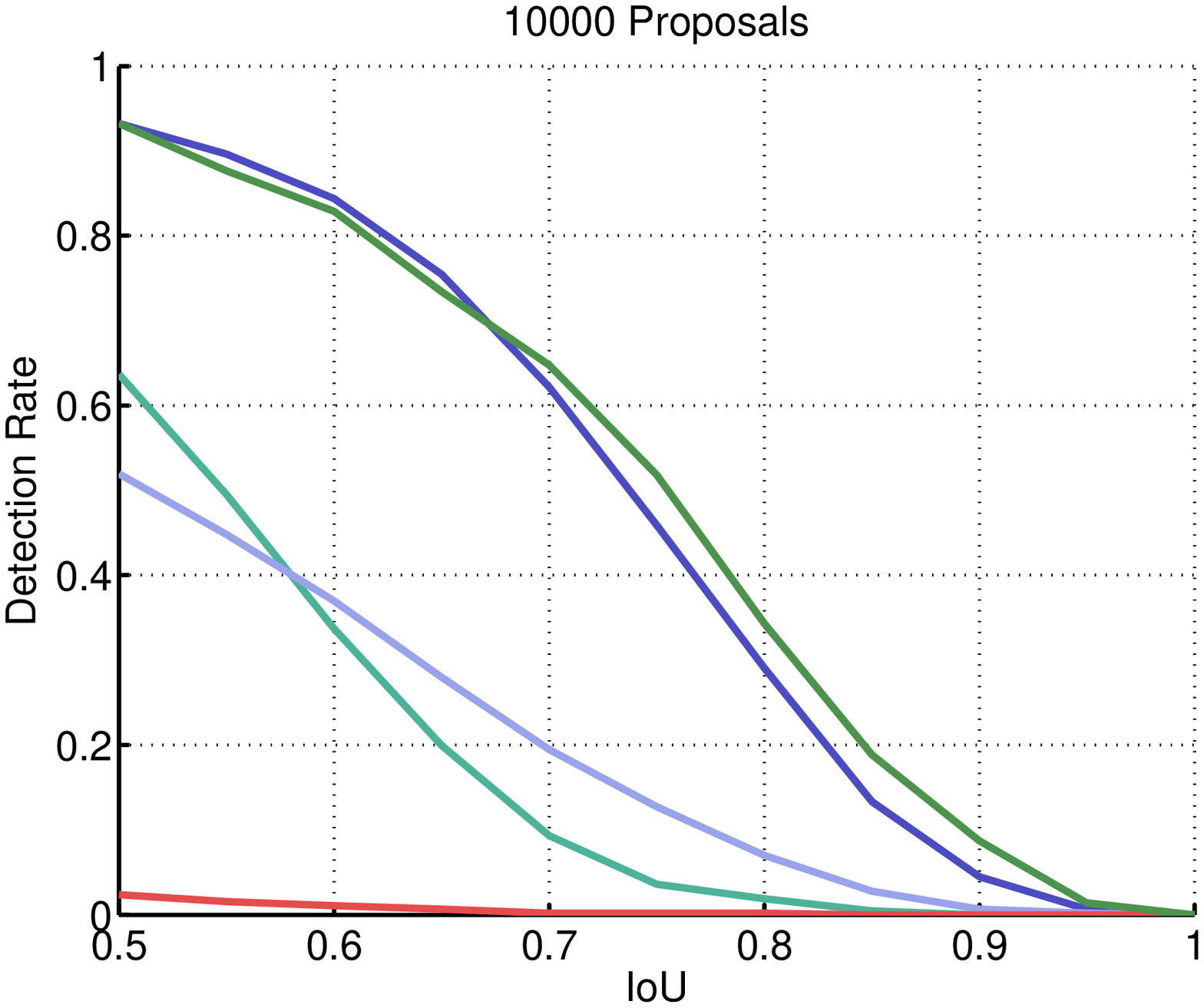}
\end{center}
   \caption{A comparison of various state-of-the-art object proposals methods in the SVT dataset. (top) Detection rate versus number of proposals for various intersection over union thresholds. (bottom) Detection rate versus intersection over union threshold for various fixed numbers of proposals.}
\label{fig:plot_svt}
\end{figure}

\begin{figure}
\centering
\includegraphics[width=\linewidth]{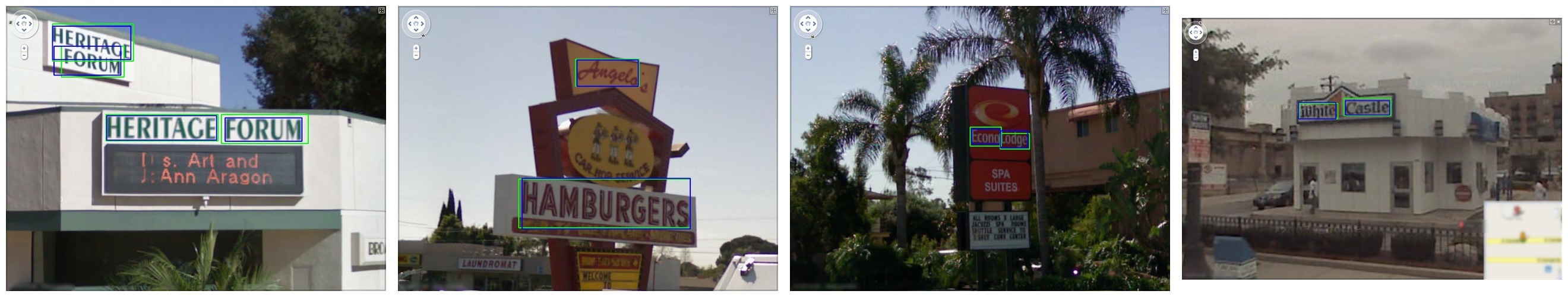}
\caption{Word proposals generated by our method (blue) with better Intersection over Union (IoU) with each of the annotated ground truth bounding boxes (green) in sample images from the SVT dataset.}
\label{fig:data_svt}
\end{figure}

Table~\ref{tab:svt_results} and Figures~\ref{fig:plot_svt} and ~\ref{fig:data_svt} show analogous experiments for the SVT dataset. 

\begin{table}
\begin{center}
\scriptsize
    \begin{tabularx}{\linewidth}{ X c c c c c}
\toprule
    Method & \# prop. & 0.5 IoU & 0.7 IoU & 0.9 IoU & time(s)\\
\midrule
    BING~\cite{cheng2014} & 2987 & 0.64 & 0.09 & 0.00 & 0.81\\  
    EdgeBoxes~\cite{zitnick2014} & 15319 & \textbf{0.94} & 0.63 & 0.04 & 2.71\\  
    RP~\cite{manen2013} & 5620 & 0.02 & 0.00 & 0.00 & 10.51\\  
    GOP~\cite{krahenbuhl2014} & 778 & 0.53 & 0.19 & 0.03 & 4.31\\  
\midrule
 \textbf{TextProposals}& 17358 & \textbf{0.94} & \textbf{0.65} & \textbf{0.09} & 3.21\\ 
\bottomrule 
    \end{tabularx}
\end{center}
\caption{Average number of proposals, recall at different IoU thresholds, and running time comparison with Object Proposals state of the art algorithms in the SVT dataset.}
\label{tab:svt_results}
\end{table}

Results on the SVT dataset exhibit a distinct scenario than in ICDAR2013. In general, recall rates are lower for all evaluated methods while still there is a clear difference between the two best performing methods (TextProposals and EdgeBoxes) and the rest. However, TextProposals and EdgeBoxes maximum recalls at $0.5$ IoU threshold are equal in SVT, while TextProposals is slightly better at $0.7$ and $0.9$. The difference between the results in the ICDAR2013 and SVT datasets can be explained because both datasets are very different in nature, SVT contains more challenging text, with lower quality and many times under bad illumination conditions, while in ICDAR2013 text is mostly well focused and flatly illuminated. 

Moreover, in this dataset our method does not provide the same stability property shown before for large IoU thresholds. This behavior is also related with the distinct nature of the datasets, but also, as discussed before, with the fact that SVT ground-truth annotations are less consistent in terms of the extra padding allowed around word instances. 

Still top plots in Figure~\ref{fig:plot_svt} demonstrate that the AUC of our TextProposals are much better that the ones of EdgeBoxes at all IoU thresholds. This provides an noticeable boost in performance when we limit our analysis to a relatively small set of proposals (e.g. as for $1000$ proposals in the bottom-left plot in Figure~\ref{fig:plot_svt}).

In order to evaluate our method in more unconstrained scenarios we conduct a similar analysis on the MLe2e and ICDAR2015 datasets. In this experiment we only evaluate the TextProposals and EdgeBoxes algorithms. The MLe2e dataset contains well-focused and horizontal text instances in four different scripts. 
On the other hand, while the type of text found in the ICDAR2015 ``Incidental Scene Text'' dataset can be considered similar to the one in SVT, the ICDAR2015 contains a large number of non-horizontal and very small text instances.

\begin{figure*}
\begin{center}
\includegraphics[width=0.32\linewidth]{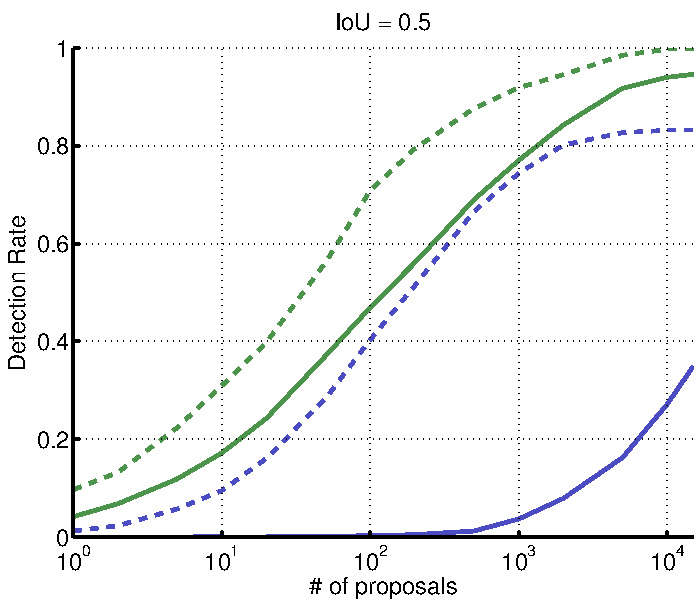} \includegraphics[width=0.32\linewidth]{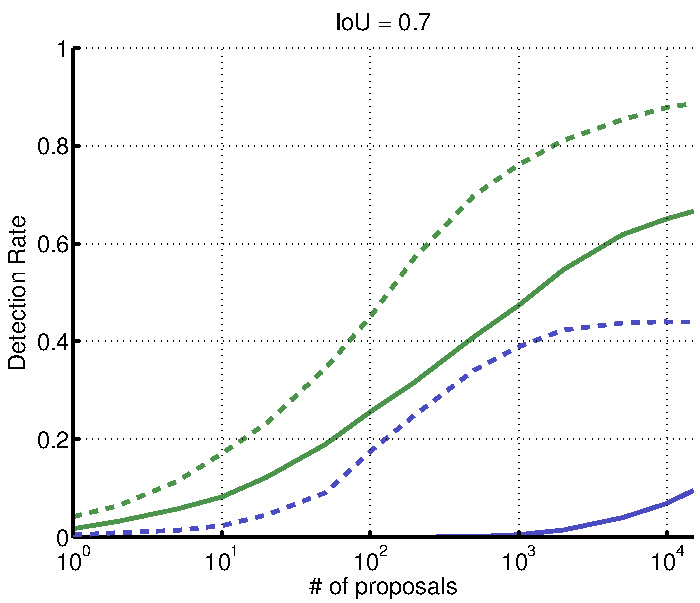} \includegraphics[width=0.32\linewidth]{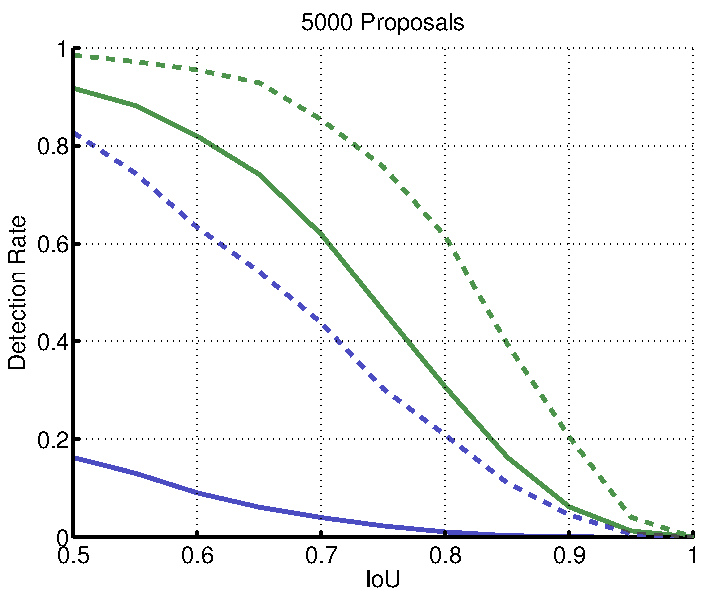}\\
\end{center}
   \caption{A comparison of our TextProposals (green) with EdgeBoxes (blue) in the MLe2e (dashed lines) and ICDAR2015 (solid lines) datasets.}
\label{fig:plot_mle2e}
\end{figure*}

\begin{figure*}
\centering
\includegraphics[width=\linewidth]{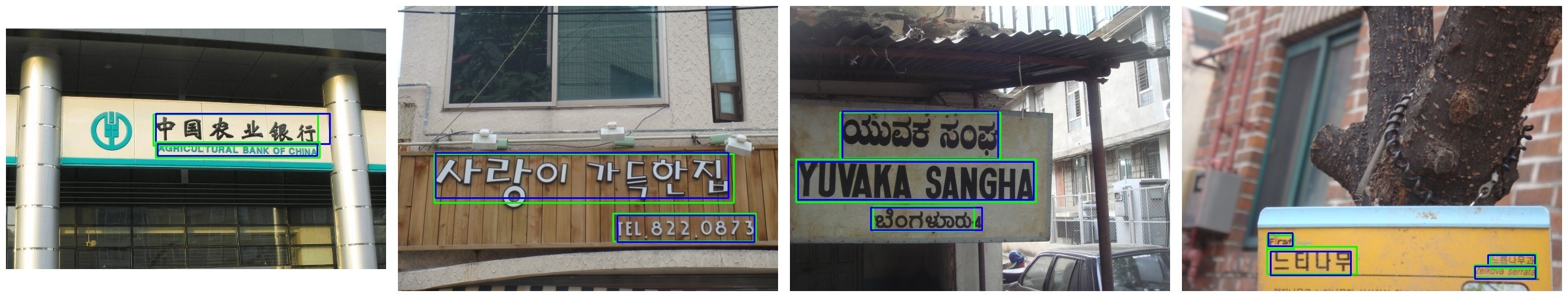}
\caption{Word proposals generated by our method (blue) with better Intersection over Union (IoU) with each of the annotated ground truth bounding boxes (green) in sample images from the MLe2e dataset.}
\label{fig:data_mle2e}
\end{figure*}

Figure~\ref{fig:plot_mle2e} shows performance plots for our TextProposals and the EdgeBoxes detector in these two datasets. Figure~\ref{fig:data_mle2e} show the best word proposals generated by our method on MLe2e images. As can be appreciated TextProposals is clearly superior to EdgeBoxes in both cases. 


The most important observation from Figure~\ref{fig:plot_mle2e} is the tiny recall rate of EdgeBoxes in the ICDAR2015 dataset. This result makes clear that the EdgeBoxes algorithm is not well suited for detecting non-horizontal and small-sized text. Contrary to EdgeBoxes, our TextProposals perform similarly in both SVT and ICDAR2015 datasets. Thus, proving to be more robust on detecting these challenging types of text.

\begin{figure}
\centering
\subfloat[]{\includegraphics[width=0.4\linewidth]{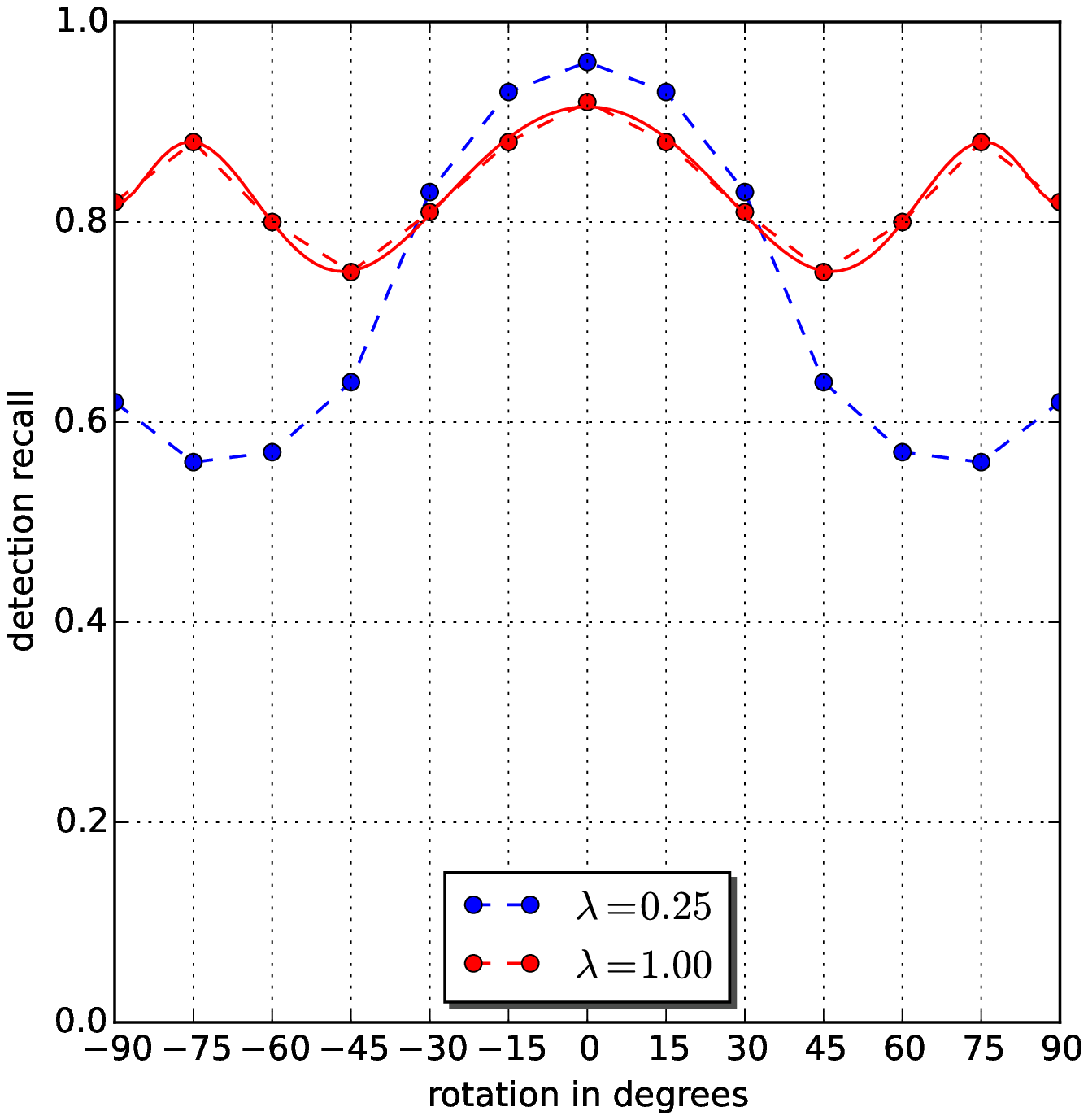}}
\subfloat[]{\includegraphics[width=0.6\linewidth]{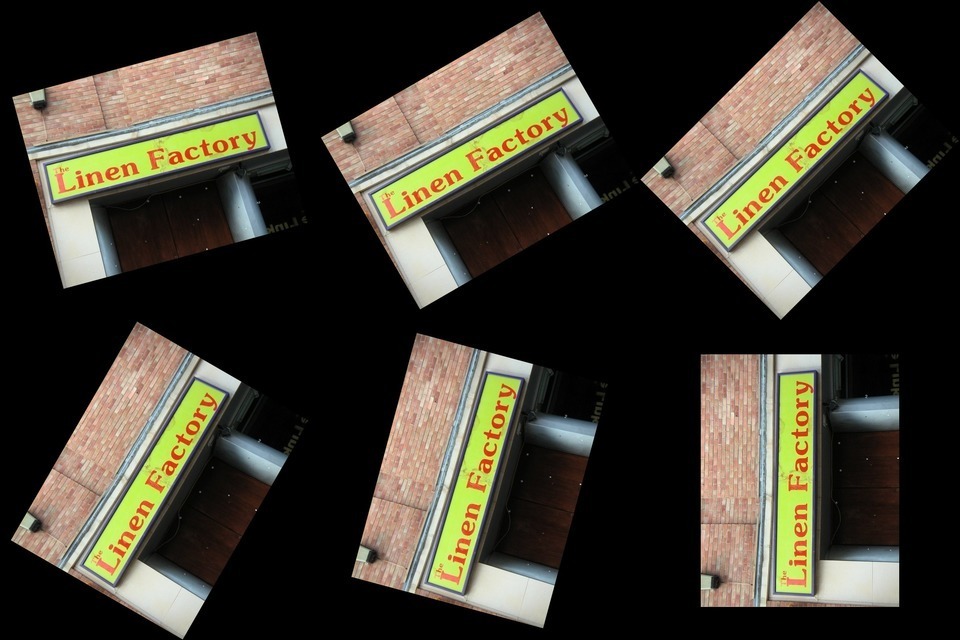}}
\caption{TextProposals detection recall at $0.7$ IoU as a function of the text orientation degree (a). We provide recall for two different values of the $\lambda$ parameter in the SLC distance metric (equation~\ref{eq:dist2}). Recall values are calculated on different rotated versions of the ICDAR2013 dataset (b).}
\label{fig:recall_vs_rotation}
\end{figure}

In order to further evaluate the performance of TextProposals on arbitrarily oriented text detection, we have done an extra experiment where our method is applied to different versions of the ICDAR2013 dataset where images were deliberately rotated at various degrees. Figure~\ref{fig:recall_vs_rotation} shows the overall detection recall of the method at $0.7$ IoU as a function of the text orientation degree, as well as the generated rotated versions ($15^{\circ}$, $30^{\circ}$, $45^{\circ}$, $60^{\circ}$, $75^{\circ}$, and $90^{\circ}$) for one sample image of the ICDAR dataset. We provide the obtained recall for two different values of the $\lambda$ parameter in the SLC distance metric (equation~\ref{eq:dist2}). As mentioned in section~\ref{sec:grouping}, setting $\lambda = 1$ makes our clustering analysis rotation invariant while smaller values are better for detecting horizontally aligned text. In fact, we have found by manual inspection that the recall oscillation for the $\lambda = 1$ curve in Figure~\ref{fig:recall_vs_rotation} is due to errors introduced on the IoU calculation for non axis-aligned bounding boxes. Moreover, we appreciate that the version used in the rest of the experiments of this section, with $\lambda = 0.25$, performs robustly for text with slight rotations (up to $30^{\circ}$) which is the common scenario in most of the cases.

\subsubsection{Error Analysis}
In this section we offer a brief analysis of failure cases in order to identify the limitations of the method. For this error analysis we focus on the SVT dataset because, as we have seen in the previous experiments, it provides a more challenging benchmark for our method. SVT text instances are many times smaller and have lower quality than the ones found in other scene text datasets, and thus are more difficult to detect. 

\begin{figure}
\centering
\includegraphics[width=\linewidth]{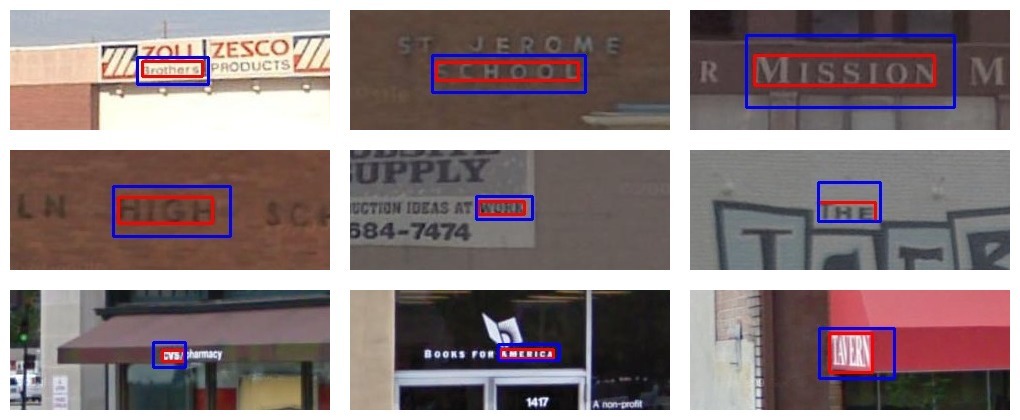}
\caption{Examples of correctly detected words that are computed as missdetections due to human annotation inconsistency. Blue boxes correspond to ground truth annotations and red boxes to TextProposals results.}
\label{fig:svt_no_err}
\end{figure}

\begin{figure}
\centering
\includegraphics[width=\linewidth]{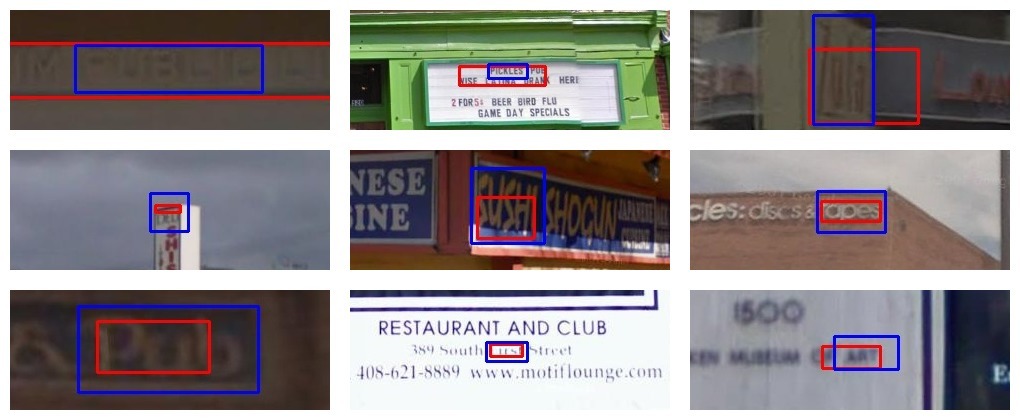}
\caption{Examples of errors of the TextProposals algorithm. Blue boxes correspond to ground truth annotations and red boxes to the TextProposals hypotheses with a larger IoU.}
\label{fig:svt_err}
\end{figure}

Our analysis consists in manually inspecting the particular images where the TextProposals algorithm does not provide correct detections. If we consider the intersection over union (IoU) detection threshold of $0.5$, we found that there are only $39$ words (out of $647$) in the SVT test set for which our method is not able to provide a correct bounding box. From these $39$ text instances, $20$ of them correspond to cases where the ground truth bounding boxes are not well annotated as shown in Figure~\ref{fig:svt_no_err}. From the remaining $19$ errors a representative set is shown in Figure~\ref{fig:svt_err}. We appreciate that they mostly correspond to extremely difficult cases in terms of contrast and/or image quality. This kind of degraded texts effectively supposes a limitation for the proposed method, as the initial segmentation will rarely provide meaningful regions for them to be detected. A possible solution to this limitation would be to use the whole component tree of the image as the set of initial regions in our algorithm. However, by doing so the number of generated proposals would be much larger.

Actually, apart of the problems in detecting those difficult text instances, we can see as a more important limitation of the method the large number of proposals provided. An interesting observation here is that while class-independent object detection generic methods suffice with near a thousand proposals to achieve high recalls, in the case of text we still need around $10000$ in order achieve similar rates. This indicates that there is a large room for improvement in specific text object proposals methods.

\subsection{End-to-end word spotting}
\label{sec:exp_end2end}
In this section we build an end-to-end scene text recognition pipeline by combining our TextProposals with two state-of-the-art holistic word recognizers: the word embeddings method of Almazan~\etal~\cite{Almazan2014} and the holistic CNN classifier of Jaderberg~\etal~\cite{jaderberg2014reading}. The evaluation framework used in all this section is the standard for end-to-end text recognition datasets~\cite{Wang2010,karatzas2015}. A result bounding box is counted as a correct match if it overlaps a ground truth bounding box by more than 50\% and the provided word transcription is correct (ignoring case). Based on this simple rule a single F-score measure is calculated for a given method from the standard precision and recall metrics.

\begin{figure*}[t]
\centering
\includegraphics[width=0.32\linewidth]{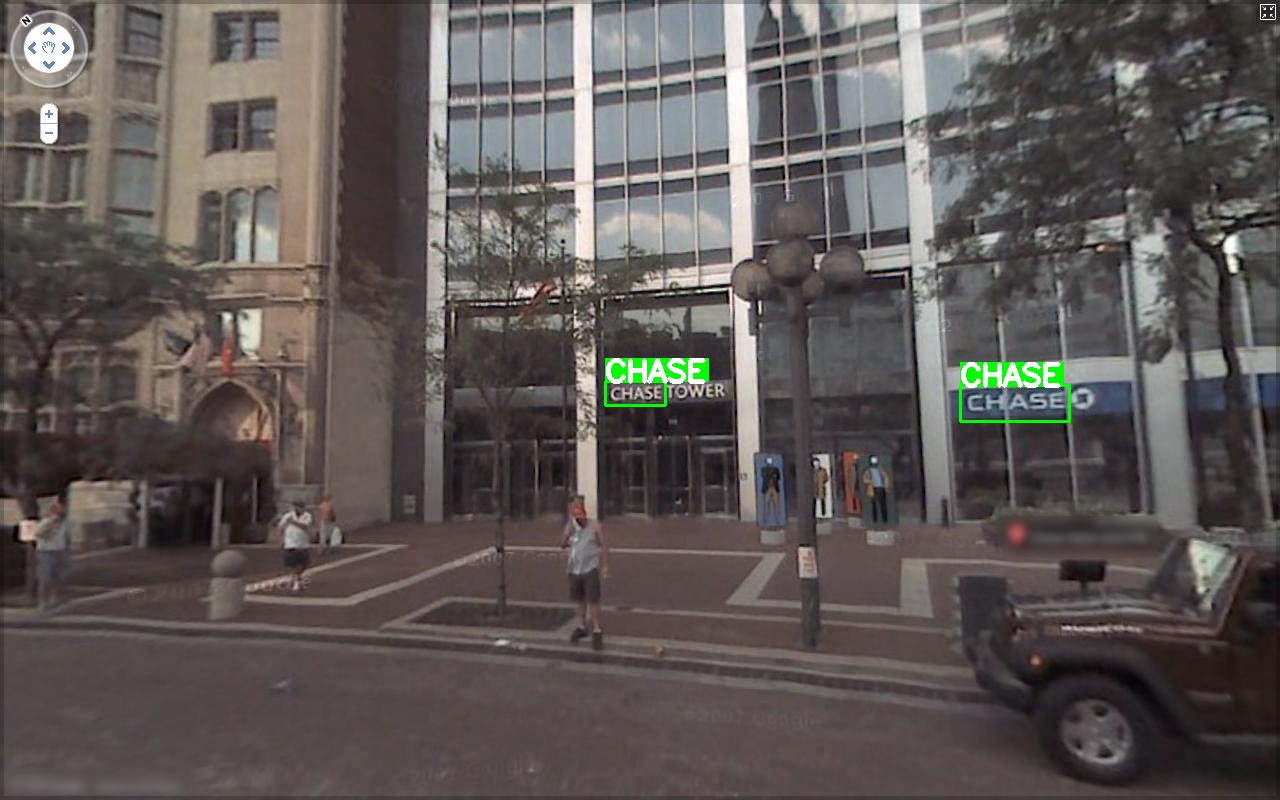}
\includegraphics[width=0.32\linewidth]{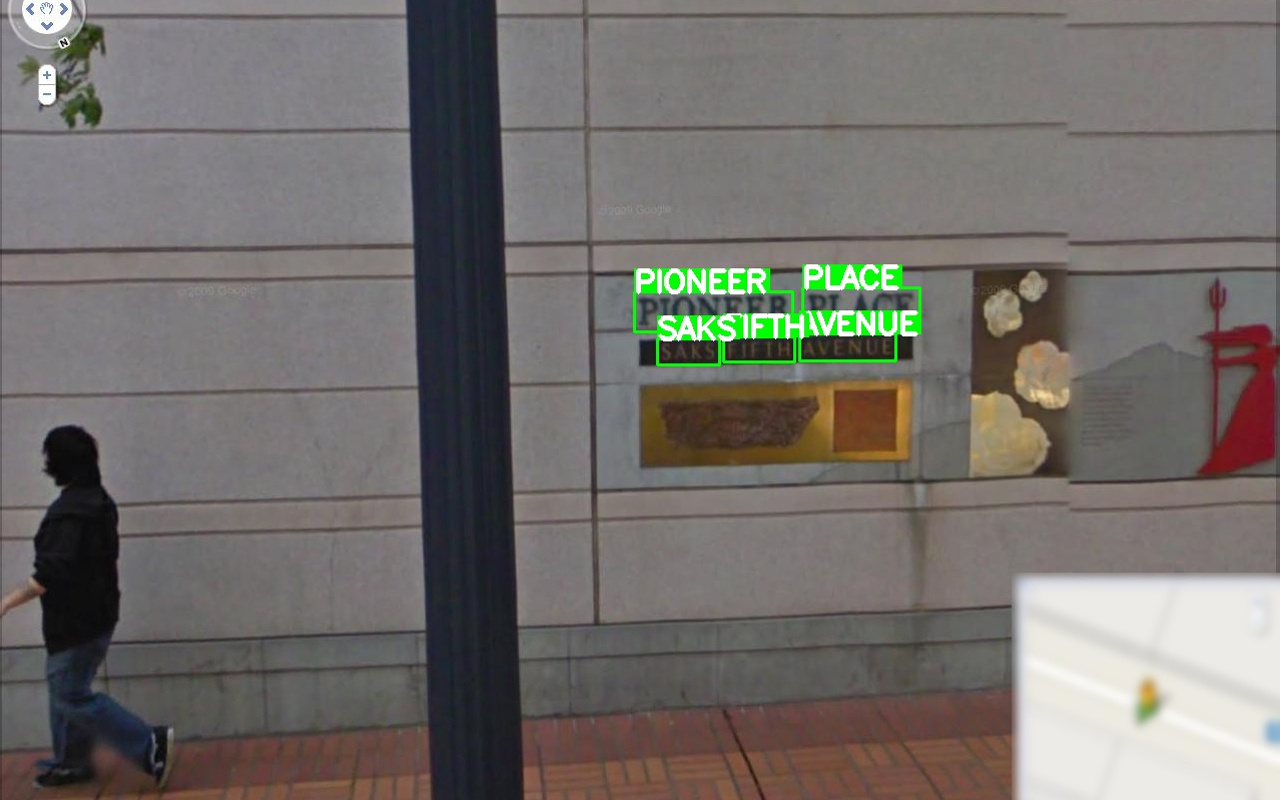}
\includegraphics[width=0.32\linewidth]{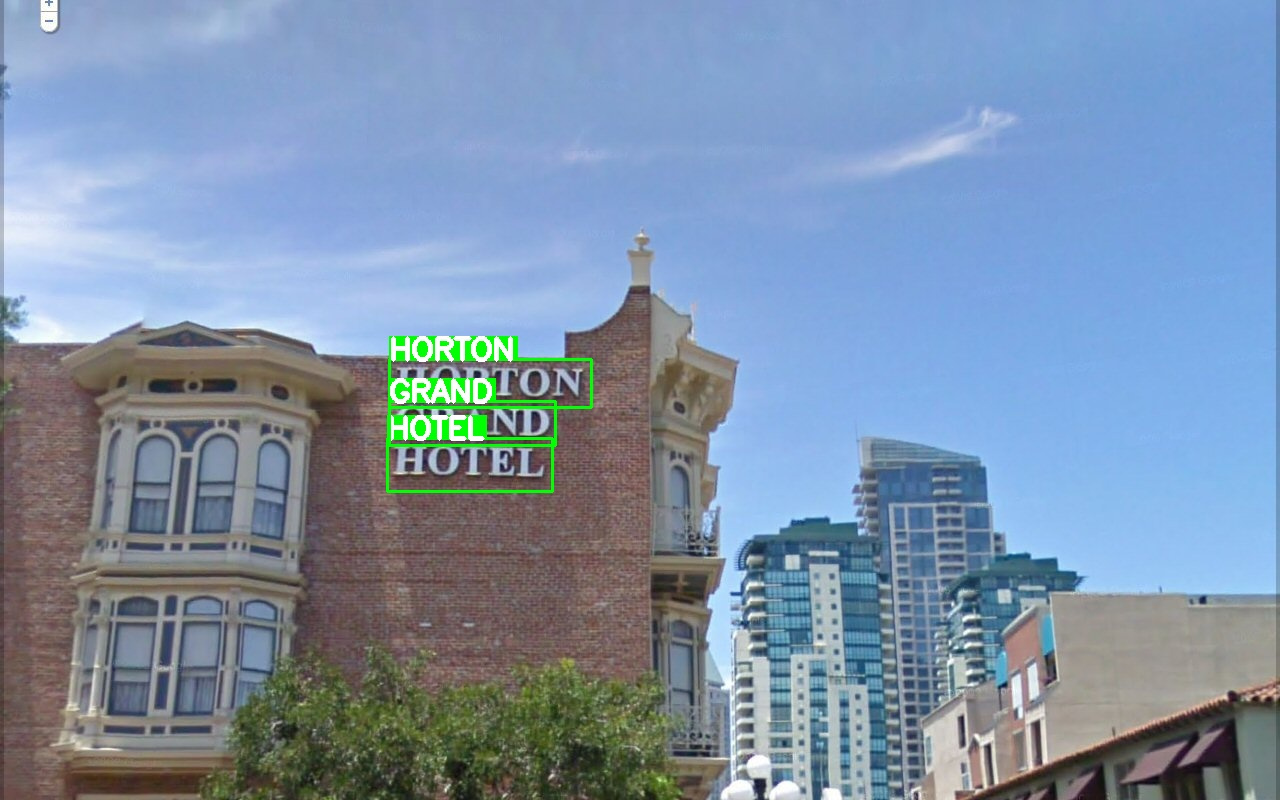}
\caption{Word spotting results using our TextProposals + DictNet~\cite{jaderberg2014reading} CNN model on SVT sample images.}
\label{fig:out_res_spotting}
\end{figure*}

Table~\ref{table:endrec1} shows the obtained end-to-end word spotting F-scores on ICDAR2003 and SVT datasets and compare them with the state-of-the-art. The combination of TextProposals with Jaderberg~\etal~\cite{jaderberg2014reading} CNN model outperforms the best previously published results, which are actually from ~\cite{jaderberg2014reading}, in most of the columns in the Table. The other variant, using Almazan~\etal~\cite{Almazan2014} for recognition, provides also competitive results. Figure~\ref{fig:out_res_spotting} shows qualitative word spotting results of the combination of TextProposals with Jaderberg~\etal~\cite{jaderberg2014reading} CNN model on ICDAR2003 and SVT sample images. 

\begin{table}[h]
\begin{center}
\scriptsize
\begin{tabularx}{\textwidth}{ X c c c c c}
\toprule
                                   		&IC03-50&IC03-Full&IC03&SVT-50&SVT\\
 \midrule
Wang \etal \cite{Wang2011}            		& 0.68  & 0.61 	& -    & 0.38 & -    \\
Wang and Wu \cite{Wang2012}           		& 0.72  & 0.67 	& -    & 0.46 & -    \\
Alsharif \cite{Alsharif2014}          		& 0.77  & 0.70 	& 0.63*& 0.48 & -    \\
Jaderberg \etal \cite{Jaderberg2014}  		& 0.80  & 0.75 	& -    & 0.56 & -    \\
Jaderberg \etal \cite{jaderberg2014reading} 	& 0.90  & 0.86 	& 0.78 & 0.76 & 0.53 \\
 \midrule
\textbf{TextProposals} + Watts \cite{Almazan2014}	& 0.82  & 0.73  & -    & 0.67 & -    \\
\textbf{TextProposals} + DictNet \cite{jaderberg2014reading} &\textbf{0.92}&\textbf{0.90}&0.75&\textbf{0.85}&\textbf{0.54}\\
\bottomrule
\end{tabularx}
\end{center}
\caption{Comparison of end-to-end word spotting F-scores on the ICDAR2003 and SVT datasets.}
\label{table:endrec1}
\end{table}

In Table~\ref{table:endrec1} it is particularly interesting the comparison with the end-to-end pipeline in~\cite{jaderberg2014reading}. Since both pipelines make use of object proposals algorithms and the same final recognition model, this comparison directly relates to the quality of the object proposals algorithms. This demonstrates that our TextProposals provide an extra boost of performance to the end-to-end system. 


Tables~\ref{tab:Results2.4} and~\ref{tab:Results4.4} show a comparison of our end-to-end pipeline with the participants in the last ICDAR Robust Reading Competition on the ICDAR2013 (focused text) and ICDAR2015 (incidental text) datasets. As can be appreciated, the combination of TextProposals with the DictNet~\cite{jaderberg2014reading} CNN recognizer shows competitive results on the ICDAR2013 dataset, while outperforms with a clear margin all the competition participants on the ICDAR2015 Incidental Text challenge.

\begin{table*}[t!]
\centering
\scriptsize
\begin{tabularx}{\textwidth}{ X c c c c c c}
\toprule
                               & \multicolumn{3}{c|}{End-to-End results} &  \multicolumn{3}{c}{Word spotting Results} \\ 
\midrule
                                          &IC13-100&IC13-Full&IC13&IC13-100&IC13-Full&IC13\\
\midrule
BeamSearch CUNI +S                         & 26.38 & 23.32 & 20.28 & 28.17 & 24.95 & 21.94\\
OpenCV+Tessaract\cite{gomez2014scene}          & 59.47 & 56.14 & 43.29 & 63.05 & 59.43 & 44.46\\
BeamSearch CUNI                            & 63.20 & 61.10 & 56.04 & 67.34 & 65.05 & 59.38\\
MSER-MRF~\cite{liu2015natural}             & 71.13 &  -    &  -    & 75.74 &  -    &  -   \\
Deep2Text-I~\cite{Yin2013,jaderberg2014reading}  & 74.36 & 74.36 & 74.36 & 76.93 & 76.93 & 76.93\\
NJU Text                                   & 74.49 &  -    &  -    & 77.96 &  -    &  -   \\ 
Deep2Text-II~\cite{Yin2013,jaderberg2014reading} & 75.29 & 75.29 &\bf{75.29}&77.37&77.37 & 77.37\\
TextSpotter~\cite{Neumann2012}    & 77.02 & 63.19 & 54.28 & 81.84 & 66.48 & 56.69\\
Stradvision-1                              & 81.28 & 78.51 & 67.15 & 85.82 & 82.84 & 70.19\\
VGGMaxBBNet~\cite{jaderberg2014reading,Jaderberg2014}    &\bf{86.18}& -  & -     &\bf{90.25}& -  &  -   \\
\midrule
\textbf{TextProposals} + DictNet \cite{jaderberg2014reading} & 81.16 &\bf{79.49}& 68.54 & 85.37 &\bf{83.58}& 70.71\\
\bottomrule
\end{tabularx}
\caption{Comparison of end-to-end recognition and word spotting F-scores on ICDAR2013 (Focused Text) dataset.}
\label{tab:Results2.4}
\end{table*}

\begin{table*}[t!]
\centering
\scriptsize
\begin{tabularx}{\textwidth}{ X c c c c c c}
\toprule
                               & \multicolumn{3}{c|}{End-to-End results} &  \multicolumn{3}{c}{Word spotting Results} \\ 
\midrule
                                          &IC15-50&IC15-Full&IC15&IC15-50&IC15-Full&IC15\\

\midrule
Beam Search CUNI +S                       & 13.26 & 10.85 & 6.86 & 14.01 & 11.48 & 7.24 \\  
OpenCv + Tessaract~\cite{gomez2014scene}      & 13.84 & 12.01 & 8.01 & 14.65 & 12.63 & 8.43 \\  
Deep2Text-MO~\cite{Yin2013,jaderberg2014reading}&16.77&16.77&16.77&17.58 & 17.58 & 17.58\\  
Beam Search CUNI                          & 22.14 & 19.80 & 17.46& 23.37 & 21.07 & 18.38\\  
NJU Text                                  & 32.63 &  -    &  -   & 34.10 &  -    &  -   \\  
Stradvision-1                             & 33.21 &  -    &  -   & 34.65 &  -    &  -   \\  
TextSpotter~\cite{Neumann2012}   & 35.06 & 19.91 & 15.60& 37.00 & 20.93 & 16.38\\  
Stradvision-2                             & 43.70 &  -    &  -   & 45.87 &  -    &  -   \\  
Megvii-Image++                            & 46.74 & 40.00 & 32.86& 49.95 & 42.71 & 34.57\\  
\midrule
\textbf{TextProposals} + DictNet \cite{jaderberg2014reading} &\bf{53.30}&\bf{49.61}&\bf{47.18}&\bf{56.00}&\bf{52.26}&\bf{49.73} \\
\bottomrule
\end{tabularx}
\caption{Comparison of end-to-end recognition and word spotting F-scores on ICDAR2015 (Incidental Text) dataset.}
\label{tab:Results4.4}
\end{table*}

It is important to notice here that ICDAR2015 is a much more challenging dataset than ICDAR2013. In the case of ICDAR2013 Focused Text dataset, specialized text detectors can still perform very well in combination with strong statistical models for recognition. See for example the very good performance of methods combining traditional text detectors with holistic word recognition CNNs (Deep2Text-I and Deep2Text-II), Recurrent Neural Networks (Megvii-Image++), or even the more traditional shape based character classifiers with dictionary-based language models (TextSpotter). 

On the other hand, the incidental text instances found in ICDAR2015 dataset are normally less suitable for traditional specialized text detectors. Incidental text is many times very small in size, has low quality, and/or is not strictly horizontal. This qualities make of our TextProposals a better solution in this dataset. 


Moreover, it is also important to notice that the DictNet~\cite{jaderberg2014reading} CNN model that we integrate with our TextProposals is limited to a list of 90k words that was not designed for the ICDAR2015 dataset. While it includes all words in ICDAR2003 and SVT test sets ground-truth, the list of 90k words matches only 70\% of the words that appear in the ICDAR2015 test set, so limiting the maximum attainable recall to $0.7$. 

\section{Conclusion}
\label{sec:conclusion}


In this paper we have presented a text specific object proposals algorithm that is able to reach impressive recall rates with a few thousand proposals in different standard datasets, including focused or incidental text, and multi-language scenarios.

We have seen how the proposed algorithm, while still rooted in the same intuitions developed in existing specialized text detectors, introduces important methodological contributions. By not making any assumption about the nature of the initial set of regions (connected components) to analyze or about the structure of the region groupings that are of our interest, we ended up with a less rigid definition of the involved grouping process. This methodological shift has proved to be beneficial from the perspective of an object proposals approach.

The performed experiments allow to conclude that text-specific object proposals are a realistic alternative to generic object proposals algorithms, and also to specialized text detectors. This is further supported by experimental evidence showing how the use of our TextProposals leads to improve the state-of-the-art in challenging end-to-end scene text recognition datasets.

%
%

\section*{Acknowledgment}
\label{sec:acknowledgement}
This project was supported by the Spanish project TIN2014-52072-P, the fellowship RYC-2009-05031, and the Catalan government scholarship 2014FI\_B1-0017.



\section*{References}
  \bibliographystyle{elsarticle-num} 
  \bibliography{GomezKaratzas_pr_2016b}





\end{document}